\documentclass[acmtog]{acmart}

\usepackage{booktabs} 
\usepackage{subcaption}

\usepackage{soul}
\usepackage{multirow}
\usepackage{xcolor}
\usepackage{xspace}
\definecolor{graytext}{RGB}{130,130,130}

\newcommand{\whitetxt}[1]{{\color{white}#1}\normalfont}
\newcommand{\ignorethis}[1]{}
\newcommand{\etal}{\textit{et al.}}

\def\methodName{Spice\texorpdfstring{$\cdot$}{\-}E\xspace} 
\def\shapE{Shap$\cdot$E\xspace}

\newcommand{\acronym}{Spice\texorpdfstring{$\cdot$}{\-}E (\textbf{S}tructural \textbf{P}riors \textbf{i}n 3D Diffusion Models using \textbf{C}ross-\textbf{E}ntity Attention)\xspace}

\newbox\jsavebox
\newcommand{\jsubfig}[2]{%
	\sbox\jsavebox{#1}%
	\parbox[t]{\wd\jsavebox}{\centering\usebox\jsavebox\\#2}%
	}

\author{Etai Sella}\authornote{Denotes equal contribution}
\affiliation{%
  \institution{Tel Aviv University}
  \city{Tel Aviv}
  \country{Israel}
}

\orcid{0009-0004-0079-0046}
\author{Gal Fiebelman}\authornotemark[1]
\affiliation{%
  \institution{Tel Aviv University}
  \city{Tel Aviv}
  \country{Israel}
}

\orcid{0009-0002-2119-9808}
\author{Noam Atia}
\affiliation{%
  \institution{Tel Aviv University}
  \city{Tel Aviv}
  \country{Israel}
}
\orcid{0009-0005-4459-6891}
\author{Hadar Averbuch-Elor}
\affiliation{%
  \institution{Tel Aviv University}
  \city{Tel Aviv}
  \country{Israel}
}
\orcid{0000-0003-3476-0940}

\title{\methodName{}: Structural Priors in 3D Diffusion using Cross-Entity Attention}
\copyrightyear{2024}
\acmYear{2024}
\setcopyright{acmlicensed}\acmConference[SIGGRAPH Conference Papers '24]{Special Interest Group on Computer Graphics and Interactive Techniques Conference Conference Papers '24}{July 27-August 1, 2024}{Denver, CO, USA}
\acmBooktitle{Special Interest Group on Computer Graphics and Interactive Techniques Conference Conference Papers '24 (SIGGRAPH Conference Papers '24), July 27-August 1, 2024, Denver, CO, USA}
\acmDOI{10.1145/3641519.3657461}
\acmISBN{979-8-4007-0525-0/24/07}

\begin{abstract}
We are witnessing rapid progress in automatically generating and manipulating 3D assets due to the availability of pretrained text-to-image diffusion models. However, time-consuming optimization procedures are required for synthesizing each sample, hindering their potential for democratizing 3D content creation. Conversely, 3D diffusion models now train on million-scale 3D datasets, yielding high-quality text-conditional 3D samples within seconds. In this work, we present \methodName{} -- a neural network that adds structural guidance to 3D diffusion models, extending their usage beyond text-conditional generation. At its core, our framework introduces a cross-entity attention mechanism that allows for multiple entities---in particular, paired input and guidance 3D shapes---to interact via their internal representations within the denoising network. We utilize this mechanism for learning task-specific structural priors in 3D diffusion models from auxiliary guidance shapes. We show that our approach supports a variety of applications, including 3D stylization, semantic shape editing and text-conditional abstraction-to-3D, which transforms primitive-based abstractions into highly-expressive shapes. Extensive experiments demonstrate that \methodName{} achieves SOTA performance over these tasks while often being considerably faster than alternative methods. Importantly, this is accomplished without tailoring our approach for any specific task. We will release our code and trained models.
\end{abstract}

\begin{teaserfigure}
\includegraphics[width=0.999\textwidth]{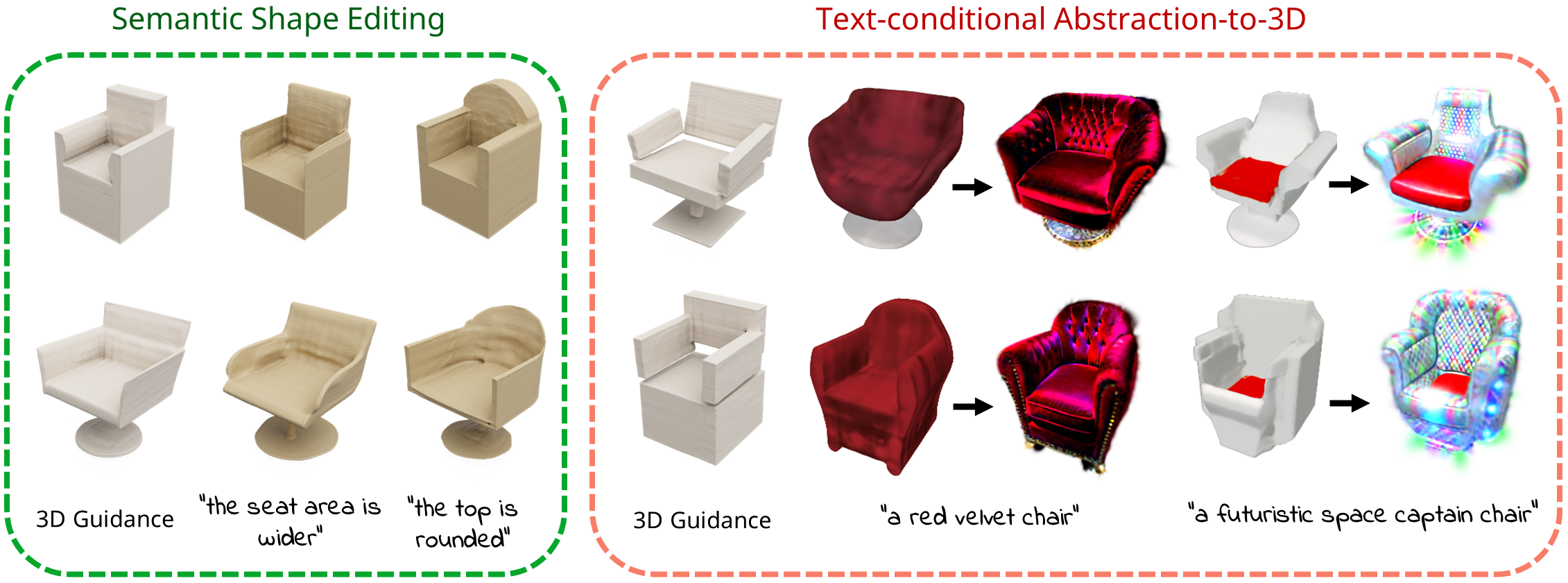}
\vspace{-15pt}
\caption{Our method adds structural guidance to 3D diffusion models. As illustrated above, this allows for generating text-conditional 3D shapes that enforce task-specific structural priors. For instance, input shapes can be semantically edited (left) and primitive-based abstractions can be transformed into high-quality textured shapes that conform with the target text (right). Our results can be optionally refined using an auxiliary process (represented by black arrows above).
}
\label{fig:teaser}
\end{teaserfigure}

\begin{document}
\maketitle



\section{Introduction}
\label{sec:intro}

Text-guided 3D generation has recently seen tremendous success, empowering us with the ability to convert our imagination into high-fidelity 3D models through the use of text~\cite{poole2022dreamfusion,wang2023prolificdreamer,wang2023score,lin2023magic3d}. 
Consequently, there has been increasing interest in leveraging this generative power for editing existing 3D objects~\cite{metzer2023latent,sella2023vox,zhuang2023dreameditor,chen2023fantasia3d}, a longstanding goal in computer vision and graphics~\cite{magnenat1988joint,lewis2023pose,igarashi2005rigid}. 
%
%
Unfortunately, these text-guided methods require timely optimization procedures for producing a single sample, as they rely on the guidance of pretrained 2D diffusion models such as Stable Diffusion~\cite{rombach2022high} over multiple rendered views, making them challenging to apply in practical settings.
%

In parallel with these advancements, million-scale 3D datasets pairing 3D data with text directly~\cite{deitke2023objaverse,deitke2023objaverse2} 
have paved the way for the creation of powerful 3D diffusion models~\cite{nichol2022point,jun2023shap}. 
These direct generative models can synthesize text-conditional 3D assets conveying complex visual concepts, and they achieve this in a matter of \emph{seconds}, orders of magnitude faster than methods utilizing 2D diffusion models. 
However, they are inherently unconstrained and lack the ability to enforce structural priors while generating 3D samples, and thereby cannot be effectively utilized in the context of 3D editing applications.

%
Inspired by recent progress adding conditional control to 2D diffusion models~\cite{zhang2023adding}, we ask: How can we provide pretrained transformer-based 3D diffusion models with task-specific structural control? And importantly, how can we achieve such structural control while preserving the model's expressive power, and to do so \emph{without} having access to (possibly) proprietary data or large computation clusters? This requires architectural modifications that maximize the utilization of pretrained weights  during model finetuning on the one hand while still acquiring task-specific structural priors from auxilary guidance shapes on the other. 
%

Accordingly, we present \acronym\footnote{pronounced ``spicy".}, a neural network that adds structural guidance to a 3D diffusion model. 
Our key observation is that the self-attention layers within transformer-based diffusion models can be modified to enable interaction between two different entities (\emph{i.e.} 3D shapes) -- one depicting the input and the other depicting the guidance entity. We introduce a cross-entity attention mechanism that mixes their latent representations by carefully combining their \emph{queries} functions, which have recently been shown for being instrumental in modifying the structure of generated images~\cite{cao2023masactrl,wu2023tune}. 
This operation allows for finetuning a 3D diffusion model to learn task-specific structural priors while preserving the model's generative capabilities. During inference, \methodName{} receives a guidance shape in addition to a target text prompt, enabling the generation of 3D shapes conditioned on both high-level text directives and low-level structural constraints. The outputs of our system can be further refined by an auxiliary process (\emph{i.e.}, \cite{yi2023gaussiandreamer}), which enhances the appearance and geometric details, albeit at the cost of increased processing time.

We show the effectiveness of our framework using  different 3D editing tasks, such as semantic shape editing and text-conditional Abstraction-to-3D, which transforms a primitive-based abstract shape into a high-quality textured shape (see Figure \ref{fig:teaser} for an illustration of these tasks). 
We perform extensive experiments, demonstrating that our approach surpasses existing methods specifically targeting these tasks,  while often being significantly faster. 

\section{Related Works}
\label{sec:related_works}
\subsection{Text-guided Shape Manipulation} 
The emergence of powerful text--image representations, most notably CLIP~\cite{Radford2021LearningTV}, has driven progress in shape editing and manipulation via language prompts. 
Several methods use CLIP for stylizing input meshes, matching their 2D image projections with a target prompt~\cite{michel2022text2mesh,chen2022tango}. 
CLIP guidance has also been exploited for generating rough un-textured shapes~\cite{sanghi2022clip,sanghi2023clip}, for optimizing a neural radiance field (NeRF)~\cite{mildenhall2021nerf} depicting the 3D object~\cite{wang2022clip,jain2022zero,lee2022understanding} and for deforming 3D meshes~\cite{gao2023textdeformer}.

This progress has been further accelerated with the rise of diffusion models, which allow for generating diverse imagery conveying complex visual concepts. 
%
%
DreamFusion~\cite{poole2022dreamfusion} introduced Score Distillation Sampling (SDS), a method that uses a 2D diffusion model to guide the optimization of a 3D model. 
SDS was later used in follow up text-to-3D works such as Prolific-Dreamer~\cite{wang2023prolificdreamer}, Score Jacobian Chaining~\cite{wang2023score}, DreamGaussian~\cite{tang2023dreamgaussian} and Magic3D~\cite{lin2023magic3d}, as well as image-to-3D techniques such as RealFusion~\cite{melas2023realfusion} and Magic123~\cite{qian2023magic123}.
In addition, this generative power has also been leveraged for editing existing 3D objects. 
Vox-E~\cite{sella2023vox} and DreamEditor~\cite{zhuang2023dreameditor} have shown that it is possible to locally edit shapes using an SDS loss. LatentNeRF~\cite{metzer2023latent} and later Fantasia3D~\cite{chen2023fantasia3d} propose a conditional text-to-3D variant, which is also provided with an input 3D shape. 

However, these aforementioned works all require timely optimization for each individual sample, and hence they are challenging to apply in practical settings. 
Several methods have been proposed for texturing 3D meshes using image diffusion models while bypassing SDS~\cite{richardson2023texture,cao2023texfusion,chen2023text2tex}. These methods, however, cannot modify the object's geometry and operate on a texture map representation, and not on the 3D representations directly.

Methods performing text-guided shape manipulation without the use of pretrained text--image models are significantly less prevalent. Text2Shape~\cite{chen2019text2shape} introduce a dataset tying 15K shapes from ShapeNet~\cite{chang2015shapenet} with textual descriptions, utilized for text-to-3D generation and also later for manipulation~\cite{liu2022towards}. ChangeIt3D~\cite{achlioptas2022changeit3d} introduce the ShapeTalk dataset, containing textual descriptions discriminating pairs of 3D shapes (also originating from ShapeNet), allowing for manipulating input shapes. LADIS~\cite{huang2022ladis} propose a disentangled latent representation which better localizes the 3D edits. We demonstrate that our technique allows for outperforming these prior 3D manipulation works, while enabling additional applications which are not necessarily restricted to specific domains.

\begin{figure*} %
\centering
\jsubfig{\includegraphics[width=0.9\textwidth]{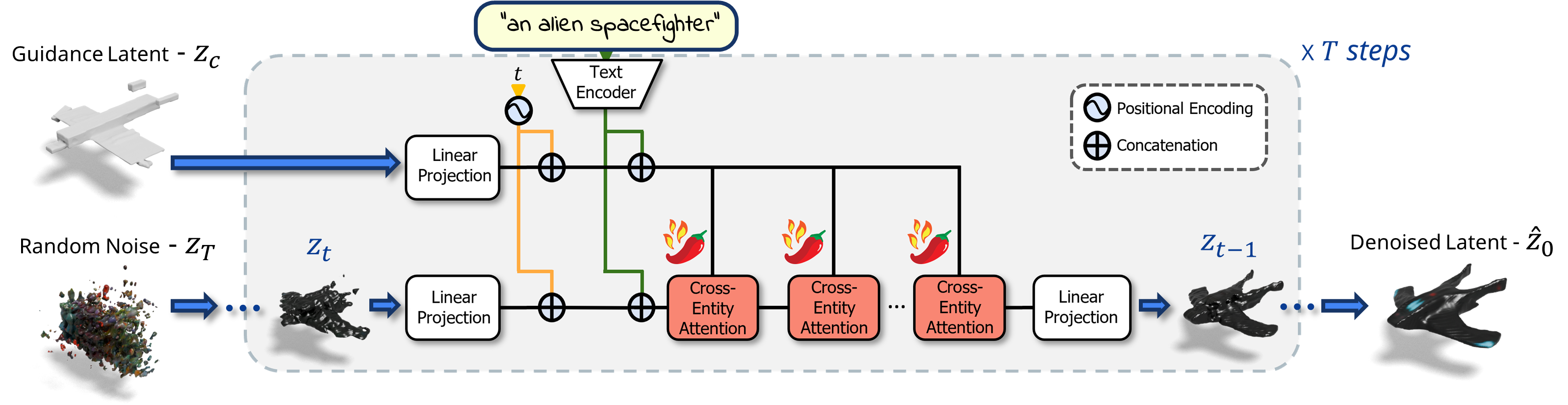}}{}
\vspace{-8pt}
\caption{\textbf{Finetuning 3D diffusion models with \methodName}. We finetune a transformer-based diffusion model~\cite{jun2023shap}, pretrained on a large dataset of text-conditional 3D assets, to enable structural control over the generated 3D shapes. The diffusion model (in gray) is modified to use latent vectors from multiple entities at each step $\mathbf{t}$ -- a conditional guidance shape $\mathbf{X}_c$ encoded into the guidance latent $\mathbf{Z}_c$ and a noisy input latent $\mathbf{Z}_t$ \ignorethis{an input 3D shape $\mathbf{X}_{in}$}. The self-attention layers are replaced with our proposed cross-entity attention mechanism. At inference time the fine-tuned diffusion model receives the guidance latent $\mathbf{Z}_c$, random gaussian noise $\mathbf{Z}_T$ and a guidance text as input and over $T$ steps gradually denoises the input to produce an output latent $\mathbf{\hat{Z}}_0$. The output latent can be decoded into the output shape $\mathbf{X}_{out}$, represented as either a neural radiance field or a signed texture field. \ignorethis{After finetuning, the output latent representation can be decoded into a 3D shape $\mathbf{X}_{out}$, represented as either a neural radiance field or a signed texture field.}}
\Description[Finetuning 3D diffusion models with \methodName]{We finetune a transformer-based diffusion model~\cite{jun2023shap}, pretrained on a large dataset of text-conditional 3D assets, to enable structural control over the generated 3D shapes. The diffusion model (in gray) is modified to use latent vectors from multiple entities at each step $\mathbf{t}$ -- a conditional guidance shape $\mathbf{X}_c$ encoded into the guidance latent $\mathbf{Z}_c$ and a noisy input latent $\mathbf{Z}_t$ \ignorethis{an input 3D shape $\mathbf{X}_{in}$}. The self-attention layers are replaced with our proposed cross-entity attention mechanism. At inference time the fine-tuned diffusion model receives the guidance latent $\mathbf{Z}_c$, random gaussian noise $\mathbf{Z}_T$ and a guidance text as input and over $T$ steps gradually denoises the input to produce an output latent $\mathbf{\hat{Z}}_0$. The output latent can be decoded into the output shape $\mathbf{X}_{out}$, represented as either a neural radiance field or a signed texture field.}
\label{fig:architecture_general}
\end{figure*}


\subsection{Controllable Shape Representations}
The problem of creating editable 3D representations has been extensively studied in recent years, not only in the context of text-guided techniques. DualSDF~\cite{hao2020dualsdf} represent shapes using two granularity levels, allowing to manipulate high resolution shapes through proxy primitive-based representations. Other works have shown that such primitive-based decompositions can also facilitate tasks such as shape completion~\cite{ganapathi2018parsing,sung2015data}. More recently, Tertikas \emph{et al.}~\shortcite{tertikas2023generating} proposed PartNeRF which generates shapes that are an assembly of distinct parts, each parameterized with a neural radiance field. 
KeypointDeformer~\cite{jakab2021keypointdeformer} discover 3D keypoints, rather than shape primitives, which can be edited for deforming 3D shapes. 
Several works couple implicit 3D representations with 2D modalities, allowing for editing the 3D shapes from 2D inputs~\cite{cheng2022cross,zheng2023locally}. 
DIF~\cite{deng2021deformed} represents shapes using a template implicit field shared across a shape category and a 3D deformation field per shape. EXIM ~\cite{liu2023exim} introduces a hybrid representation composed of an explicit part that enables coarse localization and an implicit part that enables fine global geometric editing and color modifications. SPAGHETTI~\cite{hertz2022spaghetti} propose a shape representation composed of Gaussian Mixture Models which allows for achieving part-level control. SALAD~\cite{koo2023salad} later extend this framework to incorporate a diffusion neural network using a cascaded framework. 

Several works edit shapes represented as neural fields by propagating edits from selected 2D projections~\cite{liu2021editing,yang2022neumesh}. Neutex~\cite{xiang2021neutex} represent appearance using 2D texture maps, allowing for editing textures using 2D techniques. Prior works have also shown that implicit neural fields can be coupled with an explicit mesh representation for editing them using as-rigid-as-possible deformations~\cite{garbin2022voltemorph,yuan2022nerf,xu2022deforming}. Neural Shape Deformation Priors~\cite{tang2022neural} predict a neural deformation field given a source mesh and target location of defined handles.  

In this work, we propose to manipulate shapes via text-guidance in addition to various structural priors, offering a flexible interface that can operate in various settings. Our approach bears some similarity to SDFusion~\cite{cheng2023sdfusion}, which enables conditional generation with multiple modalities including text. However, unlike SDFusion which requires training from scratch for each application, our work leverages pretrained text--3D diffusion models, allowing for a quick finetuning of these models without necessarily having access to the data or a vast number of high-end GPUs.


\subsection{Conditional Generation with Diffusion Models}
Many works are recently seeking new avenues for gaining control over the outputs generated by text-to-image diffusion models~\cite{hertz2022prompt,tumanyan2023plug,patashnik2023localizing,cao2023masactrl,wu2023tune,geyer2023tokenflow}. ControlNet~\cite{zhang2023adding} adds conditional control to 2D diffusion models, finetuning models to learn task-specific input conditions. They demonstrate image generation results using various conditions, including Canny edges and user scribbles. Our work is conceptually similar -- we modify 3D diffusion models to learn task-specific structural priors. 

To achieve structural control over the generation, we manipulate the internal representations of the denoising networks. Prior work have shown that manipulation of these representations, notably the cross-attention and self-attention layers, allows for effective editing of images and videos~\cite{ruiz2023dreambooth,chefer2023attend,geyer2023tokenflow}. In particular, several works recently demonstrate that Query features roughly control the structure of the generated images~\cite{cao2023masactrl,wu2023tune}. Cao \etal~\shortcite{cao2023masactrl} have demonstrated that Query features in the self-attention layers play a pivotal role in modifying the structure of the generated image, showing that non-rigid manipulations can be obtained by querying fixed Keys and Values. 
Similarly, Wu \etal~\shortcite{wu2023tune} keep $f_K$ and $f_V$ frozen while finetuning spatio-temporal attention blocks for creating temporally-consistent videos. Inspired by these 2D techniques, our approach carefully mixes Query features belonging to different 3D shapes to learn task-specific structural priors in 3D diffusion models, which are composed of self-attention layers, unlike 2D diffusion models that also contain cross-attention layers.

\section{Method}
\label{sec:method}
In this section, we introduce \methodName, an approach for incorporating structural priors in pretrained 3D diffusion models. We first review concepts related to the self-attention layers within a transformer-based diffusion model (Section \ref{sec:prel}). We then introduce Cross-Entity Attention, the core component of our approach (Section \ref{sec:arch}). Finally, we describe how to apply it in a transformer-based 3D diffusion model (Section \ref{sec:shap-e}, Figure \ref{fig:architecture_general}). 

\subsection{Preliminaries} 
\label{sec:prel}
We begin by describing the self-attention layers that compose the network blocks within a transformer-based diffusion model. 
At each timestep $t$, the noised latent code $\mathbf{z}_t$ is passed as input to the denoising network. 
For each self-attention layer $l$, the intermediate features of the network, denoted by $\phi_l(\mathbf{z}_t)$, are first projected to Keys ($K$), Queries ($Q$), and Values ($V$) using learned linear layers $f_Q, f_K, f_V$. Explicitly stated, $K=f_K(\phi_l(\mathbf{z}_t))$, $Q=f_Q(\phi_l(\mathbf{z}_t))$ and $V=f_V(\phi_l(\mathbf{z}_t))$.

The similarity between the Keys and Queries is initially computed, and then multiplied by the Values. Specifically, the pairwise dot product $Q\cdot K^T$ measures how relevant each key is to the corresponding query. This is then scaled by the square root of the key dimension $d$, normalised through a softmax function to obtain a unit vector and finally aggregated to produce the attention function: 
%
\begin{equation}
Attn(Q,K,V) =\text{softmax} \left ( \frac{Q\cdot K^T}{\sqrt{d}} \right )  V,
\end{equation}
which is a weighted sum of $V$, with higher weights for values whose corresponding keys have a larger dot product with the query. 

\subsection{Cross-Entity Attention}
\label{sec:arch}

\begin{figure} %
\centering
\jsubfig{\includegraphics[width=\columnwidth]{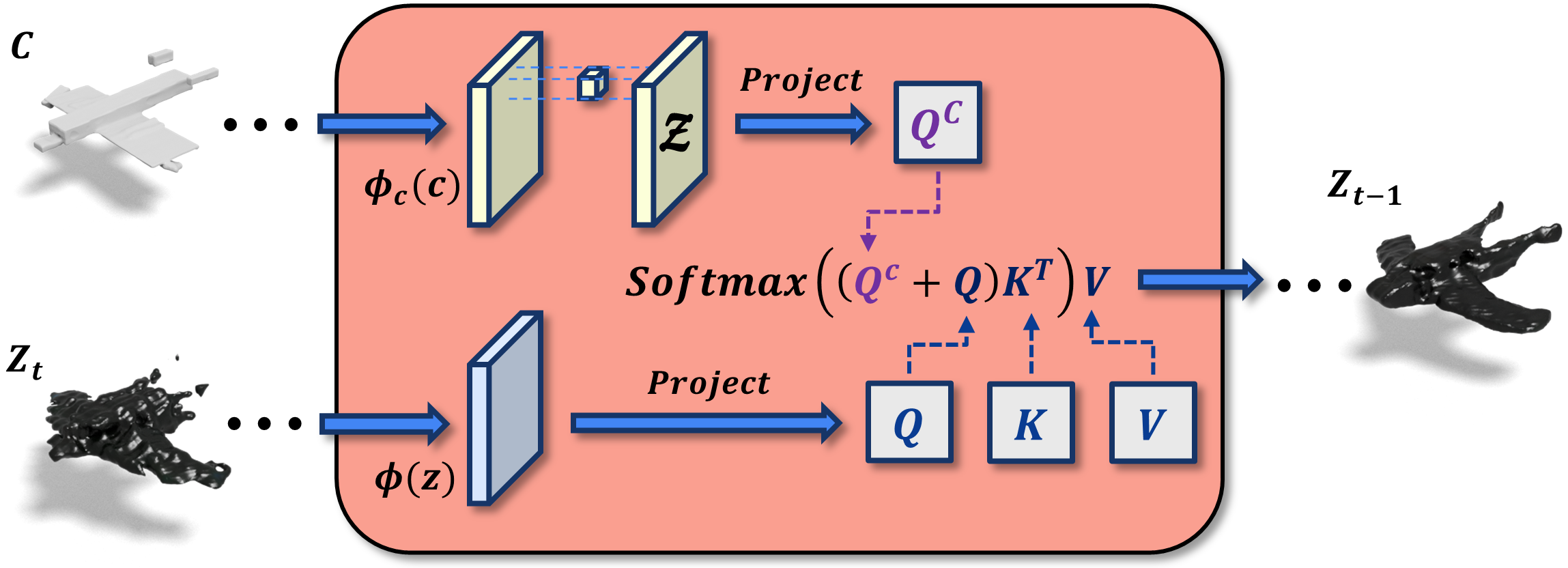}}{}\hfill
\vspace{-8pt}
\caption{\textbf{Cross-Entity Attention}. Given a pretrained self-attention block, we add a conditional latent $c$ originating from a different entity (\emph{i.e.} 3D shape). Our proposed mechanism mixes the Queries features (after a zero-convolution operator $\mathcal{Z}$ is applied to $c$), allowing for incorporating structural priors from $c$.  
}
\Description[Cross-Entity Attention]{Given a pretrained self-attention block, we add a conditional latent $c$ originating from a different entity (\emph{i.e.} 3D shape). Our proposed mechanism mixes the Queries features (after a zero-convolution operator $\mathcal{Z}$ is applied to $c$), allowing for incorporating structural priors from $c$.  }
\label{fig:architecture_detailed}
\end{figure}

%
%
Next we introduce the \emph{Cross-Entity Attention} mechanism, our core technical contribution, illustrated in Figure \ref{fig:architecture_detailed}. This mechanism modifies self-attention layers located within transformer-based diffusion models, allowing for latent vectors originating from multiple entities (\emph{i.e.} 3D shapes) to interact.
The input to our Cross-Entity Attention block is a pair of latent vectors $(\mathbf{z},\mathbf{c})$, where $\mathbf{z}$ denotes a noised latent code and $\mathbf{c}$ denotes a conditional latent that encodes structural information we would like to add to the original network. 
In our setting, the original network is a transformer based diffusion model, pretrained on millions of 3D assets.

As we are interested in preserving the capabilities of the original network, we first apply the \emph{zero-convolution} operator $\mathcal{Z}$ to $\mathbf{c}$. This is a 1 × 1 convolution layer with both weight and bias initialized to zeros, which was recently proposed for adding control to pretrained image diffusion models in ControlNet~\cite{zhang2023adding}. Due to its zero initialization, it ensures that the network will not be effected by the conditional latent code when training (or finetuning) begins.

We define the cross-entity attention mechanism over the Queries of the latent vectors, as we are interested in manipulating the structure of the shape encoded within $\mathbf{z}$, while preserving its visual appearance. Formally, the noised latent code $\mathbf{z}$ is projected to $K=f_K(\phi(\mathbf{z}))$, $Q=f_Q(\phi(\mathbf{z}))$ and $V=f_V(\phi(\mathbf{z}))$, denoting $\phi(\mathbf{z})$ as the network's intermediate features. We then perform: 
\begin{equation}
Q_{\times}=f_Q(\phi(\mathbf{z}))+f_{Q_c}(\mathcal{Z}(\phi_c(\mathbf{c}))),
\end{equation}
where $f_{Q_c}$ and $\phi_c(\mathbf{c})$ are a learned linear layer and intermediate features, initialized randomly. 

The output of our cross-entity attention block is the attention function computed over these updated Queries. That is, the output of our proposed block is $\mathbf{z}_{out} = Attn(Q_\times,K,V)$. We allow all block parameters to optimize freely during model finetuning. Intuitively, our attention mechanism acts as a fully-functional self-attention block when finetuning begins. As finetuning progresses, the network gradually learns how to utilize information from the guidance shape at each layer. Note that this is in contrast to a more simple cross-attention mechanism, such as that used in \cite{loizou2023cross}, which has no ability to retain a self-attention component.


\subsection{Structural Priors in 3D Diffusion Models} 
\label{sec:shap-e}  
In this section, we describe how our cross-entity attention mechanism can be integrated into transformer-based 3D diffusion models to enable structural control over the generated outputs. We use the recently proposed \shapE{}~\cite{jun2023shap} as a reference 3D diffusion model. \shapE{} was trained on several million 3D assets, and is capable of generating diverse high-quality 3D objects conditioned on text prompts. For completeness, we briefly describe its architecture, which we modify for creating \methodName.

Shap$\cdot$E maps a 3D shape $\mathbf{X}$ to a latent representation $\mathbf{z}\in \mathbb{R}^{d\times d}$ via an encoder $E$. Specifically, we have $\mathbf{z} = E(\mathbf{X})$, with a latent dimension $d=1024$. The input $\mathbf{X}$ is composed of both RGB point clouds and RGBA rendered images. The latent $\mathbf{z}$ can be linearly projected into the weights of either a NeRF or a signed texture field (STF) representation via a decoder $D$. Note that a STF, which is essentially a signed distance field that also provides appearance information, can be represented as a colored mesh, as further detailed in prior work ~\cite{shen2021deep,gao2022get3d}.
For text-conditional generation, this latent representation, together with pre-pended tokens representing the CLIP text embedding and the timestep embedding, is fed to a transformer-based diffusion model.   
%
The diffusion model is trained following the setup of Ho \etal~\shortcite{ho2020denoising}, directly minimizing the error between the original and predicted (de-noised) latent code.

To generate shapes conditioned on structural priors (in addition to text prompts), we modify the system's input to also use a conditional guidance 3D shape $\mathbf{X}_{c}$. 
We freeze the encoder $E$, and fine-tune the pre-trained 3D diffusion generative model (modified as detailed below) on datasets of inputs and guidance shapes that are encoded with $E$. Each self-attention block is replaced with a cross-entity attention block. To avoid overfitting, we use constant intermediate features $\phi _c$ for each block, unlike $\phi_l (z)$ which are layer dependent. 

During training, given an input latent representation $\mathbf{z}_{0}$ corresponding to an input 3D asset $\mathbf{X}_{in}$ (\emph{i.e.} $\mathbf{z}_0 = E(\mathbf{X}_{in})$), noise is progressively added to it, producing a noisy latent $\mathbf{z}_{t}$, where $t$ represents the number of timestamps noise is added. 
Given $\mathbf{z}_{0}$, a time step $t$, a text prompt $c_{text}$ and a latent representation $\mathbf{z}_{c}$ corresponding to the 3D conditional guidance shape $\mathbf{X}_{c}$, 
our model $\mathcal{M_{\theta}}$ learns to directly predict the denoised input latent representation $\mathbf{z}_{0}$ by minimizing the same objective used in \shapE{}:
\begin{equation} 
    \mathcal{L} = \mathcal{E}_{\mathbf{z}_{0},t,c_{text},c_{0}}||\mathcal{M_{\theta}}(\mathbf{z}_{t},t,c_{text},\mathbf{z}_{c}) - \mathbf{z}_{0}||_{2}^{2}
\end{equation}
An overview of our training process is shown in Figure \ref{fig:architecture_general}. 

During inference our system is only provided with the guidance shape $\mathbf{X}_{c}$ encoded into the latent $\mathbf{Z}_{c}$ with $E$ and a text prompt ($c_{text}$). \ignorethis{We encode $\mathbf{X}_{c}$ with $E$,} We sample from $\mathcal{M_{\theta}}$, starting at a random noise sample $\mathbf{z}_{T}$. This sample is gradually denoised into $\mathbf{\hat{z}}_{0}$, which is then decoded \ignorethis{then decode $\mathbf{\hat{z}}_{0}$} into our 3D output $\mathbf{X}_{out}$, represented as either a NeRF or a STF, using $D$. 
\\

\noindent \textbf{Optional Refinement}.
Our outputs can be refined using an auxiliary unsupervised iterative process that uses 2D diffusion models. Specifically, we can replace the \shapE{} initialization in GaussianDreamer \cite{yi2023gaussiandreamer} with \methodName{}. GaussianDreamer then proceeds to optimize the Gaussians initialized according to our outputs using Score Distillation, producing more detailed Gaussian splats at the expense of time, specifically increasing generation time from roughly 20 seconds to 15 minutes. See Figures \ref{fig:teaser} and \ref{fig:comparisons_abstraction} for results before and after this optional refinement stage. Note that all other reported results are provided without refinement.

\section{Tasks}
\label{sec:apps}
We demonstrate the utility of \methodName using three text-conditioned 3D-to-3D tasks: semantic shape editing (Section \ref{sec:shapetalk}), text-conditional abstraction-to-3D (Section \ref{sec:abstract}), and 3D stylization (Section \ref{sec:style}). 

For each task, we construct a dataset of latent representations and target text-prompts and fine-tune the pretrained 3D diffusion model following the procedure described in the previous section. In other words, we encode a set of input and conditional shapes $\{\mathbf{X}_{in}, \mathbf{X}_{c}\}$ via $E$ to obtain a set of latent representations $\{\mathbf{z}_{0}, \mathbf{z}_{c}\}$ which are used together with their corresponding target text prompts $\{c_{text}\}$ for finetuning. 
Below, we describe the tasks and provide experimental details, as well as discuss alternative methods and evaluation metrics. Additional details and comparisons, including perceptual studies, are provided in the supplementary material.

\subsection{Semantic Shape Editing} 
\label{sec:shapetalk}

\textbf{Task description} Several works have recently explored the problem of performing semantic fine-grained edits of shapes using language~\shortcite{achlioptas2022changeit3d,huang2022ladis}. For this task, the target text prompt describes desired semantic modifications to be performed over the input shape. For example, given an input chair, target texts include ``the legs are thinner" or ``there is a hole in the back".

\smallskip \noindent
\textbf{Experimental details}. For this task, we use the ShapeTalk dataset proposed by Achlioptas \etal~\shortcite{achlioptas2022changeit3d}.
This dataset contains pairs of \emph{distractor} and \emph{target} models (originating from ShapeNet) annotated with a textual annotation describing the shape differences from the distractor shape to the target one. For finetuning models on this task, we use distractor models as conditional guidance shapes $\mathbf{X}_c$ and target models as the input ones $\mathbf{X}_{in}$. We randomly replace $50\%$ of the distractor models with the target ones to further enforce structural similarity to the target models.
During inference, only the distractor model and the associated textual description are fed to \methodName.
We follow their setup, finetuning models for the \emph{Table}, \emph{Lamps}, and \emph{Chair} categories and using their train/set splits. 
We perform additional filtering to these sets to ensure that the distractor and target models are sufficiently close, as we observe that many pairs are geometrically very different. This yields datasets containing approximately $15\%$ of the shapes from the original ShapeTalk dataset (\emph{i.e.} 8K pairs on average for training). See the supplementary material for details.

\smallskip \noindent
\textbf{Alternative Methods}. We compare against ChangeIt3D~\shortcite{achlioptas2022changeit3d}, which operates over point cloud representations. We use their outputs directly, as these are publicly available. In the supplementary material, we also perform a qualitative comparison with LADIS~\shortcite{huang2022ladis} over results reported in their paper (as source code or trained models are not available we cannot conduct a quantitative evaluation). 

\smallskip \noindent
\textbf{Evaluation metrics}. We follow the evaluation protocol proposed by Achlioptas \etal~\shortcite{achlioptas2022changeit3d}. Specifically, we use the following metrics:

\smallskip \noindent \emph{Linguistic Association Boost} (LAB) uses their pretrained listener model for measuring the difference in the predicted association score between the input--output shapes and the target text prompt.  

\smallskip \noindent \emph{Geometric Difference} (GD) uses a standard Chamfer distance to measure the geometric difference between the input and output shapes (scaled by $10^{-2}$ in comparison to the distances reported in \cite{achlioptas2022changeit3d}), evaluating shape identity preservation.

\smallskip \noindent \emph{localized-Geometric Difference} ($l$-GD) uses a part-based segmentation model to only measure geometric differences in regions unrelated to the edit text.

\smallskip \noindent \emph{Class Distortion} (CD) uses their pretrained shape classifier for measuring the absolute difference of the shape category probability, comparing the input and output shapes. 

\subsection{Text-conditional Abstraction-to-3D} 
\label{sec:abstract}

\textbf{Task description}. Primitive-based surface reconstruction is a longstanding problem in computer vision and graphics~\cite{gal2007surface,schnabel2009completion,hao2020dualsdf}. We explore this problem in the context of our framework. Specifically, given a proxy primitive-based abstract representation and a target text prompt, we are interested in generating a corresponding high-resolution 3D shape that conforms to the target text prompt while maintaining fidelity to the input abstract shape. 

\smallskip \noindent 
\textbf{Experimental details}. We use 3D models from ShapeNet~\cite{chang2015shapenet} annotated with textual descriptions for this task. Several methods provide means of abstracting shapes of a given category into an assembly of cuboid primitives \cite{tulsiani2017learning,sun2019learning,yang2021unsupervised}. Therefore, to create corresponding primitive-based shape representations, we utilize the trained \emph{Airplane}, \emph{Chair} and \emph{Table} models given by Yang and Chen~\shortcite{yang2021unsupervised}. 
We also use their splits for constructing train/test datasets. 

\smallskip \noindent 
\textbf{Alternative Methods}. We compare against SketchShape, the variant from LatentNerf~\cite{metzer2023latent} conditioned on coarse shapes, and Fantasia3D~\cite{chen2023fantasia3d} which can optionally use a guidance  shape. 
Note that both of these methods are optimization-based, and therefore, are significantly slower at inference time.

\smallskip \noindent 
\textbf{Evaluation metrics}. We measure geometric differences (using the GD metric discussed in Section \ref{sec:shapetalk}) between the input primitive-based proxy shape and the output shape to evaluate how well the model enforces the structural priors from the guidance abstract shape. 
Furthermore, we evaluate to what extent our results are faithful to the edit prompt using the following  metrics:

\smallskip \noindent \emph{CLIP Similarity} ($\text{CLIP}_{Sim}$) measures the similarity between the output objects and the target text prompts, using the cosine-distance between their CLIP embedding. 

\smallskip \noindent \emph{CLIP Direction Similarity} ($\text{CLIP}_{Dir}$), first introduced for evaluating image edits in StyleGAN-NADA~\cite{gal2021stylegan}, measures the cosine distance between the direction of the change from the input and output rendered images and the direction of the change from an input prompt to the edit prompt. To evaluate these CLIP based metrics, we render 20 images of both the output and guidance shapes from uniformly-distributed azimuth angles around the 3D object, and average over these angles.


\subsection{3D Stylization} 
\label{sec:style}

\textbf{Task description}. 
This task aims at performing text-driven editing of an uncolored 3D asset. 
Following Michel \etal~\shortcite{michel2022text2mesh}, we define \textit{style} as the object's texturing and fine-grained geometric details.

\smallskip \noindent \textbf{Experimental details}.  
To construct a dataset for this task, we utilize the large-scale Objaverse~\cite{deitke2023objaverse} dataset. Each model in Objaverse is accompanied by metadata, which includes fields such as name, description, categories, and tags.
For our purposes, we need text prompts that describe the object's style and overall appearance. 
We observed that using the available metadata directly (\emph{e.g.} selecting specific fields) yields highly noisy target prompts. Therefore, we finetune the InstructBLIP~\cite{instructblip} model to extract target prompts from the object's metadata and associated rendered imagery (see the supplementary for additional details); the model's outputs are used as the text prompts $c_{text}$ for learning 3D stylization, along with the encoded 3D assets $\mathbf{z}_0$ and the uncolored assets $\mathbf{z}_c$. We construct a training dataset containing roughly 7.5K items overall. 

\smallskip \noindent 
\textbf{Alternative Methods}. We compare against two gradient-based optimization techniques: Latent-Paint, the variant from Latent-Nerf~\cite{metzer2023latent} that operates over 3D meshes directly (only modifying the object's texture), and Fantasia3D~\cite{chen2023fantasia3d}. For this task, we compare against two variants of Fantasia3D: One that only performs appearance modeling (henceforth denoted as Fantasia-Paint) and the full model, which also modifies the object's geometry. In the supplementary material, we also compare against Vox-E~\cite{sella2023vox}, a recent optimization-based method proposed for performing text-guided editing of 3D objects.  

\smallskip \noindent 
\textbf{Evaluation metrics}. For this task, we use the same evaluation metrics discussed above in Section \ref{sec:abstract}: $\text{CLIP}_{Sim}$, $\text{CLIP}_{Dir}$ and GD, to evaluate both the fidelity to the edit and the guidance shape. 

\section{Experiments}
We present the results and comparisons for the tasks described above in Section \ref{sec:eval}. We then ablate the design choices for the cross-entity attention block in Section \ref{sec:ablations}. Finally, we discuss limitations in Section \ref{sec:limitations}. Additional results, comparisons and ablations can be found in the supplementary material.

\subsection{Evaluation}
\label{sec:eval}

\begin{table}[t]
\setlength{\tabcolsep}{5.5pt}
 \def\arraystretch{1.1}
\centering
\resizebox{0.99\linewidth}{!}{
\begin{tabular}{lcccc}
\toprule
    Method  & LAB$\uparrow$ & GD$\downarrow$ & $l$-GD $\downarrow$ & CD$\downarrow$  \\ 
    \midrule 
  ChangeIt3D~\cite{achlioptas2022changeit3d} &  0.27  & \textbf{0.003}  & \textbf{0.009} & \textbf{0.05}  \\ 
Ours  & \textbf{0.44} & 0.007 & 0.013 & \textbf{0.05}  \\ 
\bottomrule
\end{tabular}
}
\caption{\textbf{Semantic Shape Editing Evaluation.} Above we report performance over the ShapeTalk~\cite{achlioptas2022changeit3d} test set (averaging only over highly similar shapes, as discussed in Section \ref{sec:shapetalk}). As illustrated above, our method yields significantly higher LAB scores, suggesting edits that are semantically more accurate, at the expense of slightly higher geometric differences. 
}
\label{tab:editing_comparison}
\end{table}

\begin{figure} %
\centering
\jsubfig{\includegraphics[height=1.7cm, trim={6.9cm 8.5cm 7.7cm 8cm}, clip]{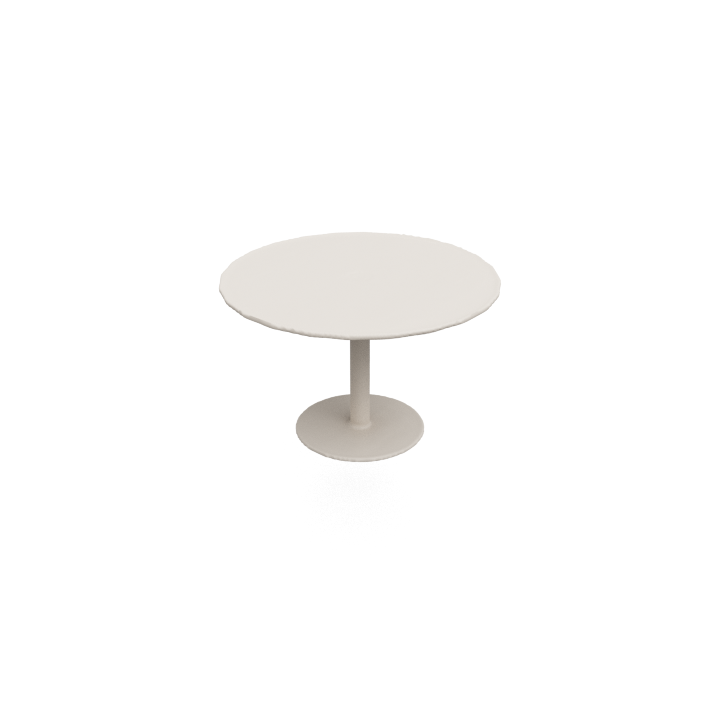}
\includegraphics[height=1.7cm, trim={6.9cm 8.9cm 7.7cm 8cm}, clip]{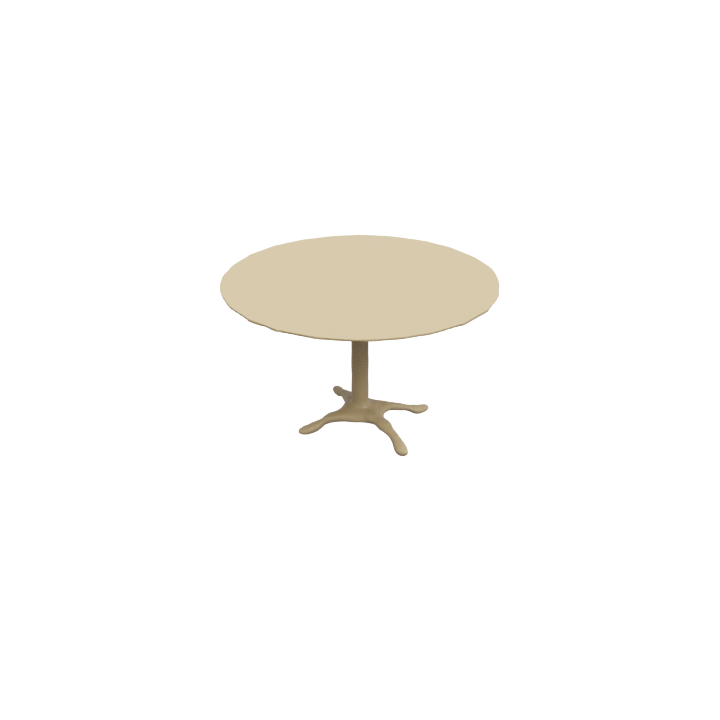}}{\footnotesize{\emph{It has four legs
}}}
\hfill 
\jsubfig{\includegraphics[height=1.7cm, trim={8cm 6cm 9cm 6cm}, clip] {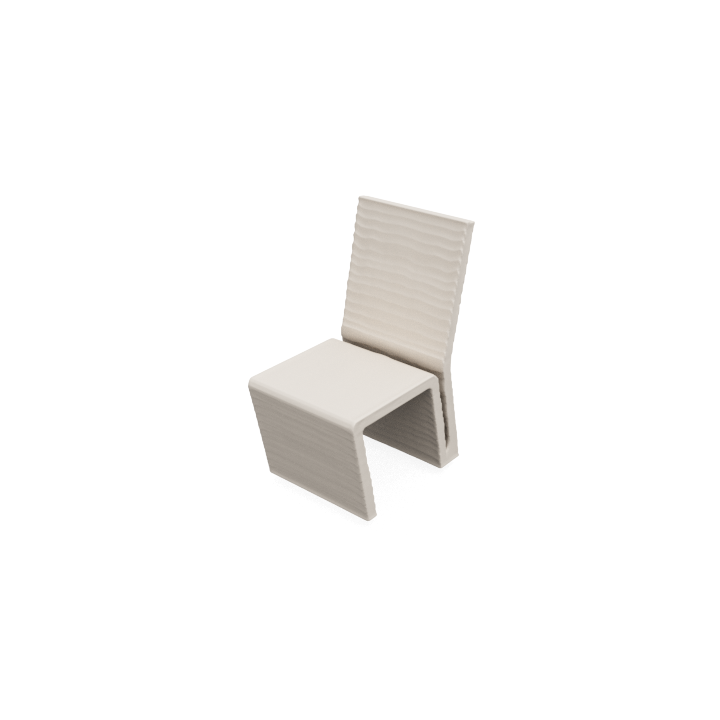}
\includegraphics[height=1.7cm, trim={8cm 6cm 9cm 6cm}, clip]{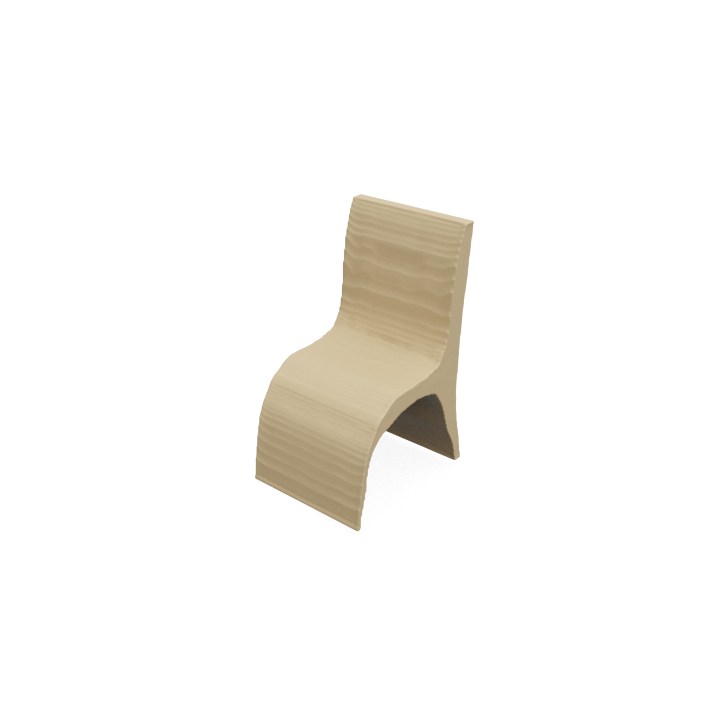}}{\footnotesize{\emph{The seat has a rounded edge}}}
\hfill
\jsubfig{\includegraphics[height=1.7cm, trim={11cm 9cm 11cm 7cm}, clip]{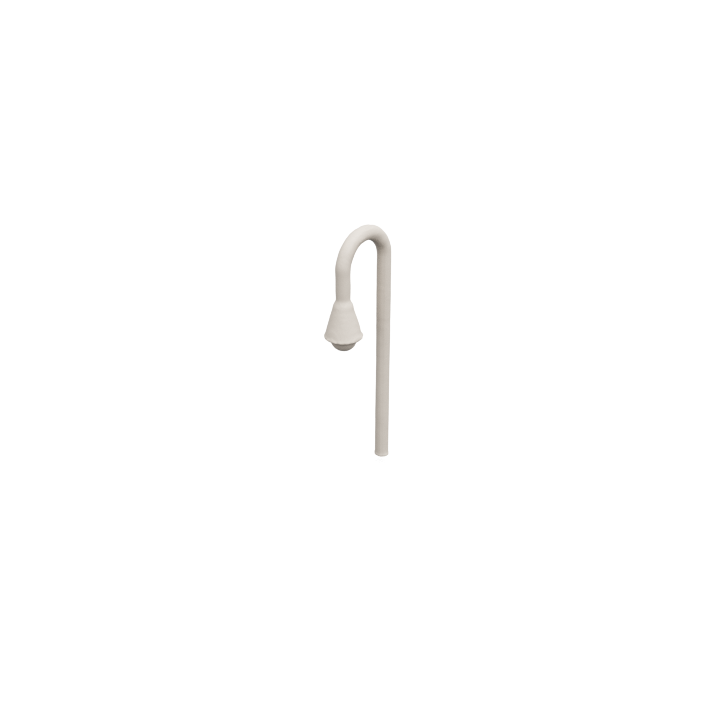}
\includegraphics[height=1.7cm, trim={11cm 9cm 11cm 7cm}, clip]{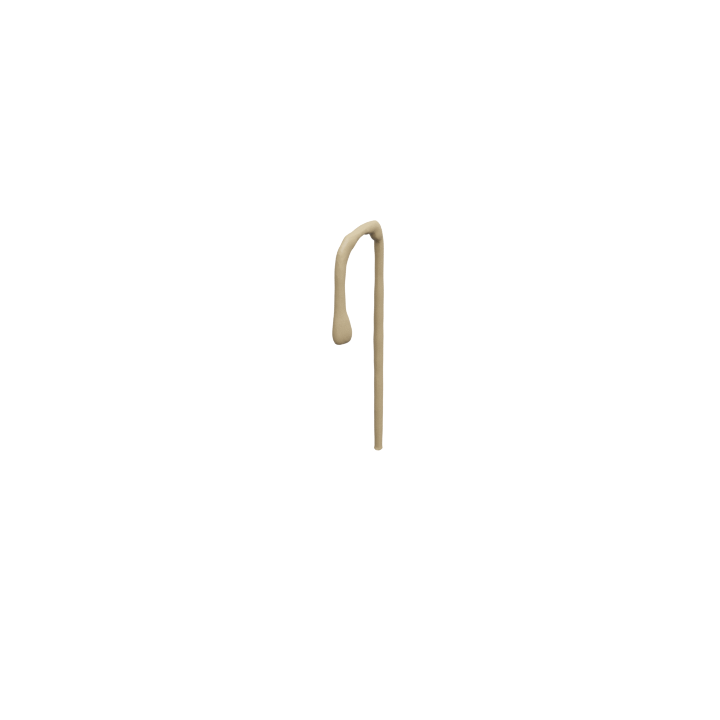}}{\footnotesize{\emph{It looks like a straw}}}
\vspace{-4pt}
\caption{Semantic shape editing results are shown above (input guidance shape on the left and edited outputs on the right, shown in different colors for visualization purposes). As illustrated in the figure, our method can semantically edit input shapes according to target prompts, while preserving the shape's structure. 
}
\label{fig:editing_res}
\end{figure}

\begin{figure} %
\centering 
\rotatebox{90}{ChangeIt3D}
\jsubfig{\includegraphics[height=1.8cm, trim={3.3cm 3.0cm 3.3cm 3.3cm}, clip]{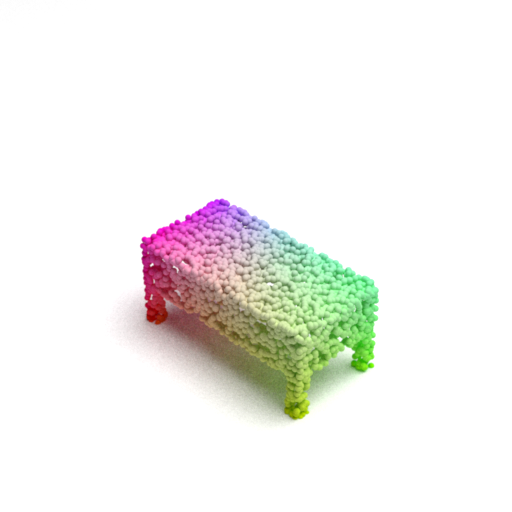}
\includegraphics[height=1.8cm, trim={3.3cm 3.0cm 3.3cm 3.3cm}, clip]{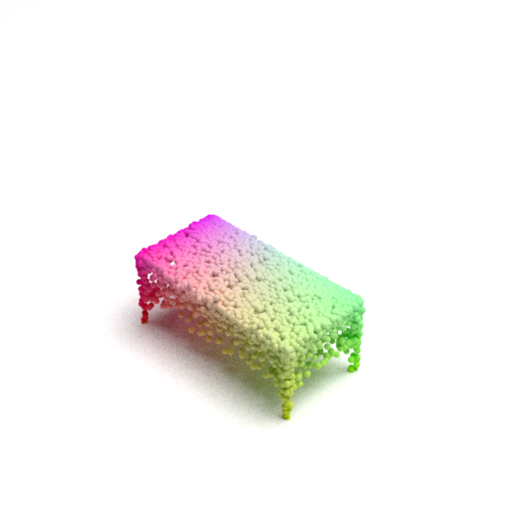}}{}
\hfill 
\jsubfig{\includegraphics[height=1.8cm, trim={1cm 2cm 2cm 2cm}, clip]{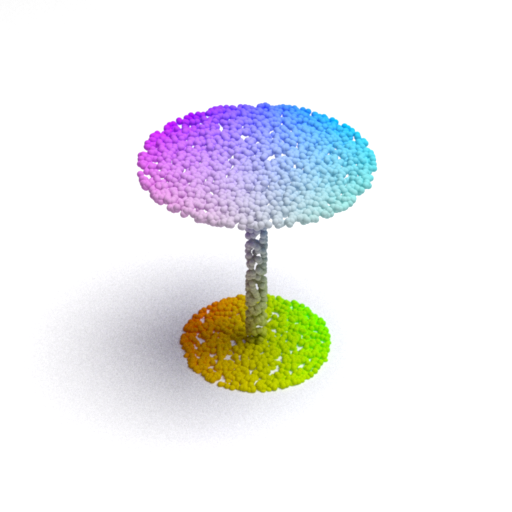}
\includegraphics[height=1.8cm, trim={1cm 2cm 2cm 2cm}, clip]{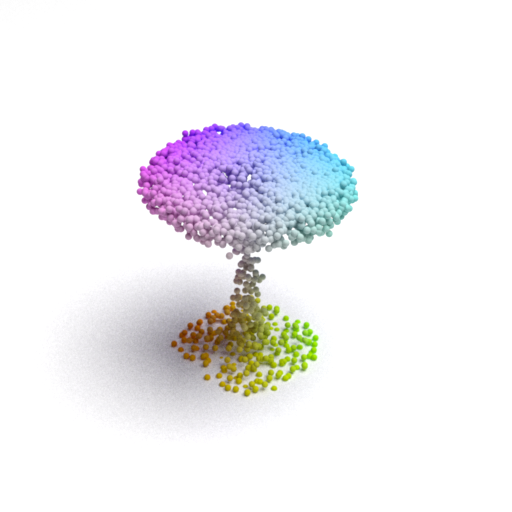}}{}
\\ 
\rotatebox{90}{\whitetxt{xxx}Ours}
\jsubfig{\includegraphics[height=1.8cm, trim={3.3cm 3.0cm 3.3cm 3.3cm}, clip]{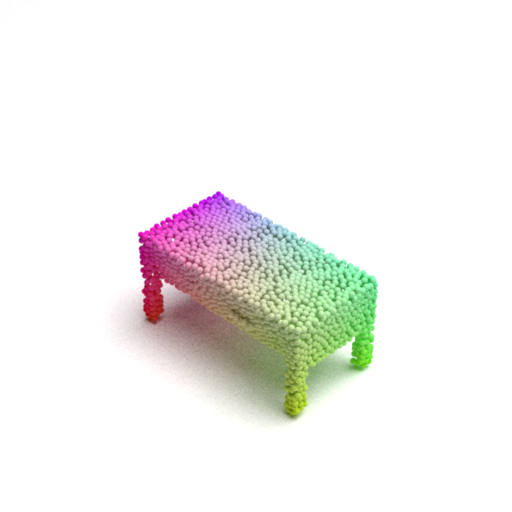}
\includegraphics[height=1.8cm, trim={3.3cm 3.0cm 3.3cm 3.3cm}, clip]{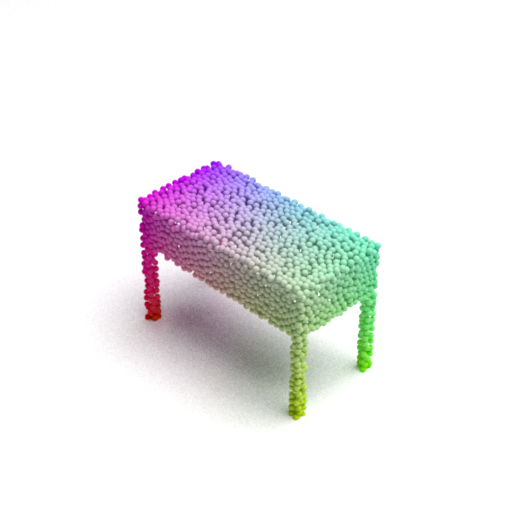}}{\footnotesize{\emph{Its legs are taller}}}
\hfill
\jsubfig{\includegraphics[height=1.8cm, trim={1cm 2cm 2cm 2cm}, clip]{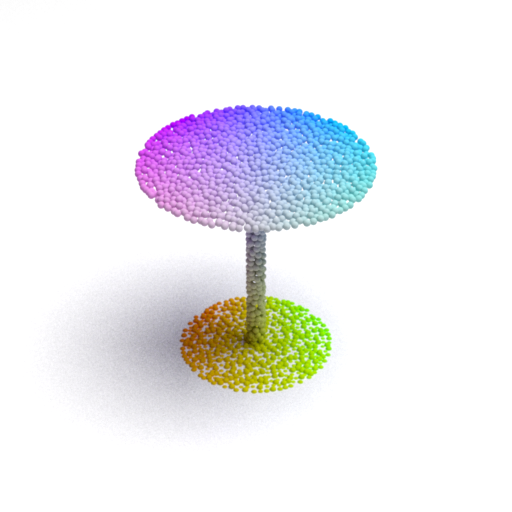}
\includegraphics[height=1.8cm, trim={1cm 2cm 2cm 2cm}, clip]{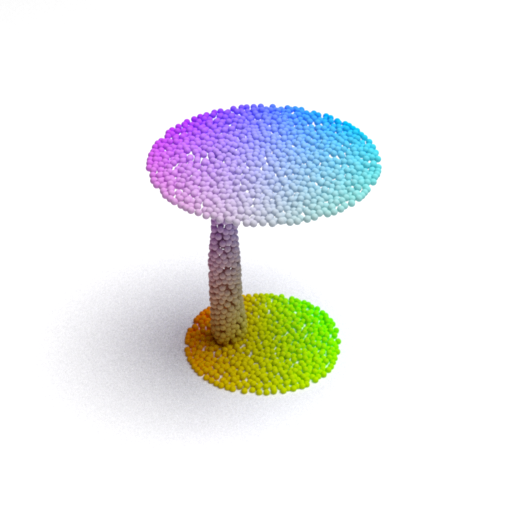}}{\footnotesize{\emph{Its top is not connected from its center to the leg}}}
\vspace{-4pt}
\caption{\textbf{Semantic Shape Editing Comparison}. We compare to prior work performing semantic shape editing above. As ChangeIt3D~\cite{achlioptas2022changeit3d} operates over a point cloud representation, we show input point clouds on the left and edited point clouds on the right. For our results, we visualize the point clouds after shape encoding, hence our inputs are not identical to theirs. 
As illustrated in the figure, our method can perform more significant edits, yielding edited shapes that better reflect the target prompts. 
}
\label{fig:editing_comp_pcs}
\end{figure}

\medskip \noindent \textbf{Semantic Shape Editing}. Results for the semantic shape editing task are reported in Table \ref{tab:editing_comparison}. As illustrated in the table, our edits better reflect the target text prompts, yielding an average LAB score of 0.44 versus 0.27 for ChangeIt3D. Both methods are capable of generating objects resembling their respective object categories, as illustrated by the low class distortion values. Our method yields slightly higher GD and $l$-GD scores. Generally, we observe that the outputs generated by ChangeIt3D often do not deviate significantly from the inputs (which is also consistent with the lower LAB scores). This is further illustrated in Figures \ref{fig:editing_comp_pcs} and \ref{fig:editing_res}. 

\begin{table}[t]
\setlength{\tabcolsep}{3.0pt}
 \def\arraystretch{1.1}
\centering
\resizebox{\linewidth}{!}{
\begin{tabular}{lcccc}
\toprule
   Method  & $\text{CLIP}_{Sim}\uparrow$ & $\text{CLIP}_{Dir} \uparrow$ & $\text{GD} \downarrow$ & Run Time \\ 
    \midrule 
    SketchShape & 0.27 & 0.01 & --- & $\sim \text{15 minutes}$\\
    Fantasia3D & 0.27 & 0.01 & 0.06 & $\sim \text{30 minutes}$  \\
    Ours & \textbf{0.28} & \textbf{0.03} & \textbf{0.01} & $\textbf{$\sim$ 20 seconds}$\\
\bottomrule
\end{tabular}
}
\caption{\textbf{Text-conditional Abstraction-to-3D Evaluation.} Above we compare the performance of SketchShape~\cite{metzer2023latent} and Fantasia3D~\cite{chen2023fantasia3d} against ours over the primitive-based shape conditioning task. As illustrated above, our method can more faithfully preserve the input structure, while exhibiting significantly faster inference time. GD is not computed for SketchShape as it outputs a NeRF representation. }
\label{tab:abstraction_comparison}
\end{table}

\begin{figure} %
\centering
\rotatebox{90}{\hspace{-4pt}\footnotesize{\emph{A SkyJet British}}}
\rotatebox{90}{\hspace{-4pt}\footnotesize{\emph{Aerospace BAe-146}}}
\jsubfig{\includegraphics[height=1.53cm, trim={7cm 7cm 7cm 7cm}, clip]{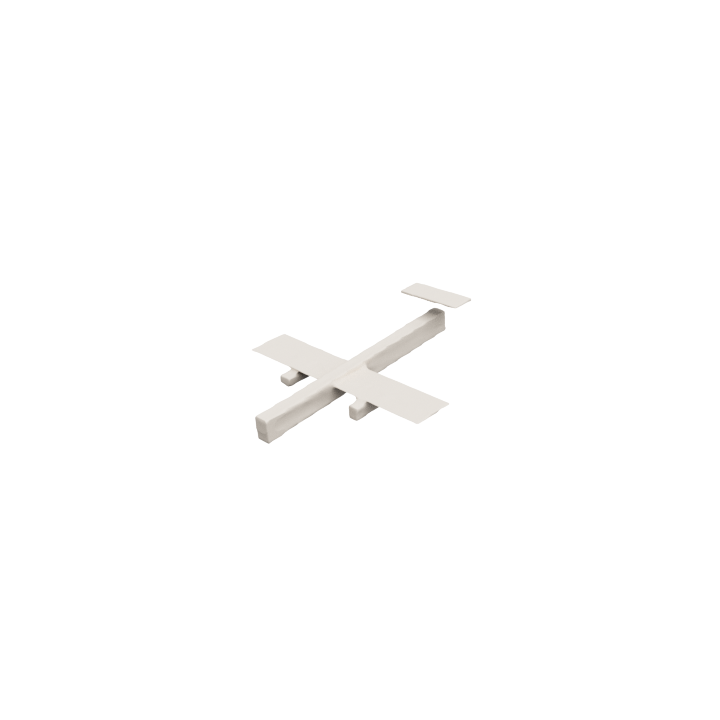}}{}\hfill
\jsubfig{\includegraphics[height=1.53cm, trim={6cm 6cm 6cm 6cm}, clip]{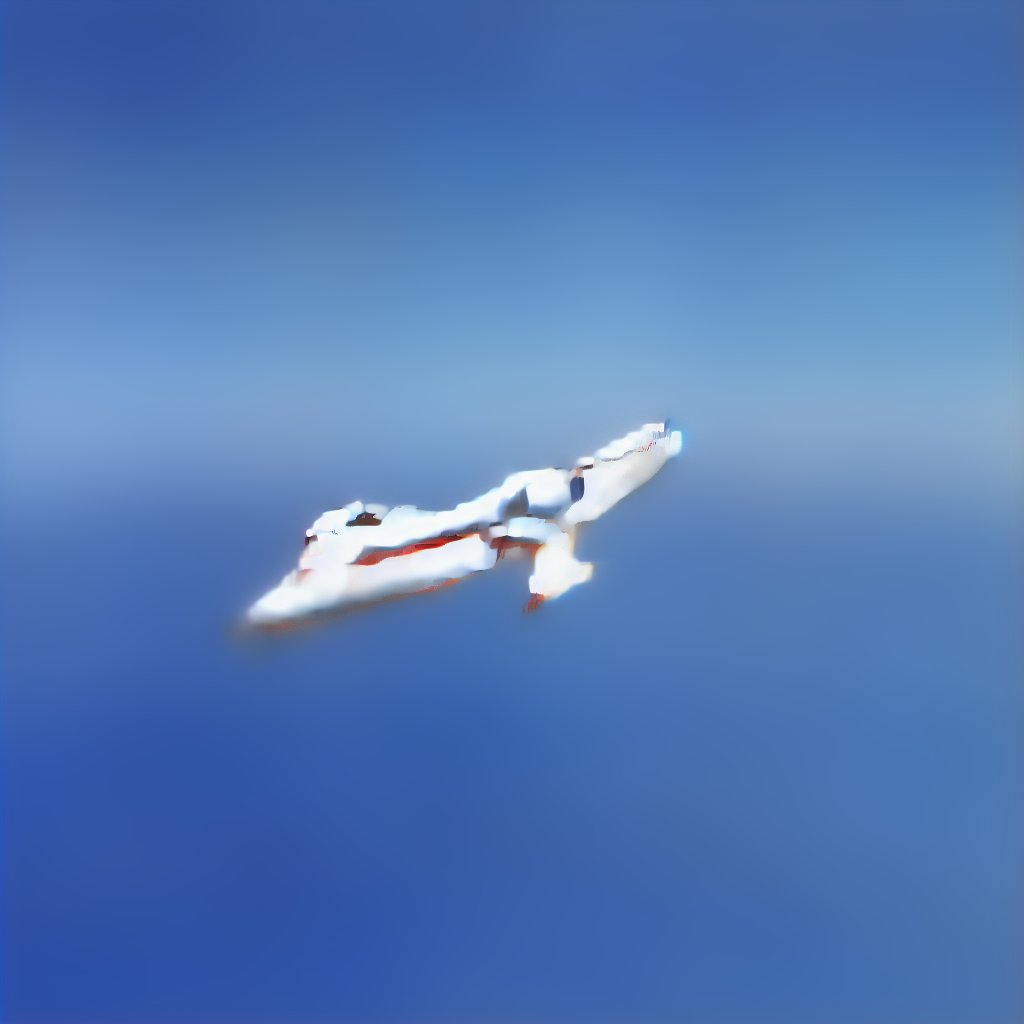}}{} \hfill
\jsubfig{\includegraphics[height=1.53cm,]{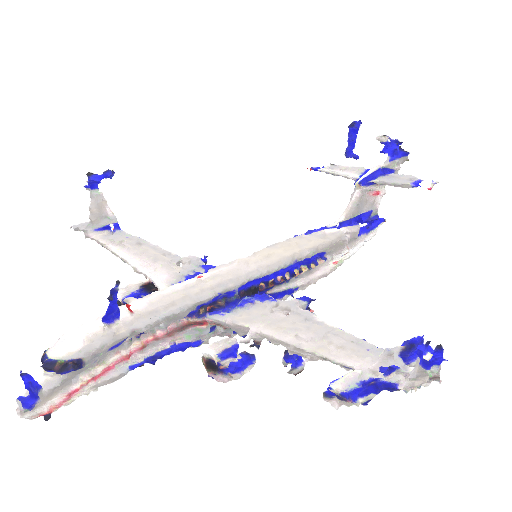}}{} \hfill
\jsubfig{\includegraphics[height=1.53cm, trim={1.1cm 1.1cm 1.1cm 1.1cm}, clip]{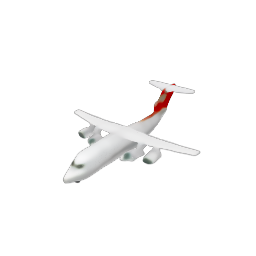}}{} \hfill
\jsubfig{\includegraphics[height=1.53cm, trim={5cm 5cm 5cm 5cm}, clip]{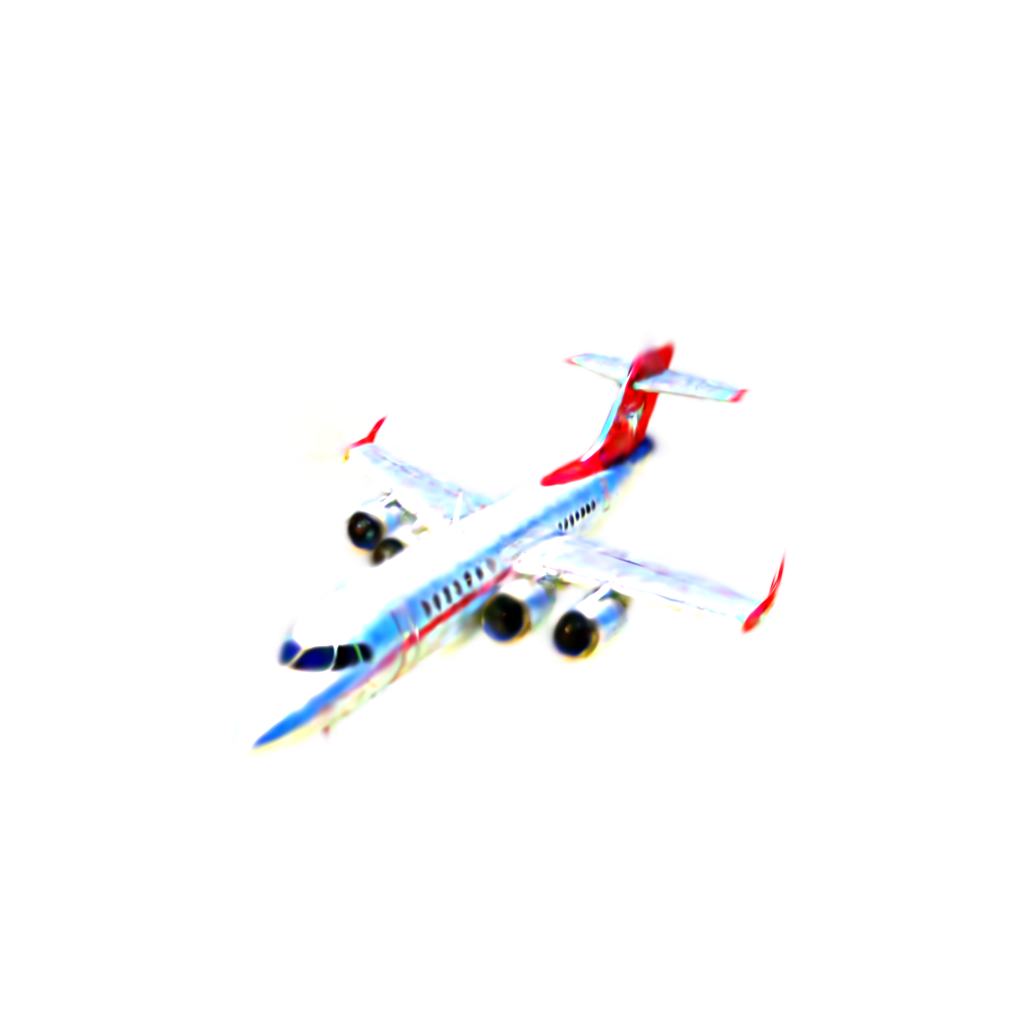}}{}
\vspace{1.5pt}
\\ \vspace{2pt}
\rotatebox{90}{\footnotesize{\emph{\whitetxt{xxxxxx}A }}}
\rotatebox{90}{\footnotesize{\emph{\whitetxt{xx}Warhawk}}}
\jsubfig{\includegraphics[height=1.53cm, trim={7cm 7cm 7cm 7cm}, clip]{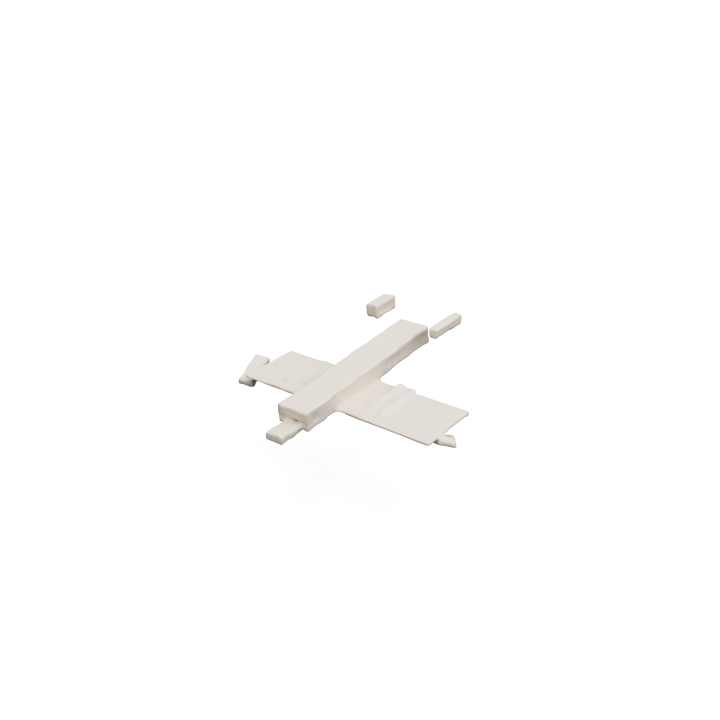}}{}\hfill
\jsubfig{\includegraphics[height=1.53cm, trim={6cm 6cm 6cm 6cm}, clip]{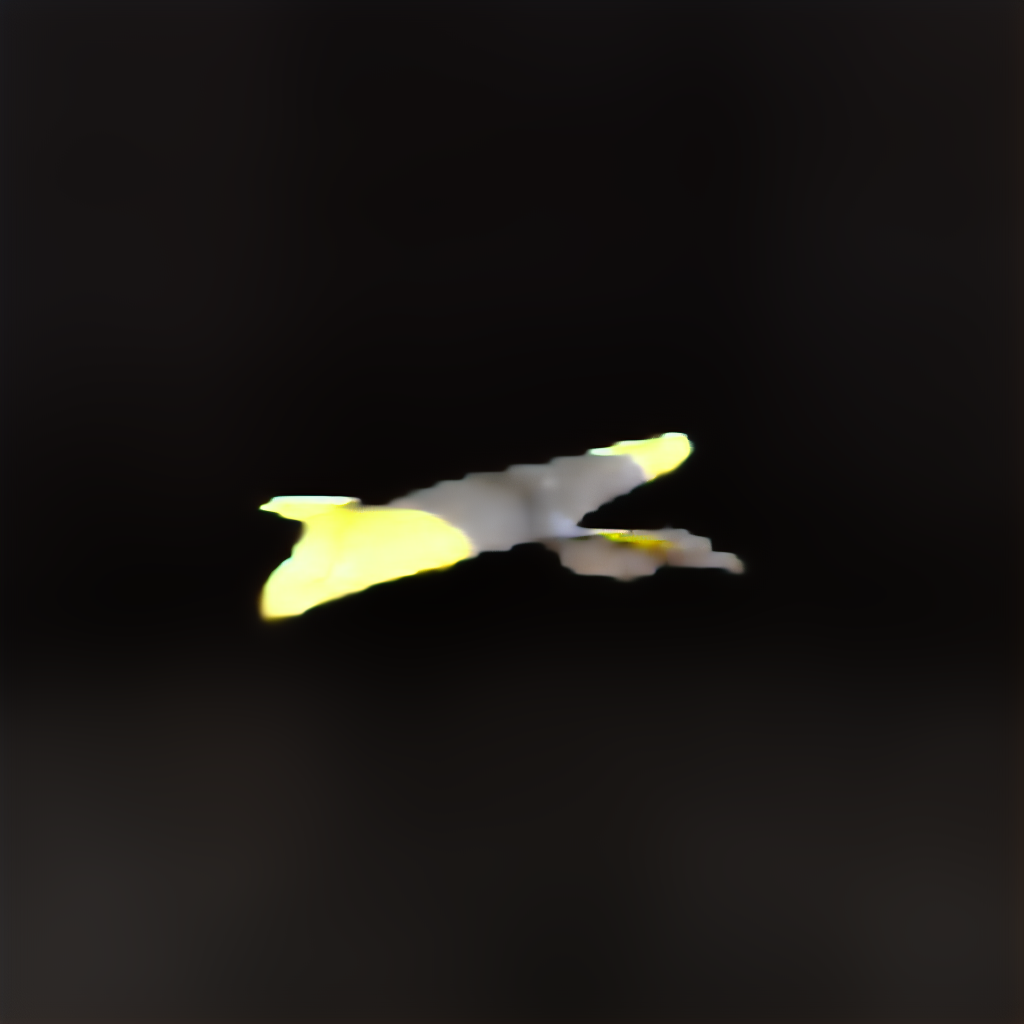}}{} \hfill
\jsubfig{\includegraphics[height=1.53cm]{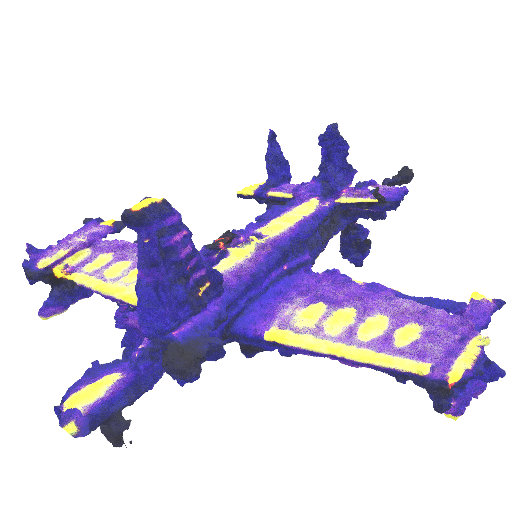}}{} \hfill
\jsubfig{\includegraphics[height=1.53cm, trim={1.1cm 1.1cm 1.1cm 1.1cm}, clip]{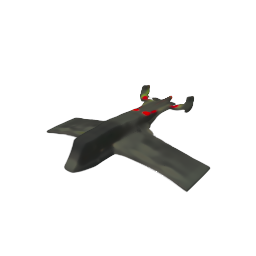}}{} \hfill
\jsubfig{\includegraphics[height=1.53cm, trim={5cm 5cm 5cm 5cm}, clip]{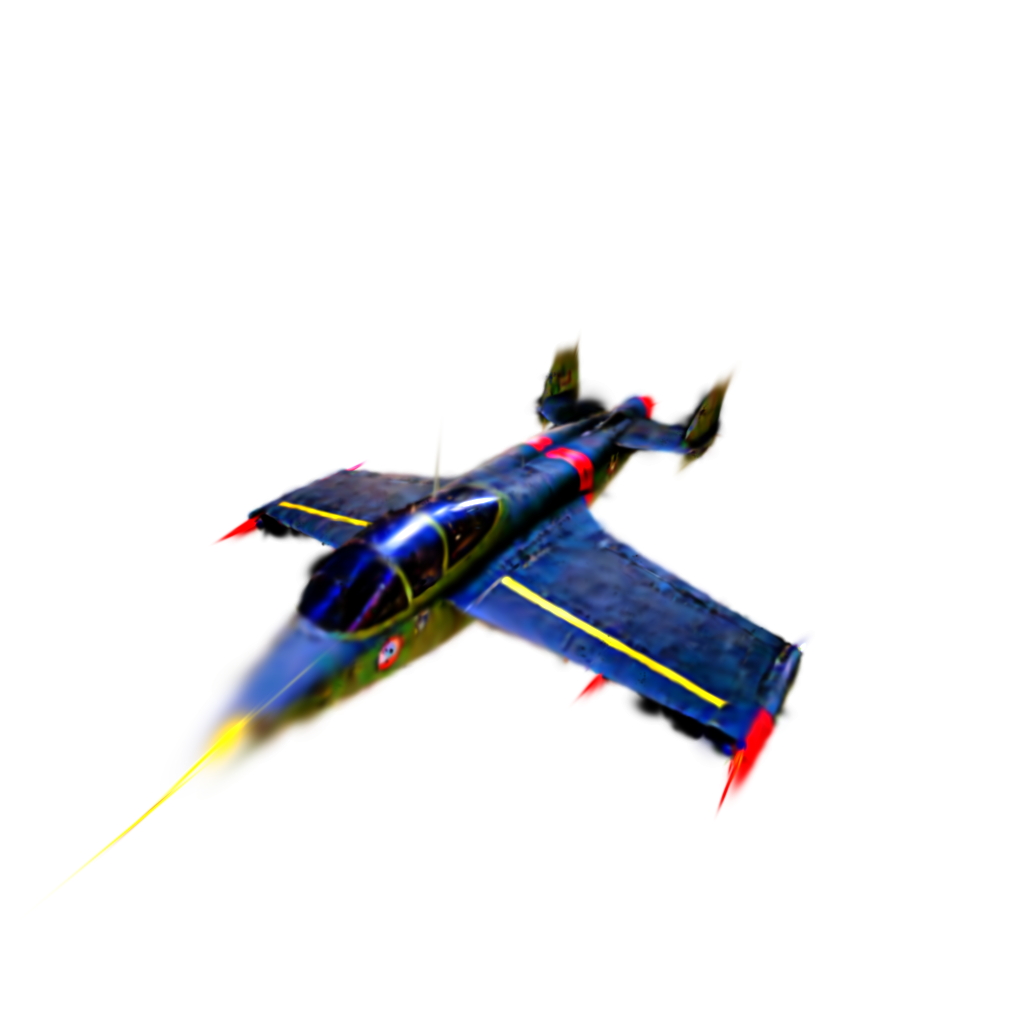}}{}
\vspace{1.5pt}
\\ \vspace{2pt}

\rotatebox{90}{\footnotesize{\emph{A modern chair}}}
\rotatebox{90}{\footnotesize{\emph{stainless steel}}}
\jsubfig{\includegraphics[height=1.53cm, trim={5.5cm 5.5cm 5.5cm 5.5cm}, clip]{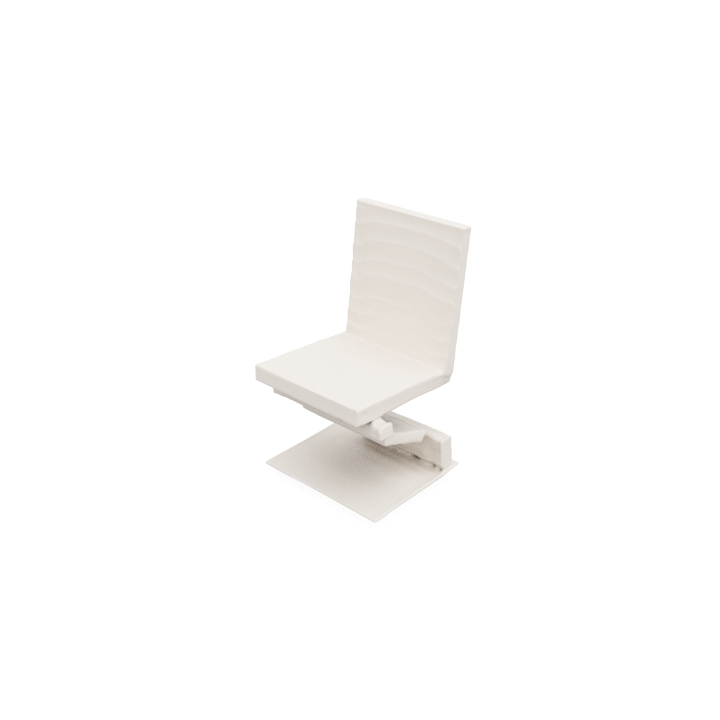}}{\footnotesize {Guidance}}\hfill
\jsubfig{\includegraphics[height=1.53cm, trim={6cm 6cm 6cm 6cm}, clip]{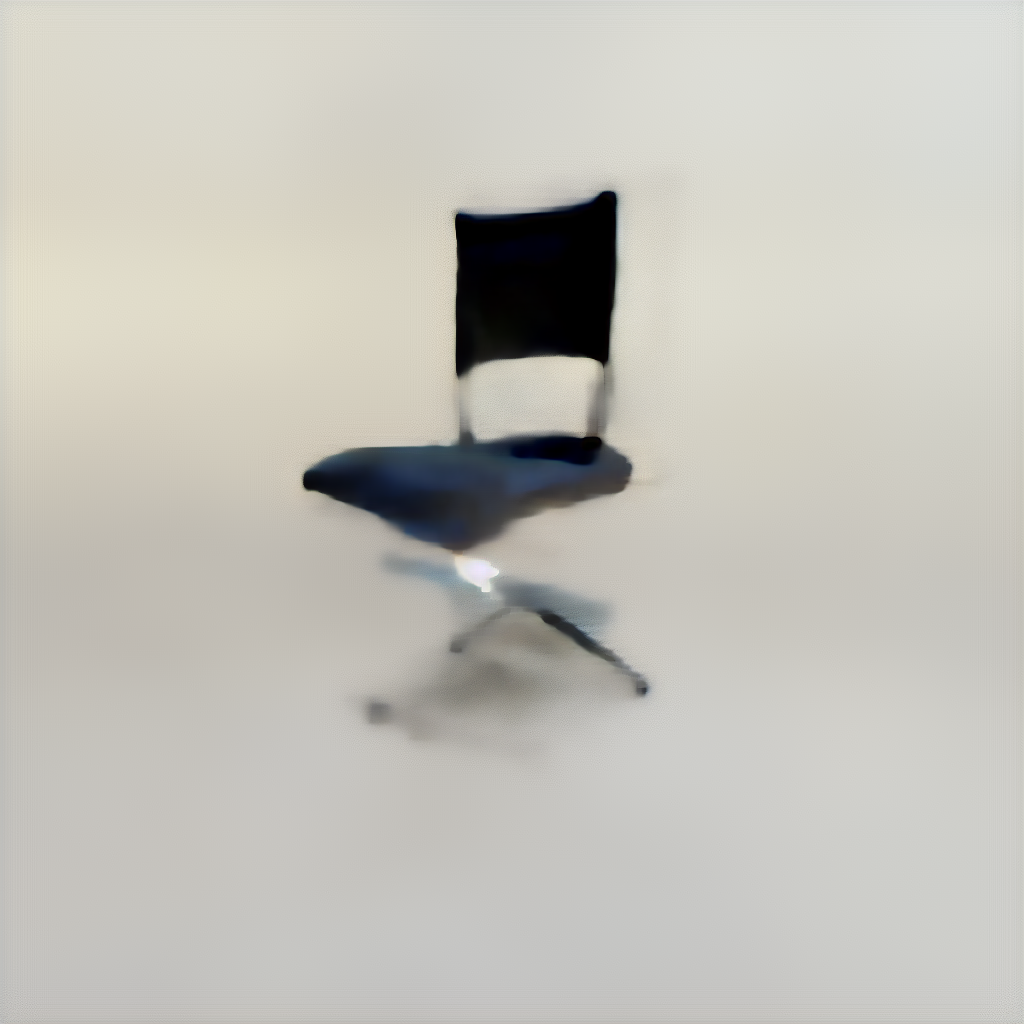}}{\footnotesize {SketchShape}} \hfill
\jsubfig{\includegraphics[height=1.53cm]{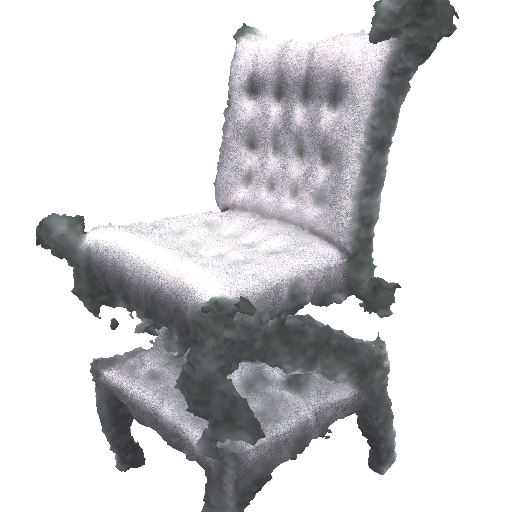}}{\footnotesize {Fantasia 3D}}\hfill
\jsubfig{\includegraphics[height=1.53cm, clip]{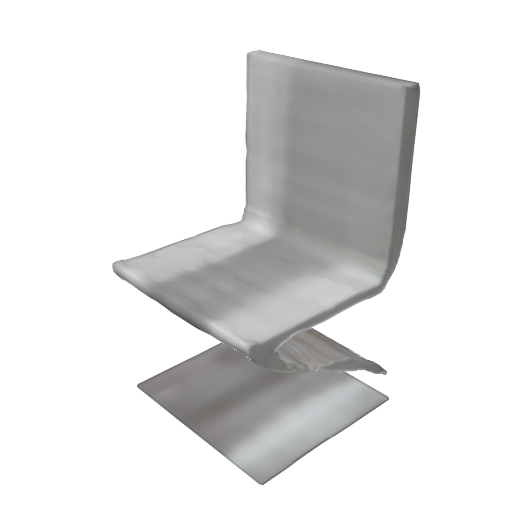}}{\footnotesize {Ours}}
\hfill
\jsubfig{\includegraphics[height=1.53cm, trim={3.5cm 3.5cm 3.5cm 3.5cm}, clip]{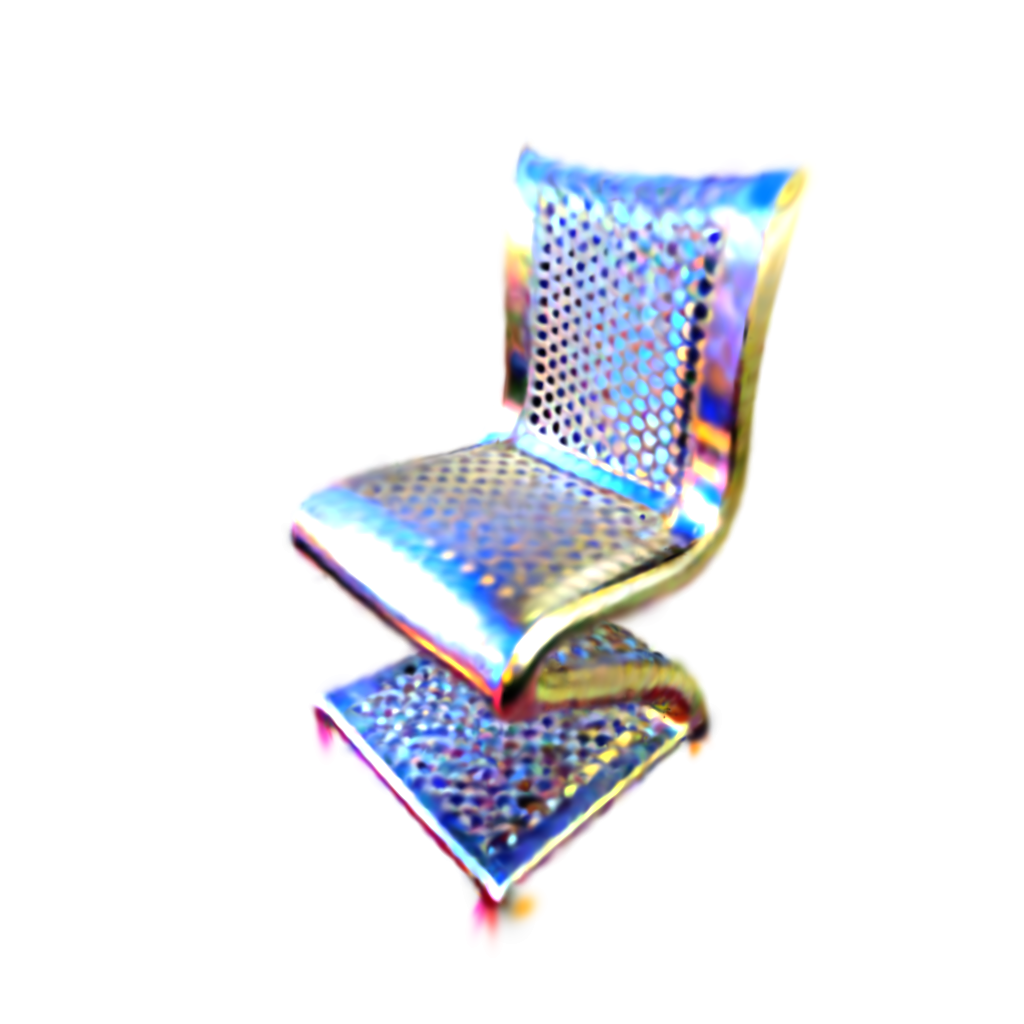}}{ \footnotesize Ours++}
\vspace{-1pt} 
\caption{\textbf{Text-conditional Abstraction-to-3D Comparison}. We compare to the results obtained using SketchShape~\cite{metzer2023latent} and Fantasia3D \cite{chen2023fantasia3d}. Methods are provided with a proxy cuboid-based abstract shape with a target prompt (left). As illustrated in the figure, our results better preserve the structure of the abstract guidance shape, while conveying the target text prompt. In the rightmost column (denoted as "Ours++"), we present results obtained after the optional refinement stage. } 

\label{fig:comparisons_abstraction}
\end{figure}

\medskip \noindent \textbf{Text-conditional Abstraction-to-3D}. Results for the text-conditional abstraction-to-3D task are reported in Table \ref{tab:abstraction_comparison}. As shown in the table, our generated 3D shapes can more faithfully preserve the abstract input guidance shapes, yielding better GD scores compared to Fantasia3D (GD is not computed for SketchShape as it outputs a NeRF representation). While Fantasia3D was not trained with any geometric supervision (thus explaining this lower GD score), we believe this metric is important in emphasizing that prior work are not suitable for this task. Note that our method also maintains high fidelity to the text prompts, outperforming both methods over $\text{CLIP}_{dir}$ while achieving comparable $\text{CLIP}_{sim}$ scores, all while exhibiting significantly faster inference times. See Figure \ref{fig:comparisons_abstraction} for a qualitative comparison, and additional results in Figure \ref{fig:abstraction_multiprompt_table}.

\begin{table}[t]
\centering
\resizebox{\linewidth}{!}{
\begin{tabular}{lcccc}
\toprule
 Method & $\text{CLIP}_{Sim}\uparrow$ & $\text{CLIP}_{Dir}\uparrow$ & $\text{GD} \downarrow$ & Run Time \\
\midrule
Latent-Paint & 0.27 & \textbf{0.01} & --- & $\sim \text{15 minutes}$ \\
Fantasia3D-Paint & \textbf{0.28} & \textbf{0.01} & --- & $\sim \text{15 minutes}$  \\ 
Fantasia3D & \textbf{0.28} & \textbf{0.01} & 0.06 & $\sim \text{30 minutes}$  \\
Ours & 0.27 & \textbf{0.01} & \textbf{0.01} & $\textbf{$\sim$ 20 seconds}$ \\  
\bottomrule
\end{tabular}
}
\caption{\textbf{3D Stylization Evaluation}. We compare against Latent-Paint~\cite{metzer2023latent} and two versions of Fantasia3D~\cite{chen2023fantasia3d} (with and without geometry modeling) over the 3D stylization task. As illustrated above, our edits are comparable with prior work and can be achieved orders of magnitude faster. GD is computed only for methods that can modify the geometry of the shape. }

\label{tab:stylization_comparison}
\end{table}
\medskip \noindent \textbf{3D Stylization}. Results for the 3D stylization task are reported in Table \ref{tab:stylization_comparison}. As illustrated, our edits capture the target text prompt well, yielding results comparable with Latent-Paint and Fantasia3D, while being orders of magnitude faster. Additional results can be seen in Figure \ref{fig:stylization_results}. 

\medskip \noindent \textbf{Additional Experiments}. To better illustrate what differences in CLIP-based metrics mean, we perform two additional experiments: (i) \emph{No Operation} baseline, measuring the $\text{CLIP}_{Sim}$ of the guidance shape to the target text, and (ii) \emph{Oracle}, measuring $\text{CLIP}_{Sim}$ on the ground-truth shape and $\text{CLIP}_{Dir}$ in the direction pointing from the guidance shape to the ground truth shape. These provide a lower and upper bound over these metrics in our setting.

For both the text-conditional abstraction-to-3D and the 3D stylization tasks, the \emph{No Operation} baseline produces lower $\text{CLIP}_{Sim}$ scores of 0.25 and 0.24 (for abstraction-to-3D and 3D stylization, respectively) and a $\text{CLIP}_{Dir}$ of 0.0, while the \emph{Oracle} produces $\text{CLIP}_{Sim}$ scores of 0.28 (for both tasks) and $\text{CLIP}_{Dir}$ scores of 0.02 and 0.05 (for abstraction-to-3D and 3D stylization, respectively). 
Indeed, for both tasks, the performance of our method, as well as competing methods, all fall in the range of the upper and lower bounds given by these baselines, with the $\text{CLIP}_{Dir}$ metric suggesting room for further improvement by future work.


\begin{figure} %
\centering
\rotatebox{90}{\footnotesize{\emph{A dasa\whitetxt{f}} }}
\rotatebox{90}{\footnotesize{\emph{dining table}}}
\jsubfig{\includegraphics[height=1.4cm, trim={0.5cm 2.0cm 3.5cm 2.0cm}, clip]{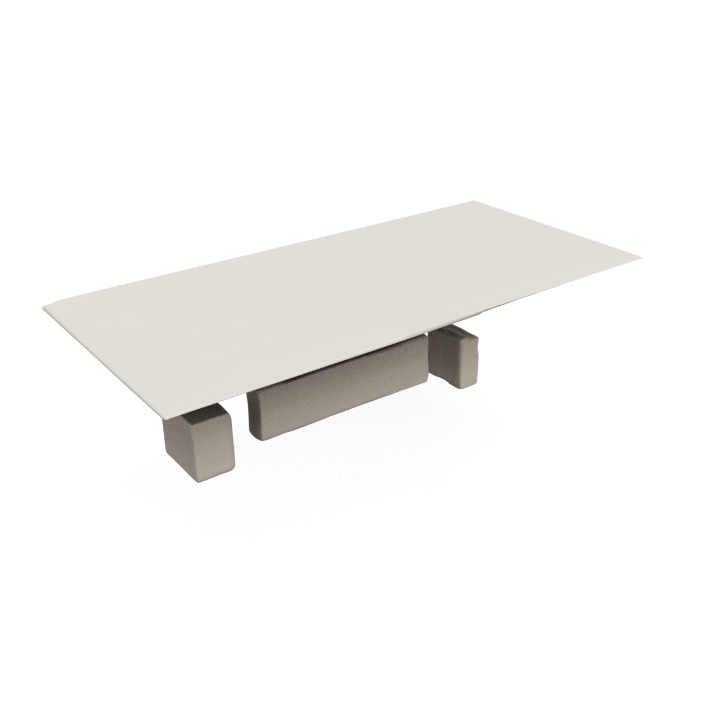}}{}\hfill
\jsubfig{\includegraphics[height=1.4cm, trim={0cm 0cm 0.5cm 0cm}, clip]{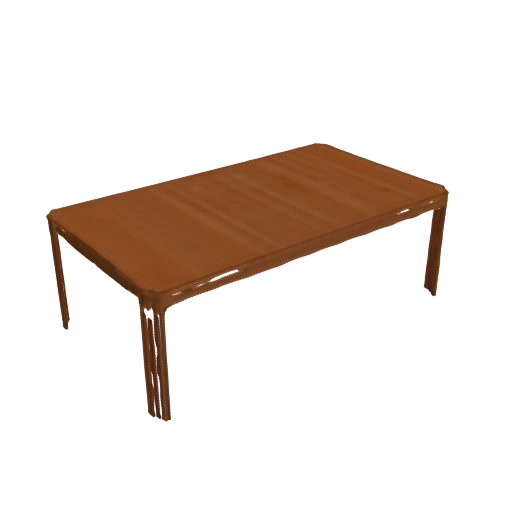}}{} \hfill
\jsubfig{\includegraphics[height=1.4cm, trim={2.0cm 2.0cm 3.5cm 2.0cm}, clip]{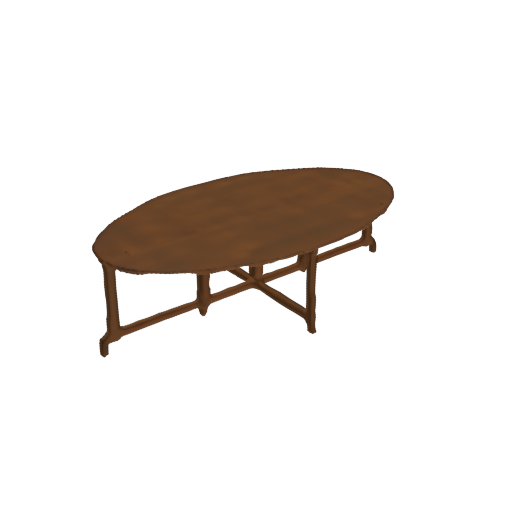}}{} 
\hfill
\jsubfig{\includegraphics[height=1.4cm, trim={2.0cm 2.0cm 4cm 2.0cm}, clip]{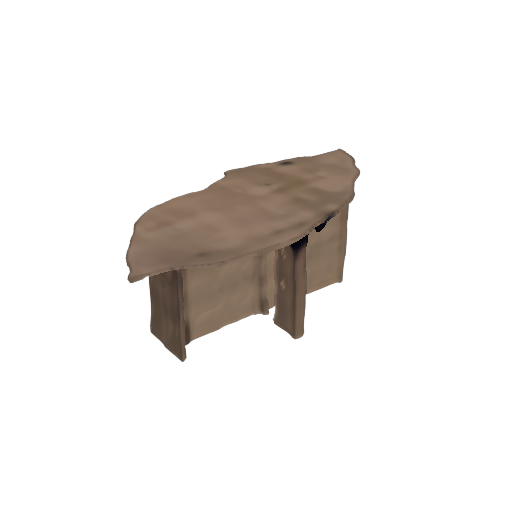}}{} \hfill
\jsubfig{\includegraphics[height=1.4cm, trim={1.5cm 2.0cm 2.0cm 2.0cm}, clip]{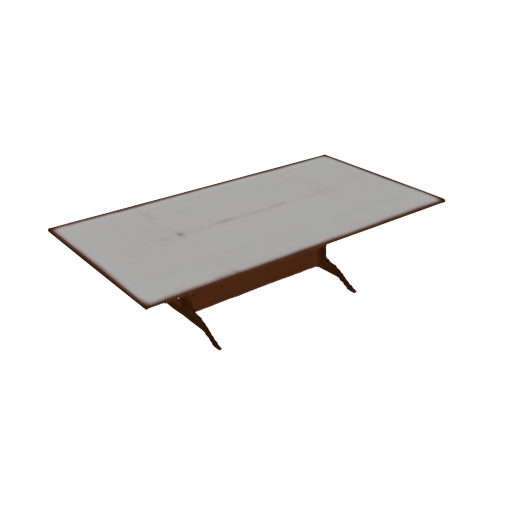}}{}
\vspace{4pt}
\\ 
\rotatebox{90}{\footnotesize{\emph{An office} }}
\rotatebox{90}{\footnotesize{\emph{lounge chair}}}
\jsubfig{\includegraphics[height=1.4cm, trim={6.5cm 7.5cm 6.5cm 6.5cm}, clip]{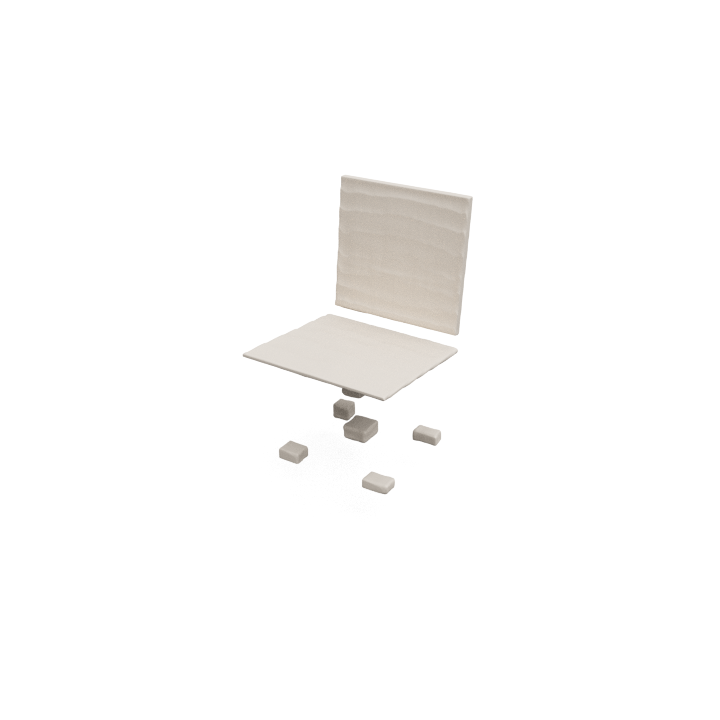}}{\footnotesize {Guidance}}\hfill
\jsubfig{\includegraphics[height=1.4cm]{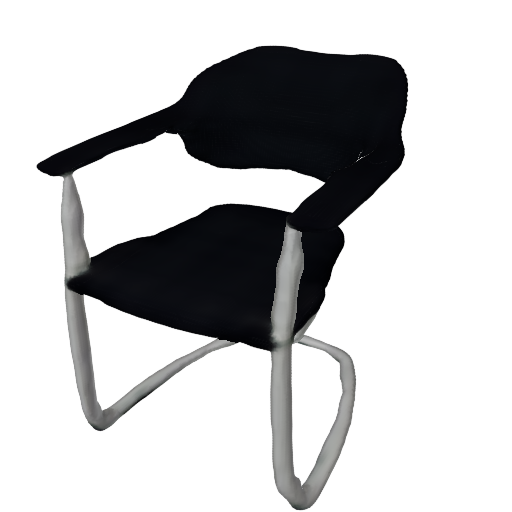}}{\footnotesize {\shapE{}$_{\text{FT}}$}}\hfill
\jsubfig{\includegraphics[height=1.4cm, trim={1.0cm 1.0cm 1.0cm 1.0cm}, clip]{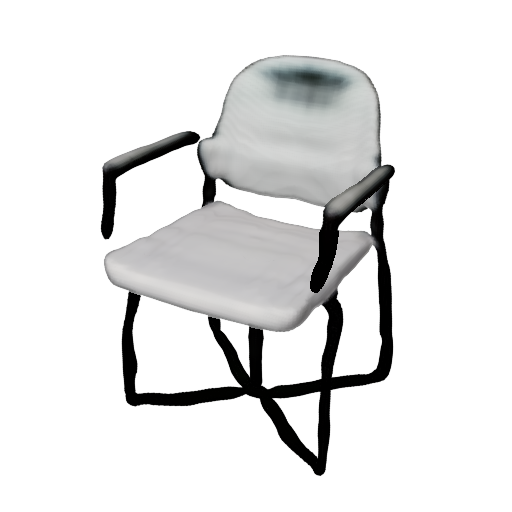}}{\footnotesize {SDEdit3D}}\hfill
\jsubfig{\includegraphics[height=1.4cm, trim={2.0cm 2.0cm 2.0cm 2.0cm}, clip]{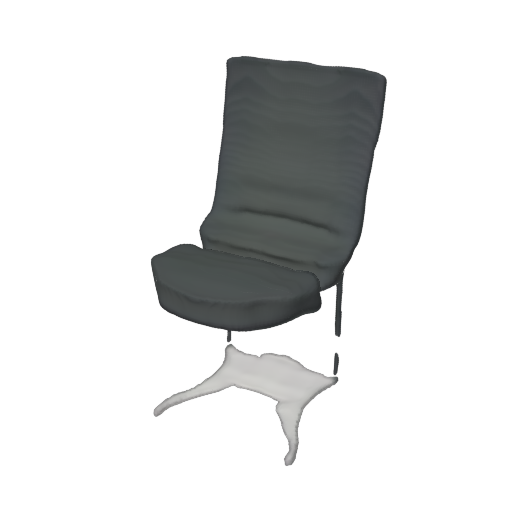}}{\footnotesize {ControlNet3D}} \hfill
\jsubfig{\includegraphics[height=1.4cm, trim={1.0cm 1.0cm 1.0cm 1.0cm}, clip]{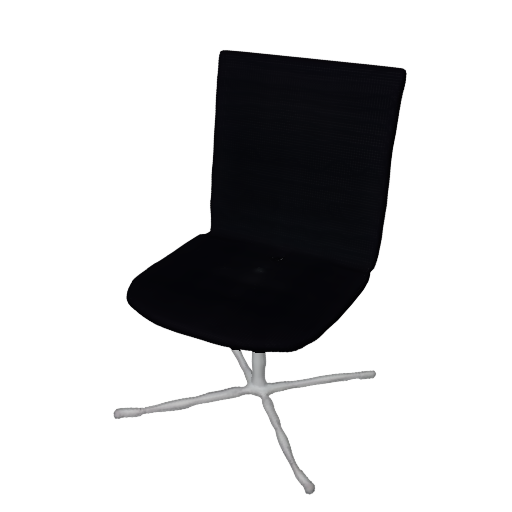}}{\footnotesize {Ours}}
\vspace{-5pt} 
\caption{\textbf{Qualitative ablation results}, obtained for test shapes from the text-conditional abstraction-to-3D task. We compare our cross-entity attention mechanism (right) with several baselines, detailed in Section \ref{sec:ablations}. As illustrated above, our approach allows for generating 3D shapes that conform to the guidance structure significantly better than baseline methods, while remaining faithful to the target text prompt.
}
\label{fig:ablations}
\end{figure}

\subsection{Ablations}
\label{sec:ablations}
Next we ablate our cross-entity attention mechanism, demonstrating that comparable structural control cannot be achieved with baseline methods. We compare to the following baselines: (i) \shapE{}$_{\text{FT}}$, the original \shapE{} model finetuned on each dataset with text guidance only (no structural guidance is added). (ii) SDEdit3D, inspired by the image editing technique SDEdit~\cite{meng2021sdedit}, which uses the \shapE{}$_{\text{FT}}$ models. During inference, noise is added to the conditional latent, and it is denoised with the target textual prompt. (iii) CrossOnly, an ablated version of our framework that uses Cross-Attention instead of our Cross-Entity Attention mechanism, i.e. using only the conditional queries. (iv) ControlNet3D, inspired by the network architecture used in ControlNet~\cite{zhang2023adding}, which freezes and clones the original network blocks of \shapE{}, creating a frozen and trainable copy of it. The guidance shape is passed through the trainable copy with intermediate outputs added to the appropriate frozen copy blocks as residuals through a zero-convolution (see the supplementary material for more details). 

We conduct experiments over the text-conditional abstraction-to-3D task. As illustrated in Figure \ref{fig:ablations}, these baselines methods cannot faithfully preserve the conditional guidance shape. For instance, the ControlNet3D results are of significantly lower quality in comparison to our method. We attribute this visual gap to the much larger number of parameters that need to be optimized in comparison to our method (50M vs. 330M additional parameters), making this method more prone to overfitting on the relatively small datasets we use (i.e. resulting in the model forgetting its pretrained knowledge). Quantitatively, the baselines yield significantly worse GD scores: 0.06 (\shapE{}$_{\text{FT}}$), 0.05 (SDEdit3D), 0.03 (CrossOnly) and 0.03 (ControlNet3D), compared to 0.01 for our approach, further showing that their outputs strongly deviate from the guidance shapes.

In the supplementary material, we also conduct additional perceptual studies to evaluate user's preference of our results over the ControlNet3D baseline. We also conduct additional ablations to motivate our design choices. 
In particular, we modify our cross-entity attention mechanism in various ways, including removing the zero-convolution operators and performing cross-attention over the Keys or Values. 
These ablations demonstrate that our proposed cross-entity mechanism allows for better preserving the structure of the guidance shape in comparison to other possible modifications. 

\subsection{Limitations}
\label{sec:limitations}

\begin{figure*}
    \begin{minipage}{0.47\textwidth}
    \jsubfig{\includegraphics[height=2.3cm, trim={7cm 8.5cm 7cm 7cm}, clip]{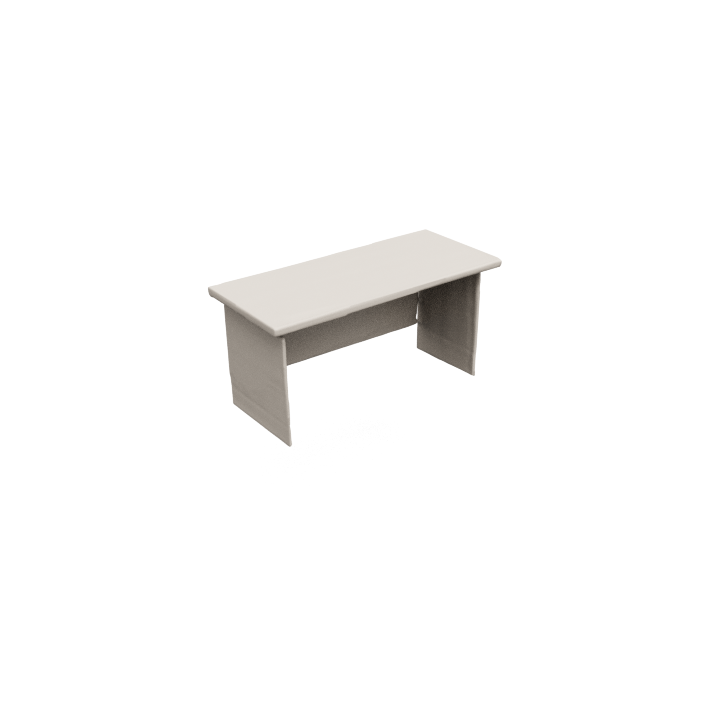}}{}\hfill
    \jsubfig{\includegraphics[height=2.3cm, trim={1cm 2cm 2cm 2cm}, clip]{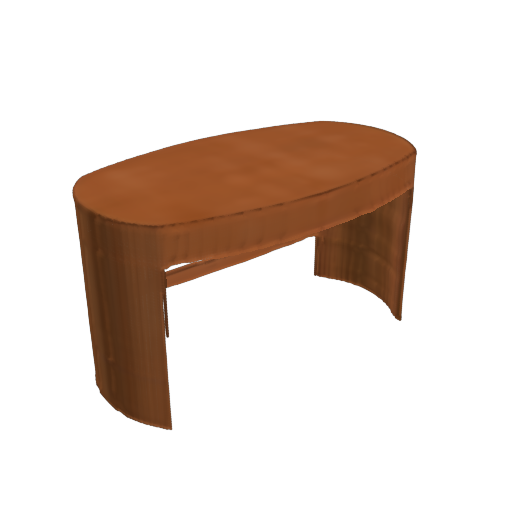}}{}\hfill
    \jsubfig{\includegraphics[height=2.3cm, trim={1cm 2cm 0.5cm 2cm}, clip]{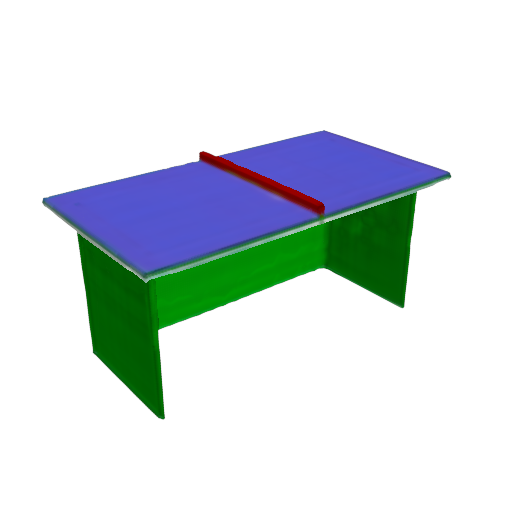}}{} \vspace{1.5pt}
    \\
    \jsubfig{\includegraphics[height=2.3cm, trim={6.5cm 9cm 7cm 6cm}, clip]{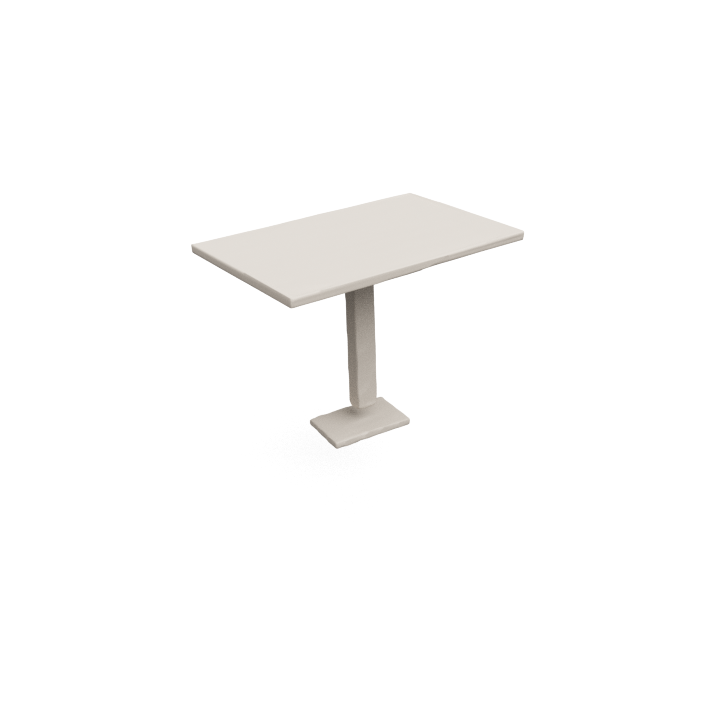}}{}\hfill
    \jsubfig{\includegraphics[height=2.3cm, trim={1cm 2.5cm 0.5cm 2cm}, clip]{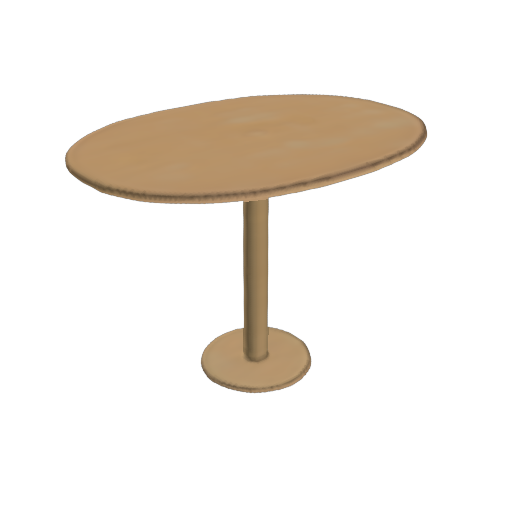}}{}\hfill
    \jsubfig{\includegraphics[height=2.3cm, trim={1.2cm 1cm 0.5cm 1cm}, clip]{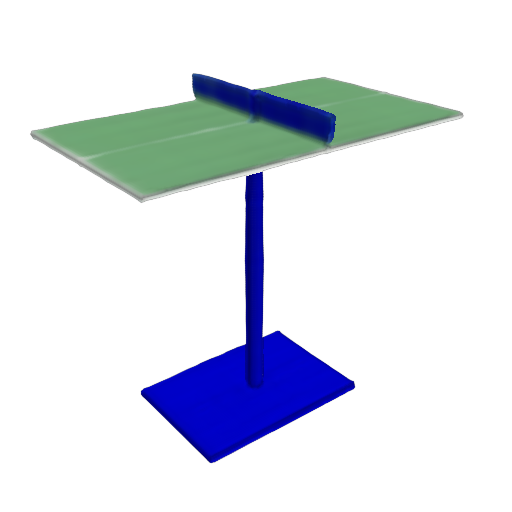}}{} \vspace{1.5pt}
    \\
    \jsubfig{\includegraphics[height=2.3cm, trim={4cm 8cm 6cm 4cm}, clip]{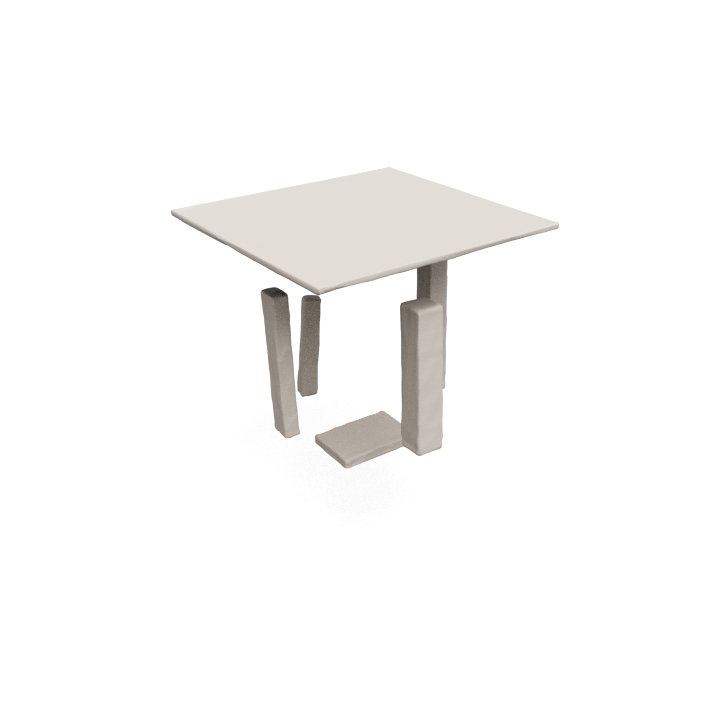}}{\footnotesize {Guidance}}\hfill
    \jsubfig{\includegraphics[height=2.3cm]{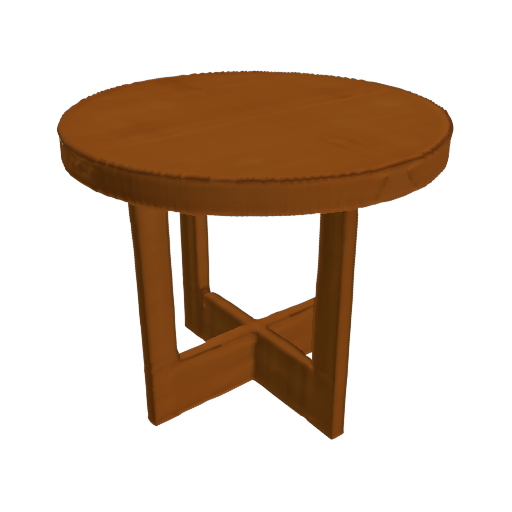}}{\footnotesize {\emph{A round coffee table}}}\hfill
    \jsubfig{\includegraphics[height=2.3cm]{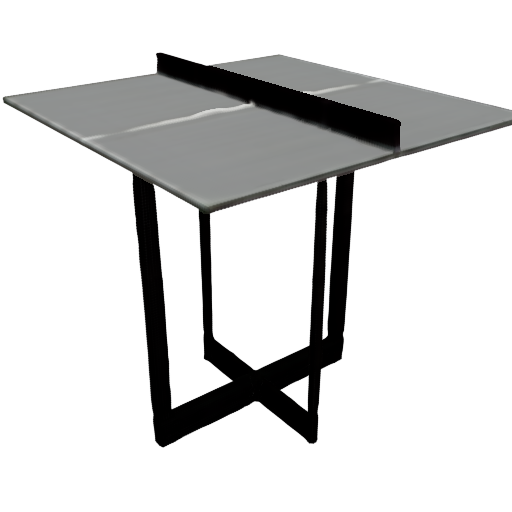}}{\footnotesize {\emph{A ping pong table}}}
    \vspace{-1pt} 
    \caption{Text-conditional abstraction-to-3D results for test shapes from the \emph{Table} category. The leftmost column displays the guidance input --- a proxy cuboid-based shape. The remaining columns showcase our results over two different target text prompts. 
    }
    \label{fig:abstraction_multiprompt_table}
\end{minipage}\hfill
\hspace{10pt}
    \begin{minipage}{0.47\textwidth}
    \vspace{-75pt}
    \jsubfig{\includegraphics[height=2cm, trim={2cm 2cm 2cm 2cm}, clip]{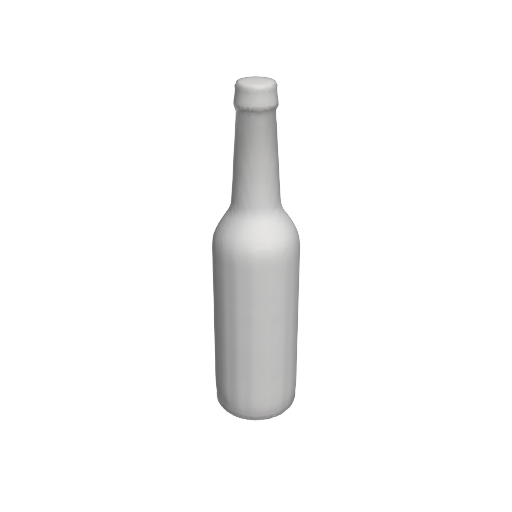}}{\footnotesize {Guidance}}\hfill
    \jsubfig{\includegraphics[height=2cm, trim={2cm 2cm 2cm 2cm}, clip]{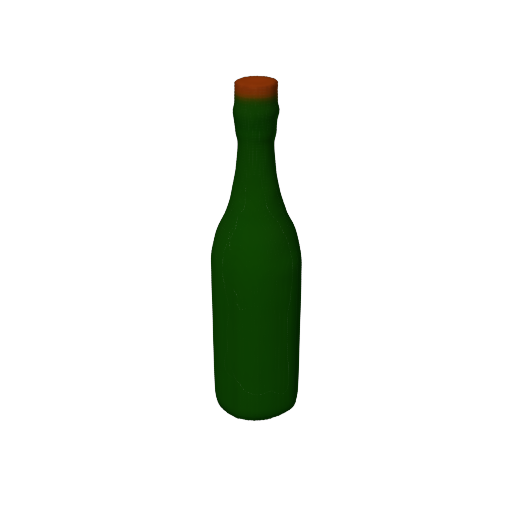}}{\footnotesize {\emph{A corked bottle}}}\hfill
    \jsubfig{\includegraphics[height=2cm, trim={2cm 2cm 2cm 2cm}, clip]{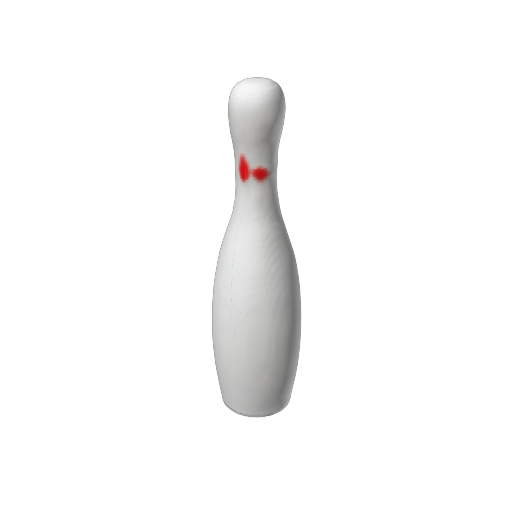}}{\footnotesize {\emph{A bowling pin}}} \vspace{1.5pt}
    \jsubfig{\includegraphics[height=2cm, trim={2cm 2cm 2cm 2cm}, clip]{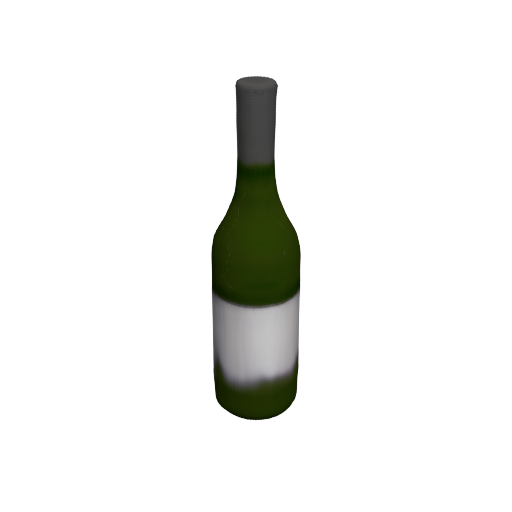}}{\footnotesize {\emph{A wine bottle}}}
    \vspace{1.5pt}
    \\
    \jsubfig{\includegraphics[height=2cm, trim={2cm 2cm 2cm 2cm}, clip]{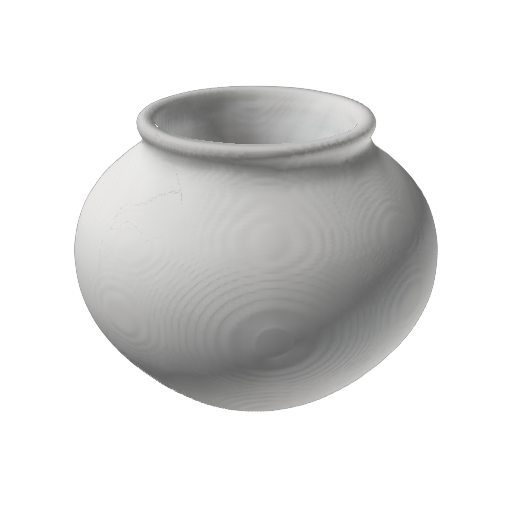}}{\footnotesize {Guidance}}\hfill
    \jsubfig{\includegraphics[height=2cm, trim={2cm 2cm 2cm 2cm}, clip]{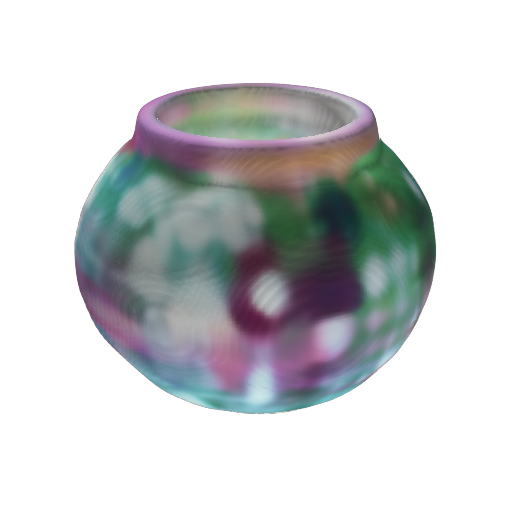}}{\footnotesize {\emph{A modern vase}}}\hfill
    \jsubfig{\includegraphics[height=2cm, trim={2cm 2cm 2cm 2cm}, clip]{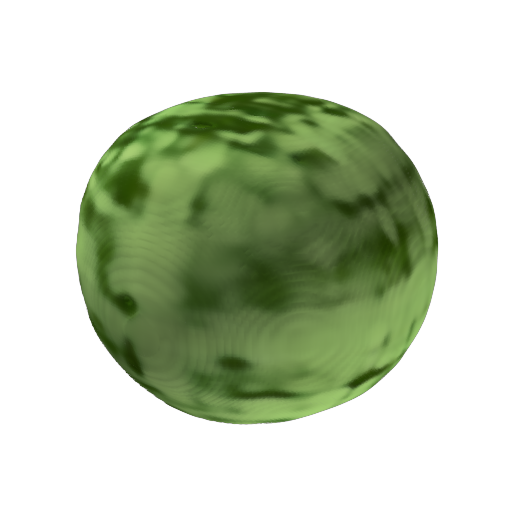}}{\footnotesize {\emph{A watermelon}}} \vspace{1.5pt}
    \jsubfig{\includegraphics[height=2cm, trim={2cm 2cm 2cm 2cm}, clip]{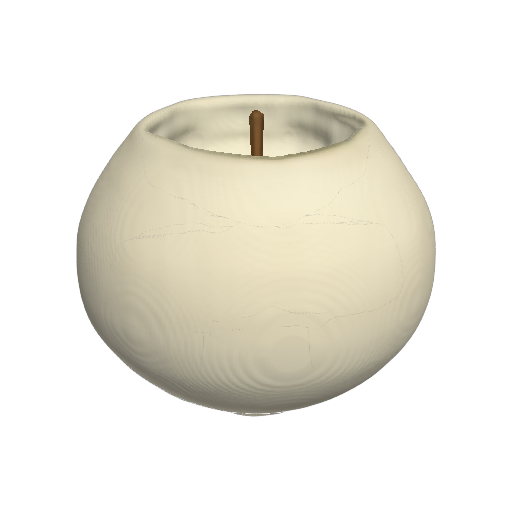}}{\footnotesize {\emph{A candle}}}
    \caption{3D stylization results results are shown above. The leftmost column displays the guidance input --- an uncolored 3D asset. The remaining columns showcase how the guidance input is styled according to the target text prompt.}
    \label{fig:stylization_results}
\end{minipage}
\bigskip
    \begin{minipage}{0.47\textwidth}
    \jsubfig{\includegraphics[height=2cm, trim={2cm 2cm 2cm 2cm}, clip]{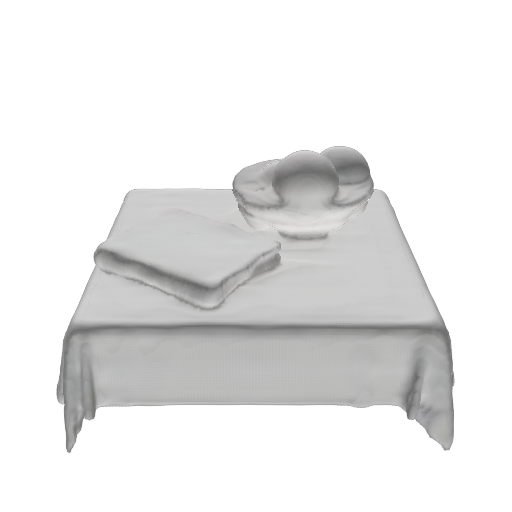}}{\footnotesize {Guidance}}\hfill
    \jsubfig{\includegraphics[height=2cm, trim={2cm 2cm 2cm 2cm}, clip]{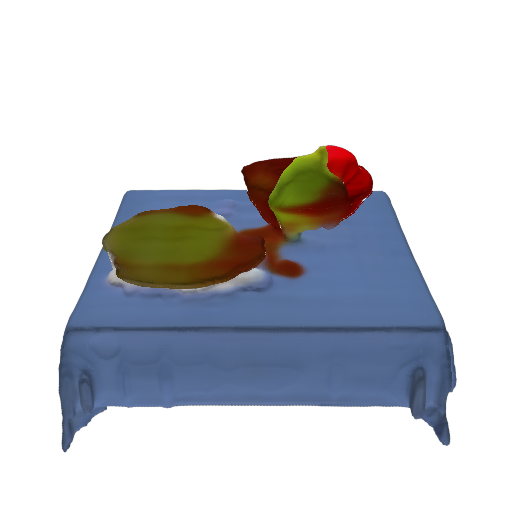}}{\footnotesize {\emph{A bowl of fruit next to a book on a tablecloth}}}\hfill
    \jsubfig{\includegraphics[height=2cm, trim={3cm 3cm 3cm 3cm}, clip]{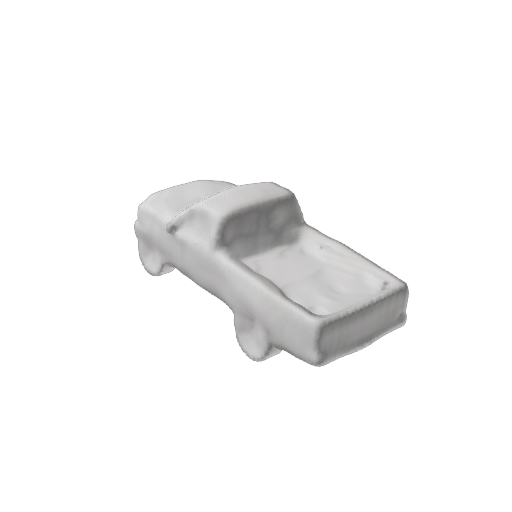}}{\footnotesize {Guidance}} \vspace{1.5pt}
    \jsubfig{\includegraphics[height=2cm, trim={3cm 3cm 3cm 3cm}, clip]{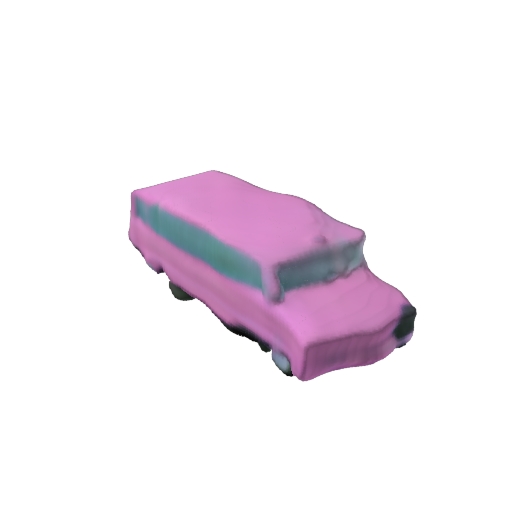}}{\footnotesize {\emph{A pink bus}}}
    \caption{\textbf{Limitations}. Above, we present two failure cases. These likely result from incorrect multiple attribute binding (the fruit bowl and the book colored similarly) or insufficient preservation of the guidance structure (changing the guidance pickup truck into a bus and switching the back of the guidance truck to the front of the bus).}
    \label{fig:limitations}
\end{minipage}\hfill
\label{fig:additional}
\end{figure*}

Our method allows for learning various types of structural priors for generating text-conditional shapes guided by 3D inputs, but there are several limitations to consider, as also shown in Figure \ref{fig:limitations}. 
First, our approach inherits limitations from diffusion-based techniques and in particular from \shapE{}, which we build our method upon. While \shapE{} can generate diverse high-quality 3D shapes, it still struggles to bind multiple attributes to objects, limiting the scope of possible object edits. Furthermore, we observe that highly complicated shapes are not often successfully encoded, leading to noisy data used for both training and evaluation. As our approach can be added on top of other transformer-based 3D diffusion models, we expect that with the emergence of stronger backbones, more powerful edits can be achieved.

Additionally, our approach does not offer explicit control over the tradeoff between the fidelity to the input guidance shape and the consistency with the target prompt. This often leads to results which are either not functionally plausible (for instance, see the ping pong table on the top row of Figure \ref{fig:abstraction_multiprompt_table} where the table's legs make it challenging for the table to correctly function as intended) or conversely do not sufficiently preserve the guidance structure.

\section{Conclusion}

In this work, we presented \methodName{}, a new approach for adding structural control to 3D diffusion models. We demonstrated that our method facilitates several text-conditional 3D editing tasks, without the need for tailoring the network architectures or training objectives.  Our work represents a step towards the goal of democratizing 3D generation, 
making 3D object editing more accessible to non-experts by providing them with task-specific structural control within seconds. Technically, we introduced the cross-entity attention mechanism, which allows for mixing latent representations corresponding to different 3D shapes while preserving the capabilities of the pretrained 3D diffusion model. We believe that our mechanism could potentially improve a wide variety of applications where guidance is injected into a generative framework, beyond the realm of 3D shape generation and manipulation. 

\begin{acks}
We thank Peter Hedman, Ron Mokadi, Daniel Garibi, Itai Lang and Or Patashnik for helpful discussions. This work was supported by the Israel Science Foundation (grant no. 2510/23) and by the Alon Scholarship.
\end{acks}

\bibliographystyle{ACM-Reference-Format}
\bibliography{main}
\clearpage

\clearpage
\appendix
\noindent {\LARGE\textbf{\methodName{} Supplementary Material}}
\section{Additional Details}
\label{sec:details}

\subsection{\methodName{} Implementation Details}
\label{sec:imp_details}

\medskip \noindent \textbf{Network Architecture}.

Overall, we add a total of 50M new parameters to the original \shapE{} architecture, which uses 315M parameters for optimizing the transformer-based diffusion model. These additional parameters include the weights and biases of the 'zero-conv' blocks, the linear projection layer for the conditional input, as well as the additional query projection blocks added to each residual attention block of the \shapE{} backbone.


\medskip \noindent \textbf{Training and Inference}.
Tasks are trained using their task-specific datasets as detailed in Section 4 in the main paper. We use a batch size of 6 and a learning rate of $2.8\cdot10^{-5}$. Given the varying dataset sizes, we use a different number of iterations for each application (derived from the number of epochs, the batch sizes, and the size of the training dataset). Training runtimes, along with the number of epochs and iterations, are reported Table \ref{tab:training_times}. Additionally, to encourage our network to rely more on the structural guidance and less on text we replace the guidance text with an empty string with a probability of 0.5. We perform this replacement for the text-conditional abstraction-to-3D and 3D stylization tasks, while for semantic shape editing we use the guidance text always as we do not want to encourage random shape edits. Instead, we encourage structural guidance by feeding the target shape instead of the conditional guidance shape with a probabilitiy of 0.5 (as mentioned in the main paper). 

We follow the inference procedure used by \shapE{} and sample latents with 64 denoising steps. All experiments and training are conducted  on a single RTX A5000 GPU (24GB VRAM). 

\subsection{Evaluation}
\label{sec:eval_details}
%
We evaluate our results using two CLIP-based metrics for the text-conditional abstraction-to-3D and the 3D stylization tasks. 
The CLIP model we used for both of these metrics is ViT-B/32 and $\text{CLIP}_{Dir}$ is calculated for each rendered image from the output 3D object in relation to the corresponding rendered image from the input 3D conditional prior shape. For these tasks, we transform our output 3D object and the input 3D conditional shape into meshes, and calculate Chamfer distance over point clouds (with length $N=4096$) sampled from these meshes. For sampling and computing Chamfer distance, we use \href{https://pytorch3d.org/}{\texttt{the following library}}. 

For the semantic editing task, we run the evaluation script \\ \href{https://github.com/optas/changeit3d/blob/main/changeit3d/scripts/evaluate_change_it_3d.py}{\texttt{evaluate\_change\_it\_3d.py}} available in the publicly available implementation of \href{https://github.com/optas/changeit3d}{\texttt{ChangeIt3D}}. To compute the metrics on our results, we decode both the input guidance and the output latent to generate point clouds that could be employed for a quantitative evaluation. Note that for evaluating ChangeIt3D we utilize the original ShapeTalk point clouds (not the ones encoded by \shapE{}) in conjunction with the point clouds generated by ChangeIt3D. 

\begin{table}[t]
\centering
\resizebox{\linewidth}{!}{
\begin{tabular}{lccc}
\toprule
 Task & Iterations & Epochs & Training Time \\
\midrule
Semantic Shape Editing - \emph{Chair} & 35K & 25 & 11.5 hours \\
Semantic Shape Editing - \emph{Table} & 44K & 25 & 14.5 hours \\
Semantic Shape Editing - \emph{Lamp} & 27.1K & 25 & 9 hours \\
Text-Conditional Abstraction-to-3D - \emph{Table} & 15.3K & 20 & 5 hours \\
Text-Conditional Abstraction-to-3D - \emph{Airplane} & 19.4K& 20 & 6.5 hours \\
Text-Conditional Abstraction-to-3D - \emph{Chair} & 9.4K & 20 & 3 hours \\
3D Stylization & 27.5K & 20 &  9 hours \\
\bottomrule
\end{tabular}
}
\caption{\textbf{Training Runtimes}. Number of iterations, number of epochs and the overall training time for each model is reported in the table above.}

\label{tab:training_times}
\end{table}

\begin{figure*} %
\centering
\jsubfig{\includegraphics[width=1.95\columnwidth]{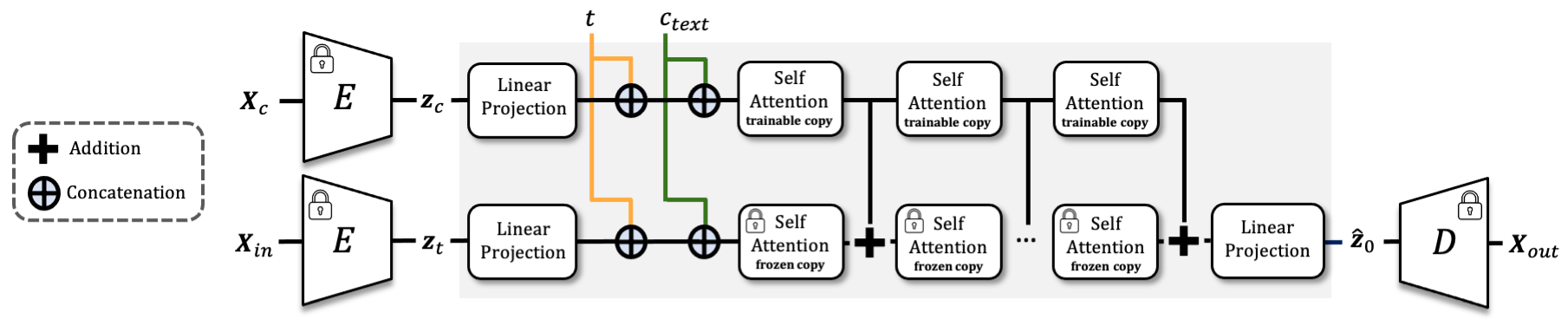}}{}\hfill
\vspace{-8pt}
\caption{\textbf{ControlNet3D Baseline Architecture}. To ablate our proposed architecture, we finetune a 3D diffusion model (\emph{i.e.} \shapE{}) with an architecture inspired by the network architecture used in ControlNet~\cite{zhang2023adding}. The original network blocks of \shapE{} are frozen and trainable copies of them are created. The guidance shape is passed through the trainable copy with intermediate outputs added to the appropriate frozen copy blocks as residuals through a zero-convolution. After finetuning, the output latent representation can be decoded into a 3D shape $\mathbf{X}_{out}$, represented as either a neural radiance field or a signed texture field.
}
\label{fig:architecture_controlnet3d}
\end{figure*}

\subsection{Comparisons and Ablations}
\label{sec:prior}
Below we provide all the details needed to reproduce the comparisons and ablations shown in the paper. 
\subsubsection*{SketchShape}
We use the \href{https://github.com/eladrich/latent-nerf}{\texttt{code} provided by the authors} and the input parameters used are the default parameters in the \\ \href{https://github.com/eladrich/latent-nerf/tree/main/scripts}{\texttt{train\_latent\_nerf.py} script}. The input for this method is a mesh obtained from our input 3D guidance shape and the corresponding target text prompt.

\subsubsection*{Latent-Paint}
As in SketchShape, we used the code provided by the authors and the default input settings provided for latent paint, which are given in the  \href{https://github.com/eladrich/latent-nerf/tree/main/scripts}{\texttt{train\_latent\_paint.py} script}. The input for this method is a mesh obtained from our input 3D guidance shape and the corresponding target text prompt. We compared our method to Latent-Paint using only the two CLIP-based metrics as this method does not modify the geometry of the input mesh and transforms only the appearance of the input mesh. 

\subsubsection*{Fantasia3D}
We compare against two variants of Fantasia3D, one that only performs appearance modeling, denoted as Fantasia-Paint, and the full model, which modifies both the object’s geometry and appearance. We used the \href{https://github.com/Gorilla-Lab-SCUT/Fantasia3D}{\texttt{code} provided by the authors}. For the model that modifies both geometry and appearance, we first run the \texttt{train.py} script with the default input settings provided for in the \texttt{Gundam\textunderscore{}geometry.json} configuration file, only changing the \texttt{sdf\textunderscore{}init\textunderscore{}shape\textunderscore{}scale} parameter to be 1 for all dimensions (as opposed to the default 1.2 for all dimensions). 
The input for this first part is a mesh obtained from the input 3D guidance shape and the corresponding input text prompt. We then use the output mesh from this first part as the input mesh to the second part, while using the same text prompt, and run  
the \texttt{train.py} script with the the default input settings provided for in the \texttt{Gundam\textunderscore{}appearance.json} configuration file.
For the comparisons to Fantasia-Paint we run the \texttt{train.py} script with the the default input settings provided in the \texttt{Gundam\textunderscore{}appearance.json} configuration file. The input for this method is a mesh obtained from the input 3D guidance shape and the corresponding target text prompt.


\subsubsection*{ChangeIt3D}
The authors of Changeit3D provided a \href{https://docs.google.com/forms/d/e/1FAIpQLSdOouzvK0zmjvmBoiQhbfnhe1Kac72XNmHXzshn6_KUEjw8QQ/viewform}{link} for obtaining the ShapeTalk dataset and the weights of their trained model. We used the script \href{https://github.com/optas/changeit3d/blob/main/changeit3d/scripts/evaluate_change_it_3d.py}{\texttt{evaluate\_change\_it\_3d.py}} in order to load their model and extract their results. 

\subsubsection*{ControlNet3D}

To ablate our proposed architecture, we finetune a \shapE{} model with a network architecture inspired by the one used in ControlNet~\cite{zhang2023adding} for controlling 2D diffusion models. We freeze and clone the original network blocks of \shapE{}, thus creating a frozen and trainable copy of each network block. For the first network blocks $B_{0_{frozen}}$ and $B_{0_{trainable}}$, the guidance shape latent $z_{c}$, used here as the control condition, is passed through a zero-convolution operator $\mathcal{Z}_{0}$, and summed with the input latent $z_{t}$ and passed through $B_{0_{trainable}}$, such that $y_{0}=B_{0_{trainable}}(\mathcal{Z}_{0}(z_{c})+z_{t})$, while the input latent $z_{t}$ is passed through  $B_{0_{frozen}}$, such that $x_{0}= B_{0_{frozen}}(z_{t})$. We then sum $y_{0}$, after the zero-convolution operator, with $x_{0}$, resulting in $x_{1}=x_{0}+\mathcal{Z}_{1}(y_{0})$. We pass $x_{1}$ as the input to the next frozen network block $B_{1_{frozen}}$, and we pass $y_{0}$ as input to the following trainable network block $B_{1_{trainable}}$. We continue in the same manner for all remaining network blocks. See Figure \ref{fig:architecture_controlnet3d} for a visualization of the modified network architecture for this ablation.
We train the ControlNet3D models using the same training parameters as mentioned in Section \ref{sec:imp_details}.

\ignorethis{
The results of this survey can be found in Table \ref{tab:fantasia3d_survey}. As is evident in the results, a significant majority of users preferred our results when compared against those produced by the Fantasia3D baseline. In the Abstract-to-3D setting the results also show that user preference of our method was especially distinct for results from the "table" category. Our intuition for this is that this category usually requires somewhat less detail in order to convey the text in a convincing manner, this makes it ideal for our method, which is more inclined to convey convincing structure over generating fine detail.
}

\begin{figure}
\jsubfig{\fcolorbox{cyan}{cyan}{\includegraphics[height=2cm]{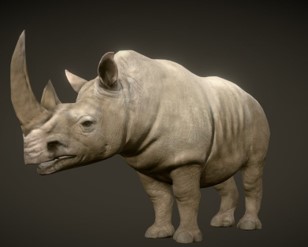}}}{\textit{{\footnotesize a white rhino with horns}}}
\hfill
\jsubfig{\fcolorbox{orange}{orange}{\includegraphics[height=2cm]{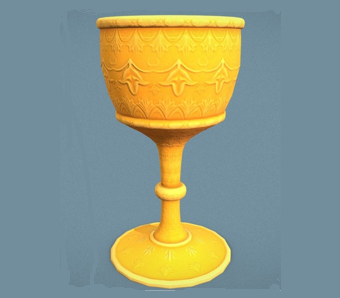}}}{\textit{{\footnotesize a yellow goblet with an ornate design}}} 
\hfill
\jsubfig{\fcolorbox{purple}{purple}{\includegraphics[height=2cm]{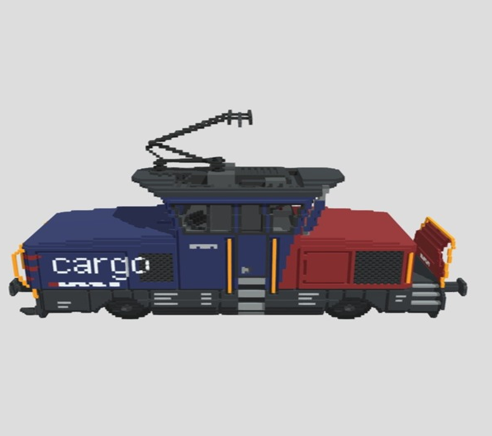}}}{\textit{{\footnotesize EEM923 train car model}}} 
  {\begin{flushleft} \leftskip=0.1pt   \footnotesize{
  \textbf{Instruction}: Provide a description of the object in the image represented by the following metadata. 
Answer in one sentence, and write "unknown" if the answer is unclear.
 \\
 \textbf{Image metadata}: \\
  \textcolor{cyan}{
  \quad\textbf{Key}: rhinoceros. \\
  \quad\textbf{Name}: Sculptober Day 17: Horn. \\
  \quad\textbf{Tags}: white, africa, rhino, rhinoceros, horn, realism, sculptober, \\
  \quad sculptober2020, whiterhinoceros. \\
  \quad\textbf{Categories}: animals-pets, nature-plants. \\
  \quad\textbf{Caption}: My entry for day 17 of sculptober horn. 
  }
  \\
   \textcolor{orange}{
  \quad\textbf{Key}: chalice. \\
  \quad\textbf{Name}: The ancient  one!\\
  \quad\textbf{Tags}: prop, substancepainter, maya, zbrush, stylized. \\
  \quad\textbf{Categories}: furniture-home
 \\
  \quad\textbf{Caption}: This trinket is said to have been passed down from generation\\
  \quad to generation. The owners never lived to see their grandchildren. 
  \\
  \textcolor{purple}{
  \quad\textbf{Key}: cabin. \\
  \quad\textbf{Name}: EEM923. \\
  \quad\textbf{Tags}: train, track, mod, railway, swiss, switzerland. \\
  \quad\textbf{Categories}: cars-vehicles. \\
  \quad\textbf{Caption}: }
  }}
 \end{flushleft}} %
  \caption{
  \textbf{Enriching Objaverse dataset using InstructBLIP}. Presented above are three exemplars illustrating the outputs of the fine-tuned InstructBLIP model (shown in \emph{italics} directly below the images). These outputs were generated from the instruction and input metadata (shown using the same colors as their associated images). As illustrated above, this procedure allows for extracting meaningful text descriptions from the noisy Internet metadata.  
  }\label{fig:objaverse}
\end{figure}
\subsection{Datasets}
\label{sec:dataset}
\subsubsection*{Text-conditional Abstraction-to-3D} 
We follow the train/test splits defined in Yang and Chen~\cite{yang2021unsupervised}. For obtaining reasonable evaluation runtimes, we limit the test sets to 100 randomly sampled test objects from each category (as the alternative methods require at least 15 minutes inference time per test sample). 
\subsubsection*{3D Stylization}
To extract target text prompts for 3D assets in the Objaverse dataset, we finetune 
%
InstructBLIP \cite{instructblip} using a subset of 500 manually annotated examples. During this fine-tuning process, the model's input consisted of an RGB image of the object and an instruction prompt requesting a concise description of the object in the image given the accompanying metadata, as illustrated on Figure \ref{fig:objaverse}. In order to fine-tune the model efficiently, we used Low-Rank Adaption (LoRA) \cite{hu2022lora} with attention dimension of 16, alpha parameter for scaling of 32 and dropout probability of 0.05, adapting only the 'queries' and 'values' projections. We perform 3 epochs of fine-tuning with a learning rate of 1e-4. 

The test set for this task is made up of 100 random examples not included in the training set. 

\subsubsection*{Semantic Shape Editing}
We follow the train/test splits defined in Achlioptas \etal~\cite{achlioptas2022changeit3d}. To only train and evaluate over distractor and target models that are structurally similar, we use only \emph{Hard Context} examples, and remove pairs with extremely large geometric differences (as measured using Chamfer distance; 0.140 for tables, 0.080 for chairs and 0.100 for lamps). We also filter noisy pairs that have negative LAB scores (signifying that the distractor object conveys the target prompt more than the target) and a high class distorion score (greater than 0.5).

\section{Additional Comparisons and Ablations}
\label{sec:ab}

\begin{figure} %
\centering 
\rotatebox{90}{\hspace{14pt}\emph{The left side}}
\rotatebox{90}{\hspace{16pt}\emph{is longer}}
\jsubfig{\includegraphics[height=2.5cm, trim={5.6cm 5.6cm 5.6cm 5.6cm}, clip]{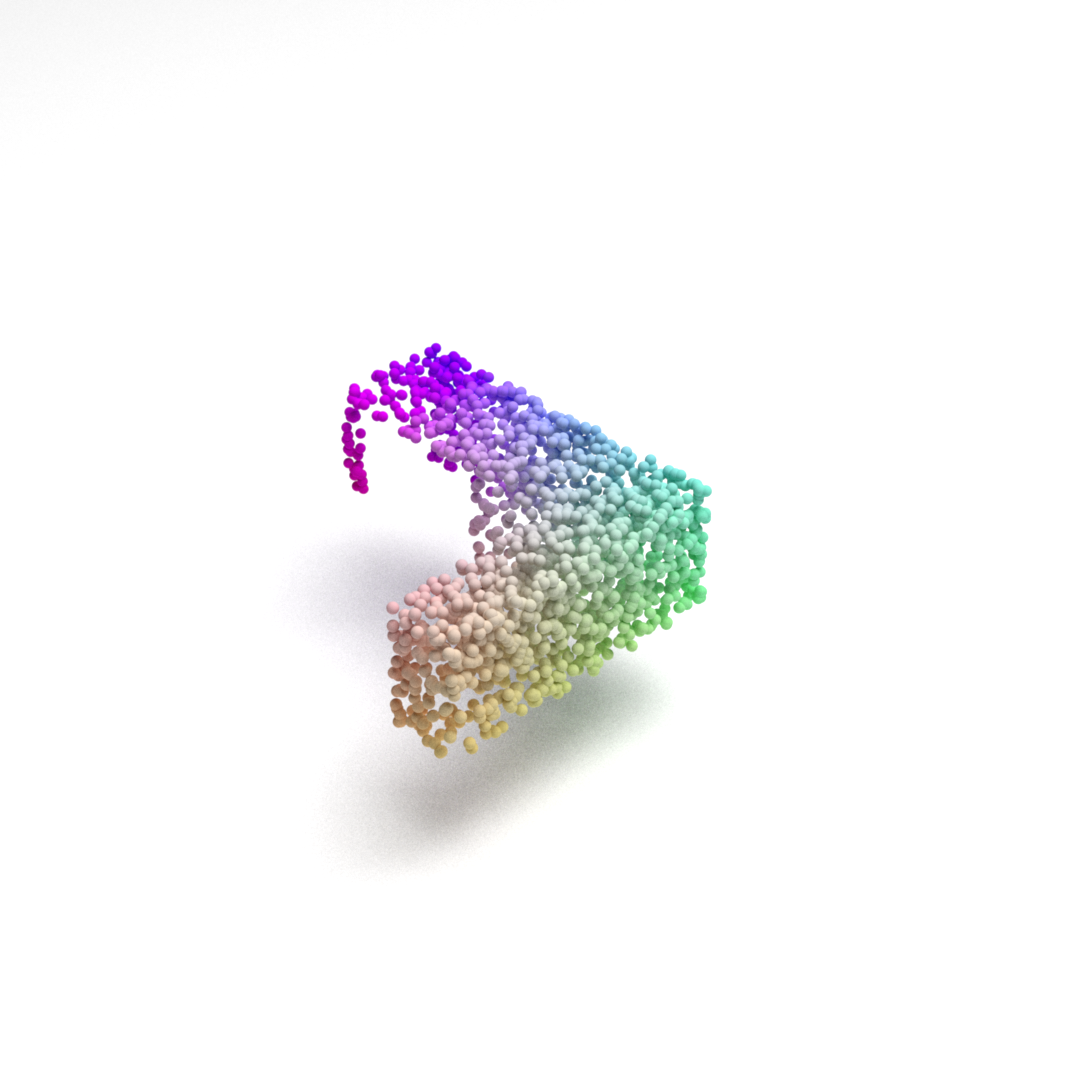}}{}
\jsubfig{\includegraphics[height=2.5cm, trim={6.3cm 5.5cm 6.3cm 7.5cm}, clip]{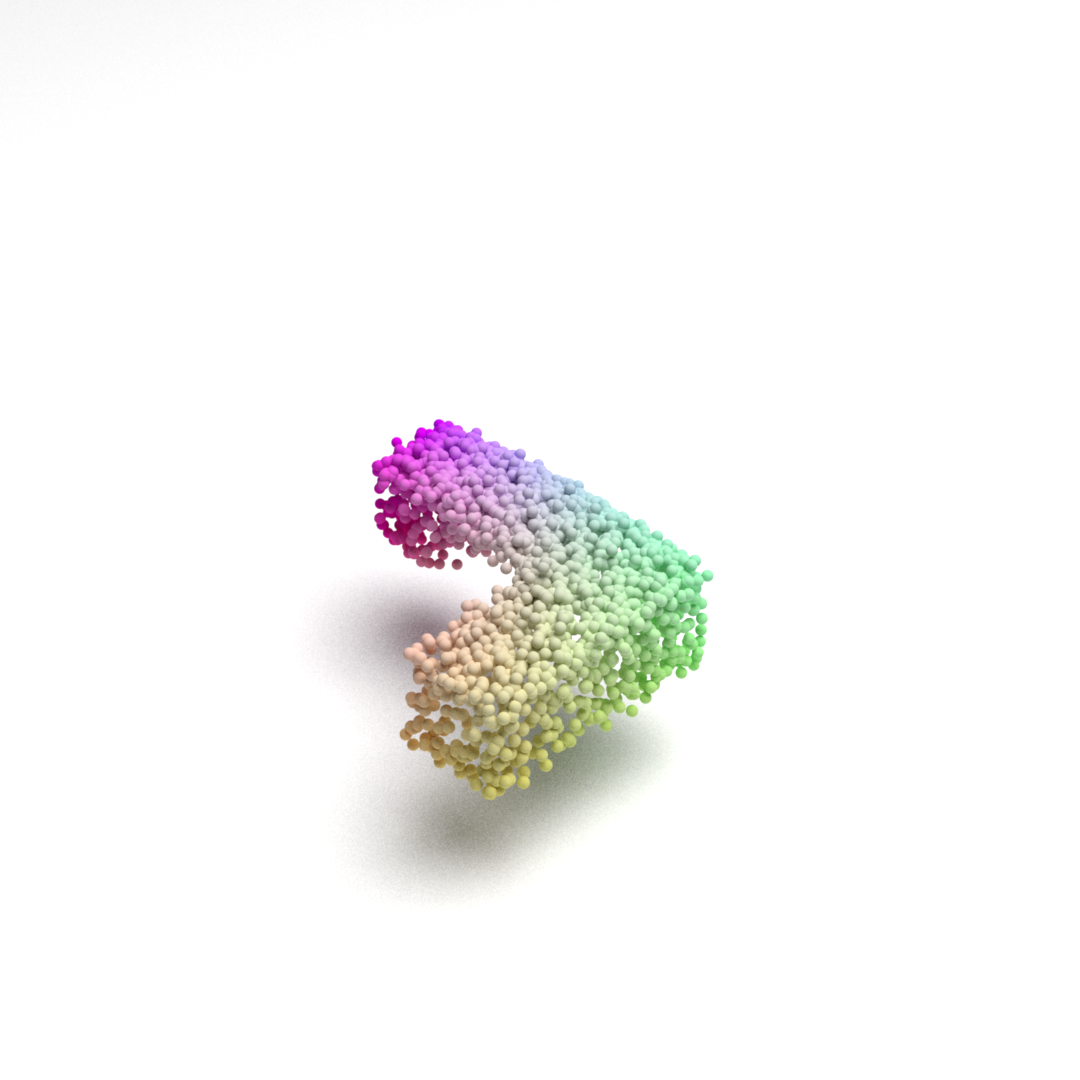}}{\hspace{-10pt}$[$\textcolor{red}{0.33}, \textcolor{teal}{0.005} $]$}
\jsubfig{\includegraphics[height=2.5cm, trim={5cm 5cm 5cm 5cm}, clip]{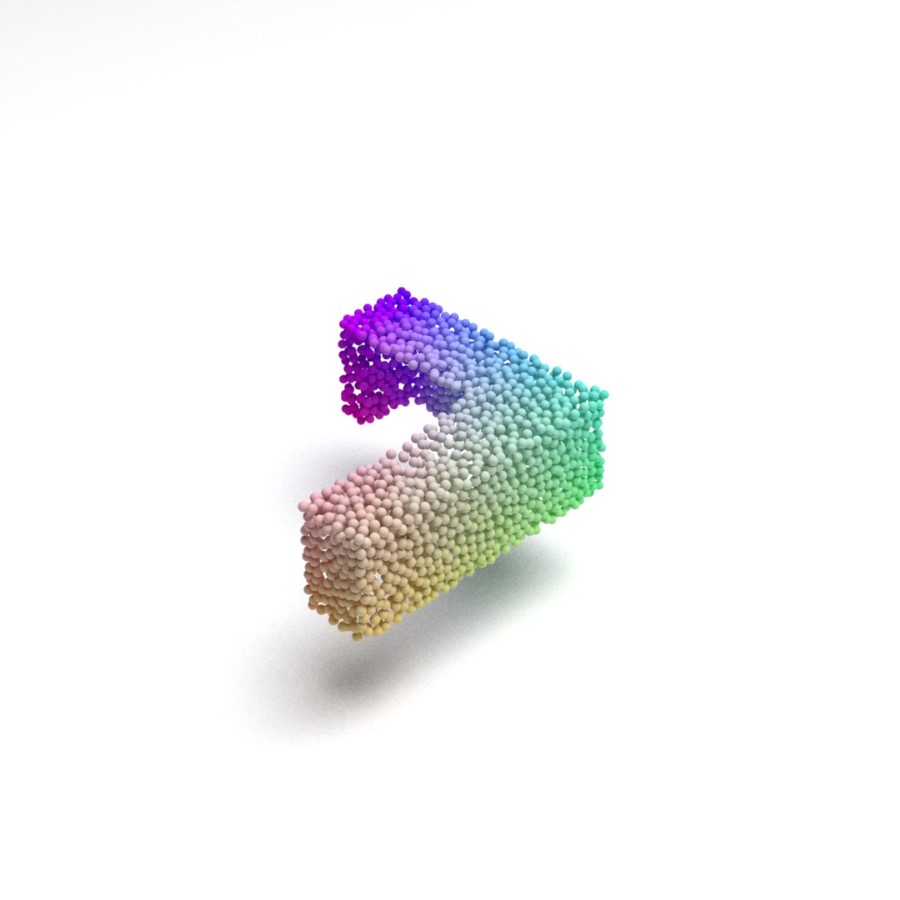}}{$[$\textcolor{red}{0.48}, \textcolor{teal}{0.007}$]$}
\\ 
\rotatebox{90}{\hspace{14pt}\emph{The seat is}}
\rotatebox{90}{\hspace{16pt}\emph{flipped up}}
\jsubfig{\includegraphics[height=2.5cm, trim={5cm 5cm 5cm 5cm}, clip]{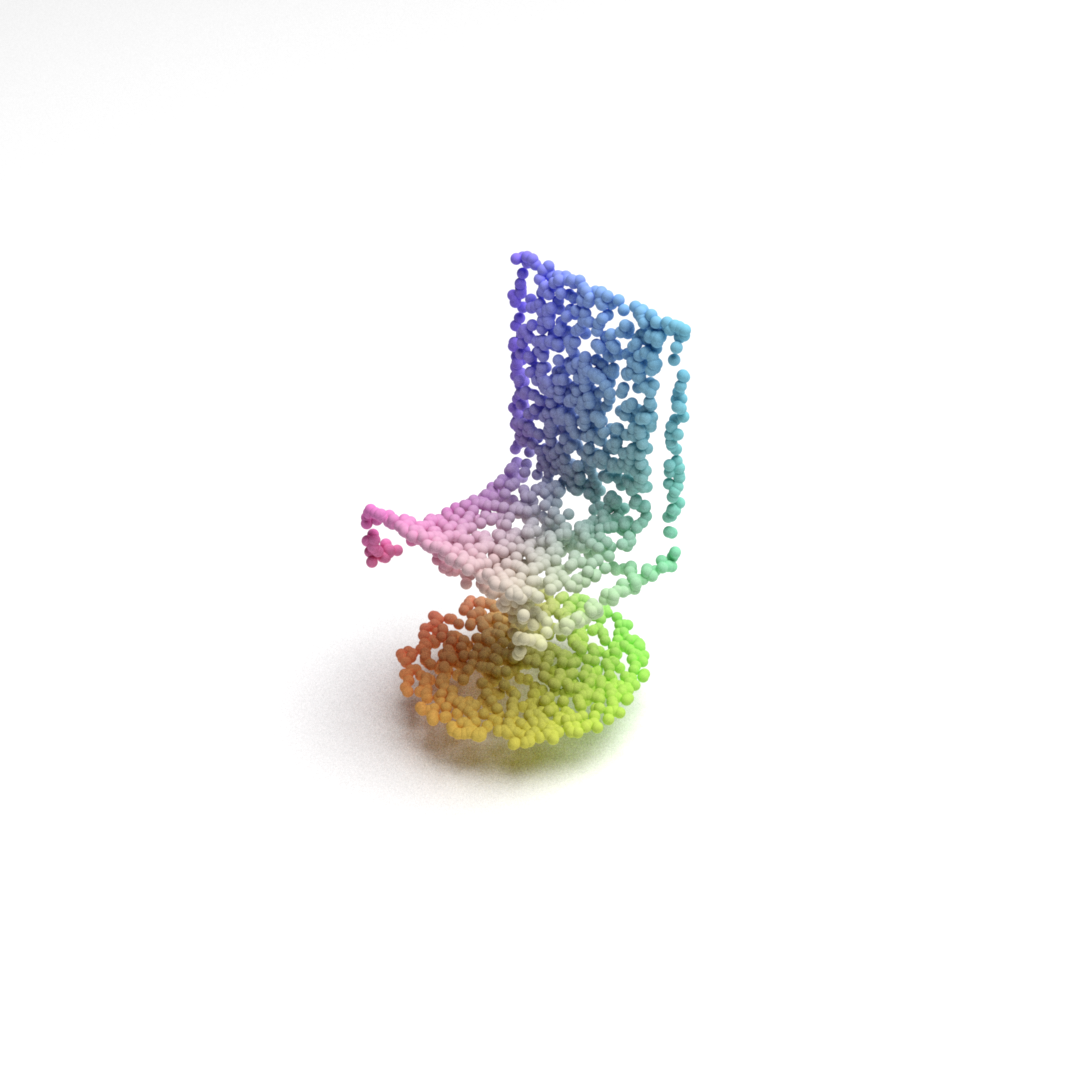}}{}
\jsubfig{\includegraphics[height=2.5cm, trim={5cm 5cm 5cm 5cm}, clip]{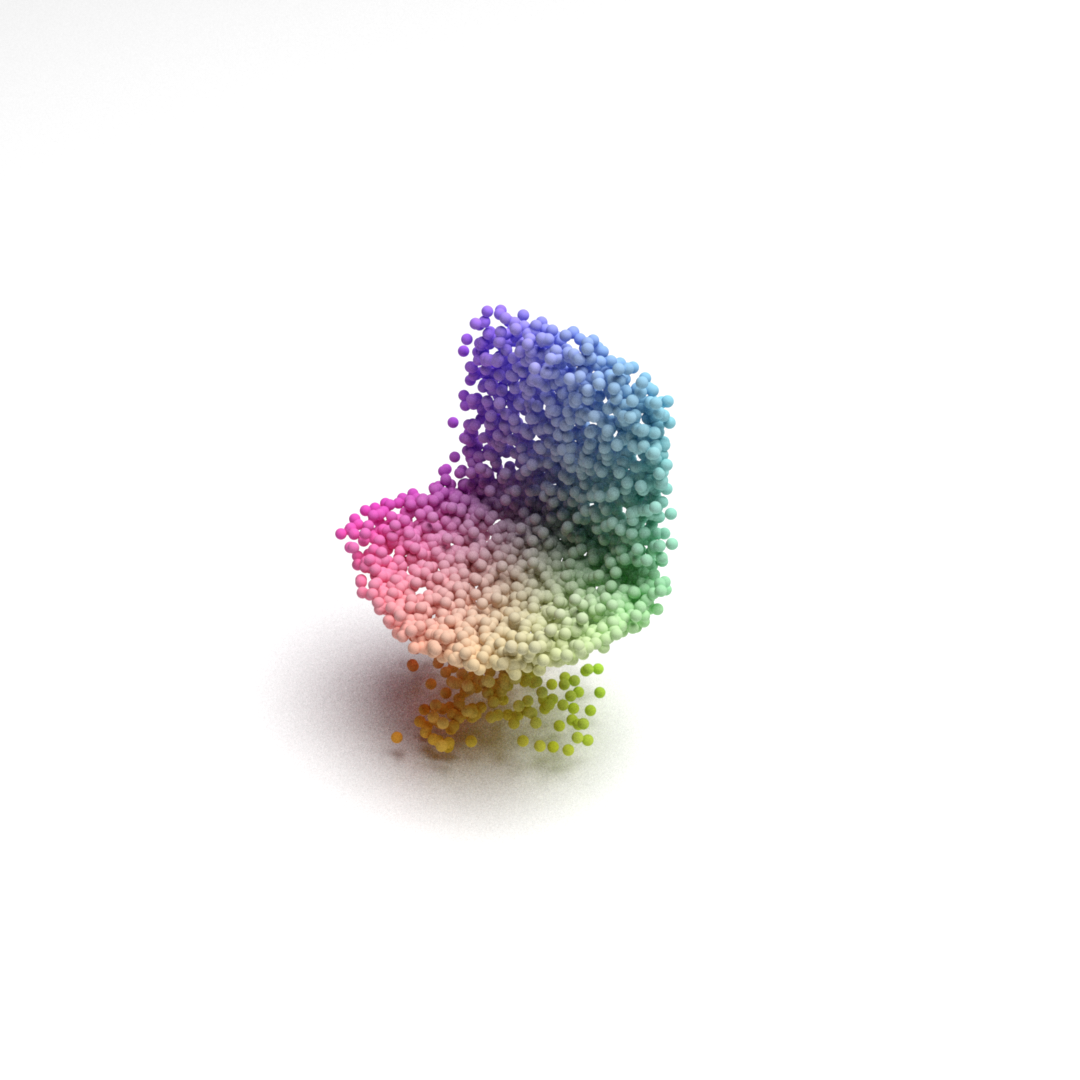}}{\hspace{-10pt}$[$\textcolor{red}{0.29}, \textcolor{teal}{0.004} $]$}
\jsubfig{\includegraphics[height=2.5cm, trim={5cm 5cm 5cm 5cm}, clip]{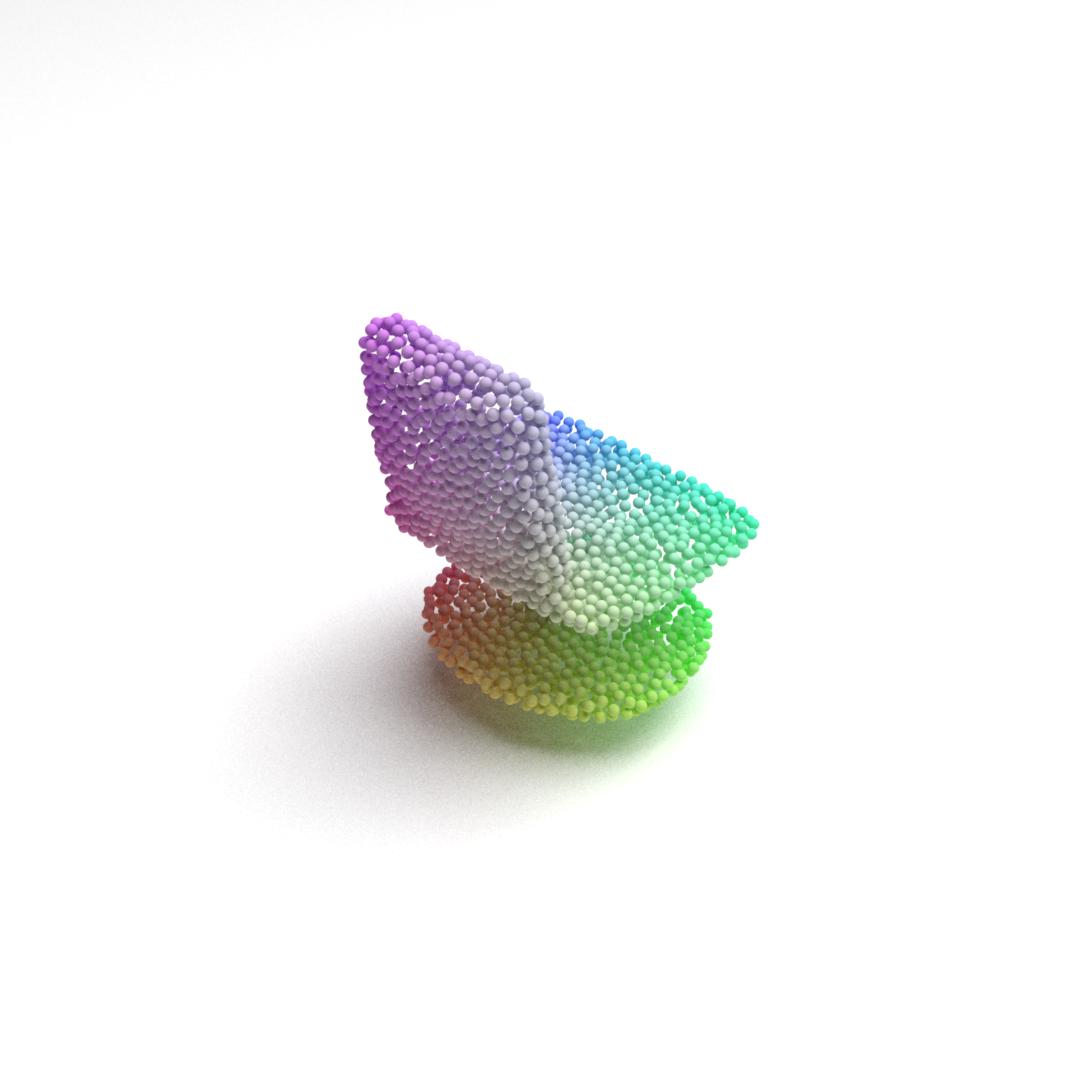}}{$[$\textcolor{red}{0.72}, \textcolor{teal}{0.005}$]$}
\\ 
\rotatebox{90}{\hspace{14pt}\emph{The stretcher}}
\rotatebox{90}{\hspace{16pt}\emph{is lower}}
\jsubfig{\includegraphics[height=2.5cm, trim={5cm 5cm 5cm 5cm}, clip]{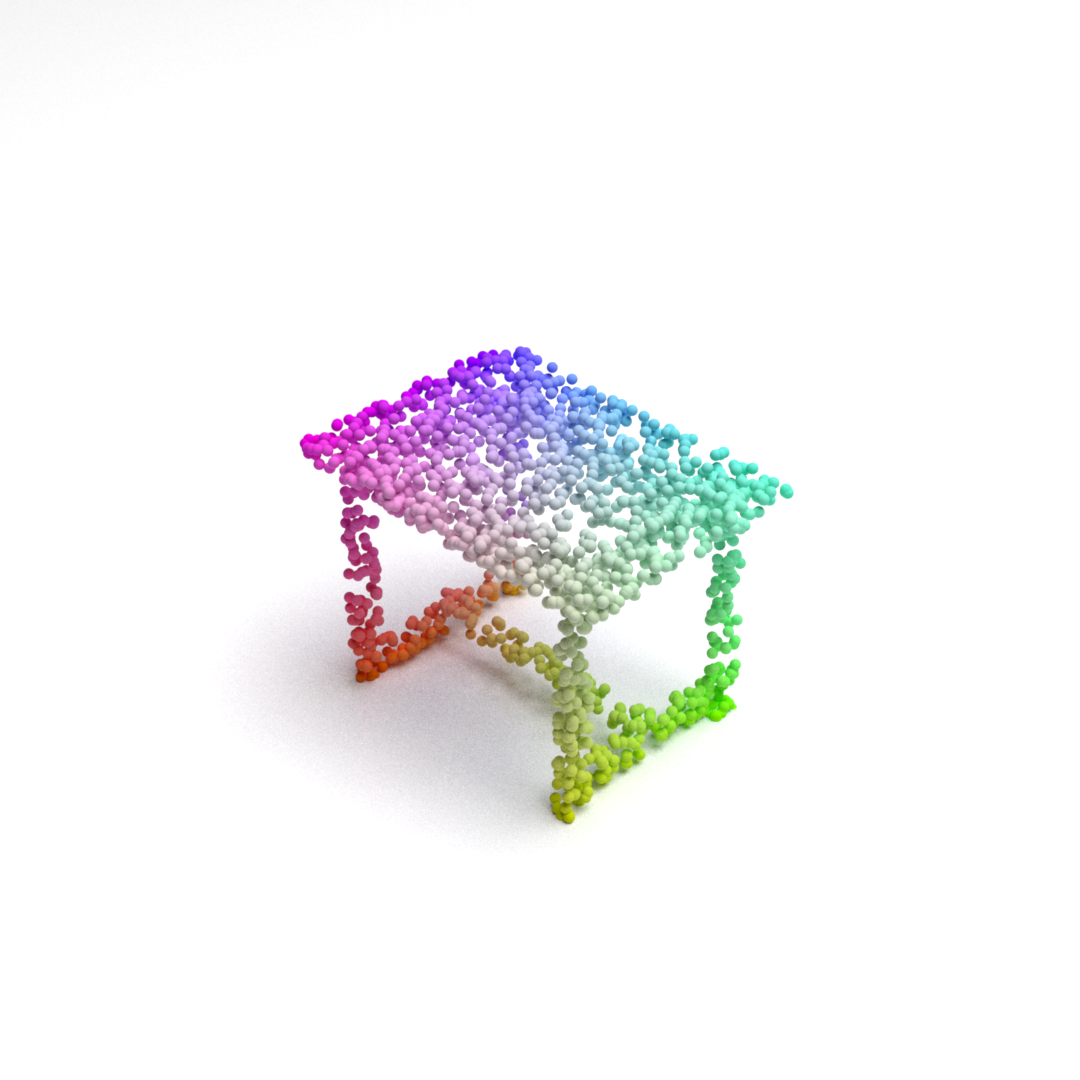}}{}
\jsubfig{\includegraphics[height=2.4cm, trim={5cm 5cm 5cm 7cm}, clip]{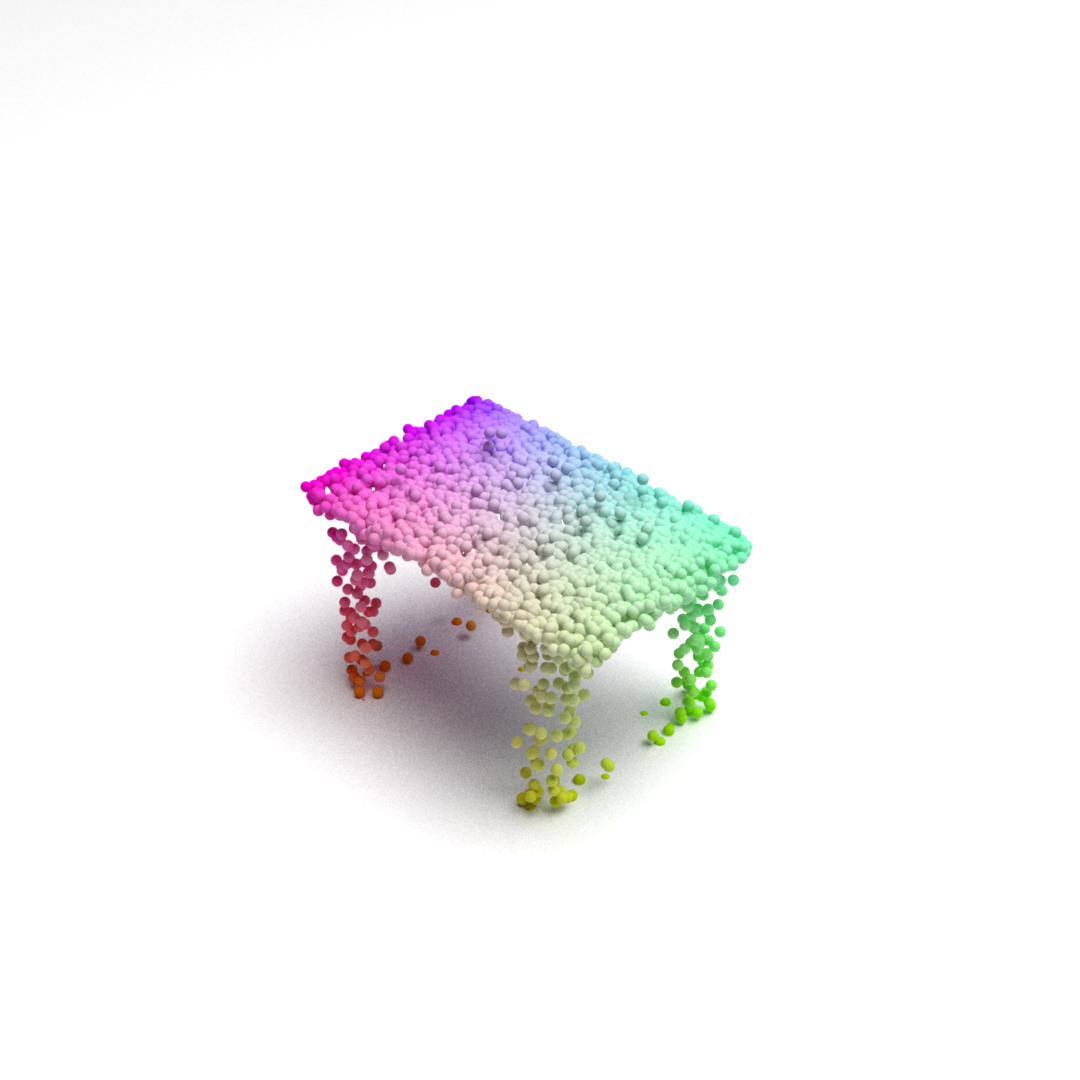}}{\hspace{-10pt}$[$\textcolor{red}{0.01}, \textcolor{teal}{0.003} $]$}
\jsubfig{\includegraphics[height=2.5cm, trim={5cm 5cm 5cm 5cm}, clip]{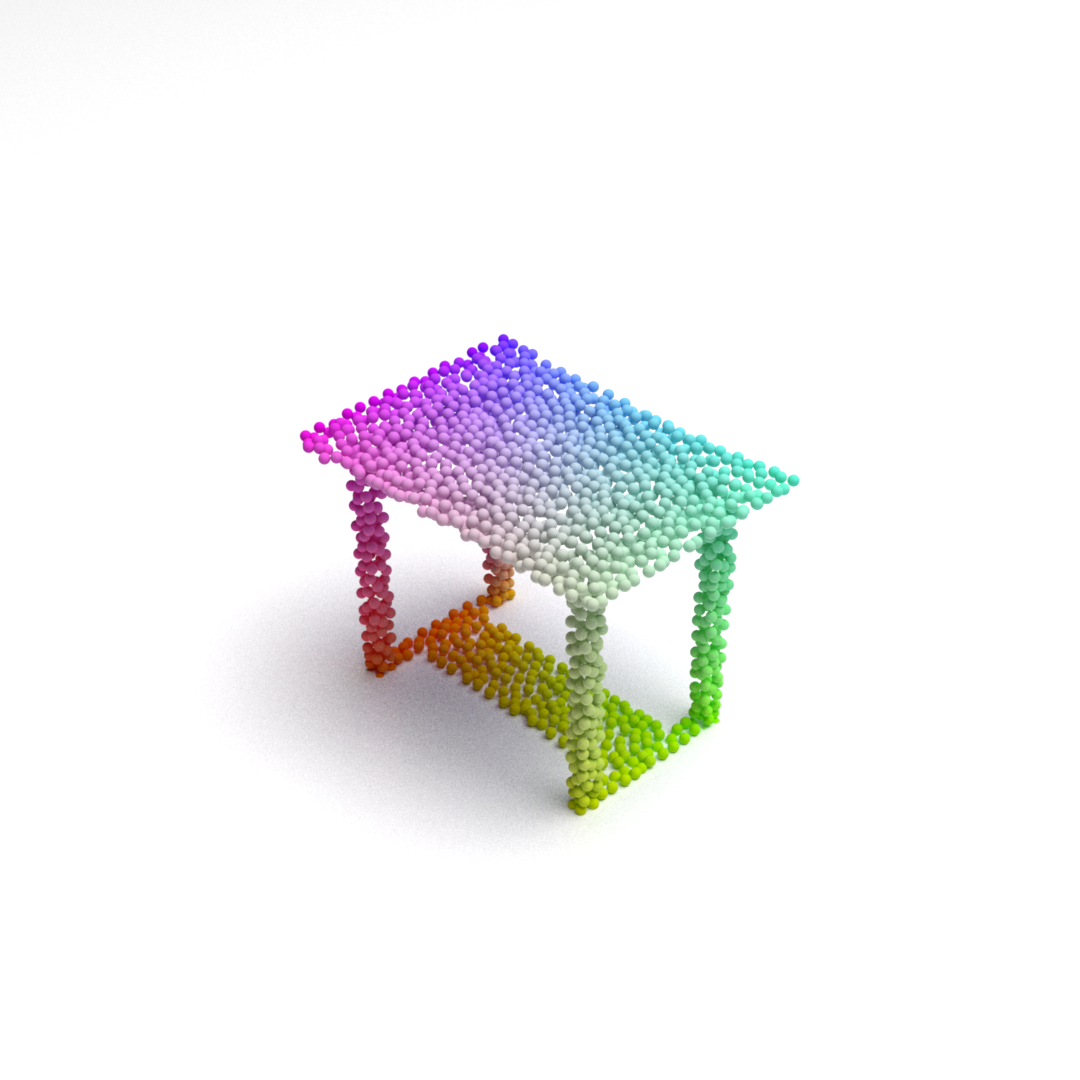}}{$[$\textcolor{red}{0.40}, \textcolor{teal}{0.004}$]$}
\\
\rotatebox{90}{\hspace{5pt}\emph{The shade hangs}}
\rotatebox{90}{\hspace{7pt}\emph{down lower}}
\jsubfig{\includegraphics[height=2.5cm, trim={7cm 7cm 7cm 7cm}, clip]{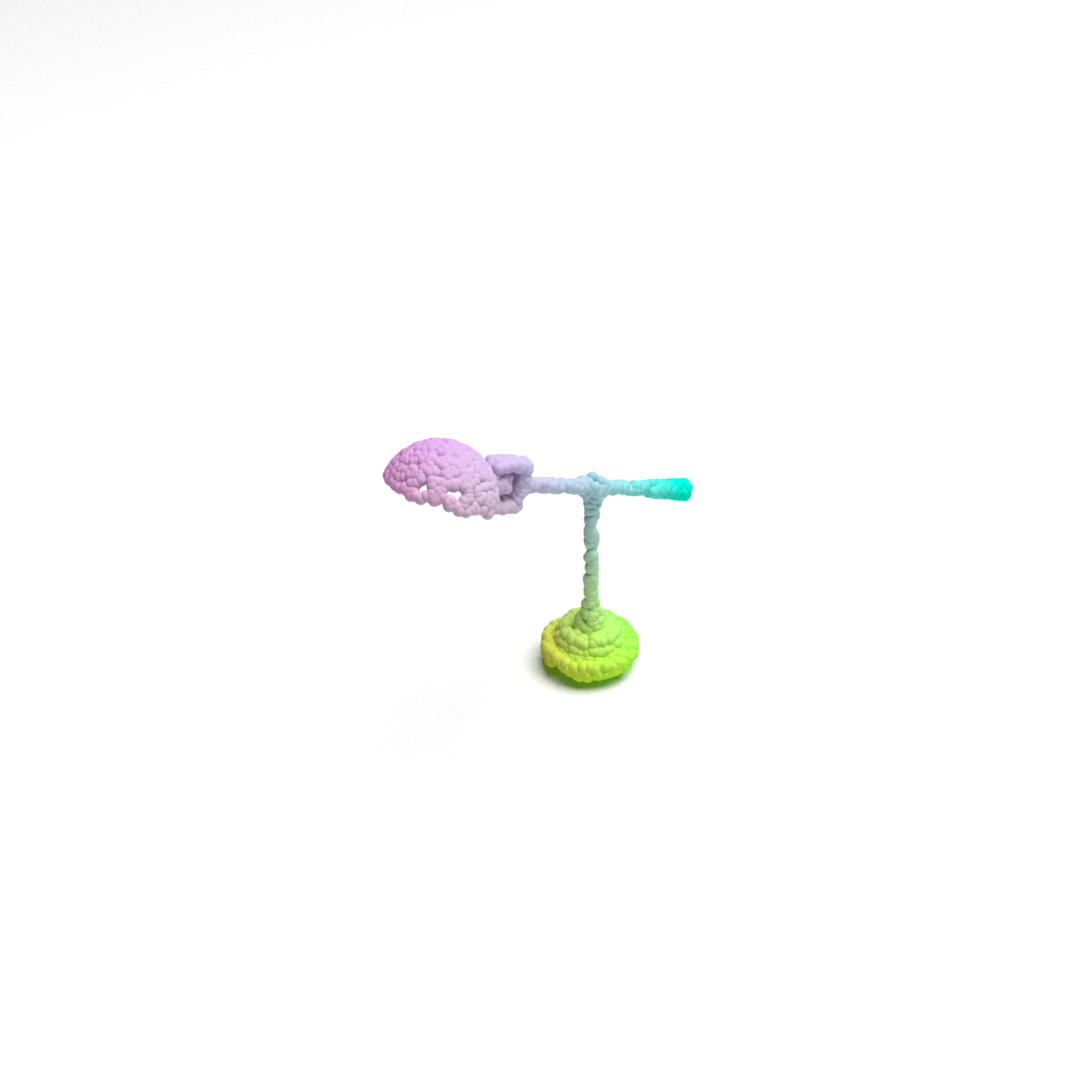}}{}
\jsubfig{\includegraphics[height=2.5cm, trim={6cm 6cm 6cm 6cm}, clip]{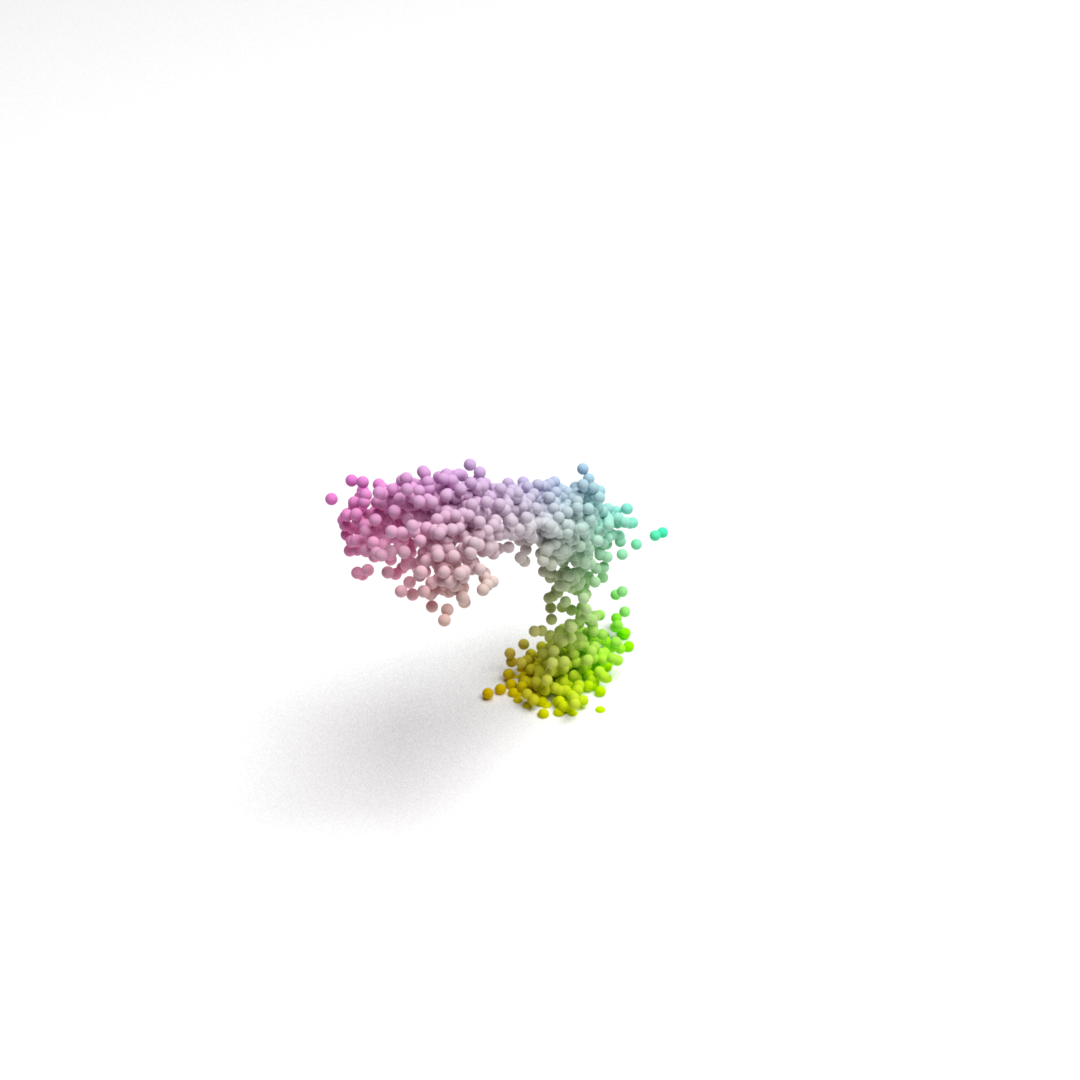}}{\hspace{-10pt}$[$\textcolor{red}{0.60}, \textcolor{teal}{0.003} $]$}
\jsubfig{\includegraphics[height=2.5cm, trim={5cm 5cm 5cm 5cm}, clip]{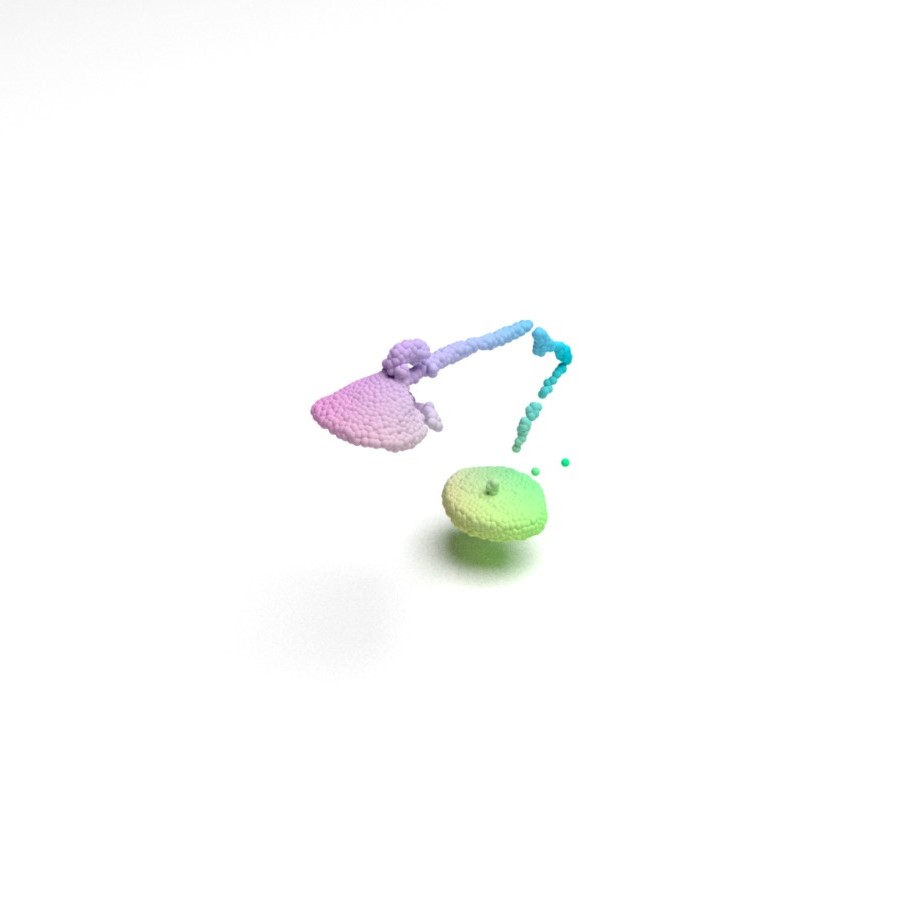}}{$[$\textcolor{red}{0.72}, \textcolor{teal}{0.013}$]$}
\\ 
\rotatebox{90}{\hspace{12pt}\emph{Target has}}
\rotatebox{90}{\hspace{13pt}\emph{less length}}
\jsubfig{\includegraphics[height=2.2cm, trim={5cm 5cm 5cm 7cm}, clip]{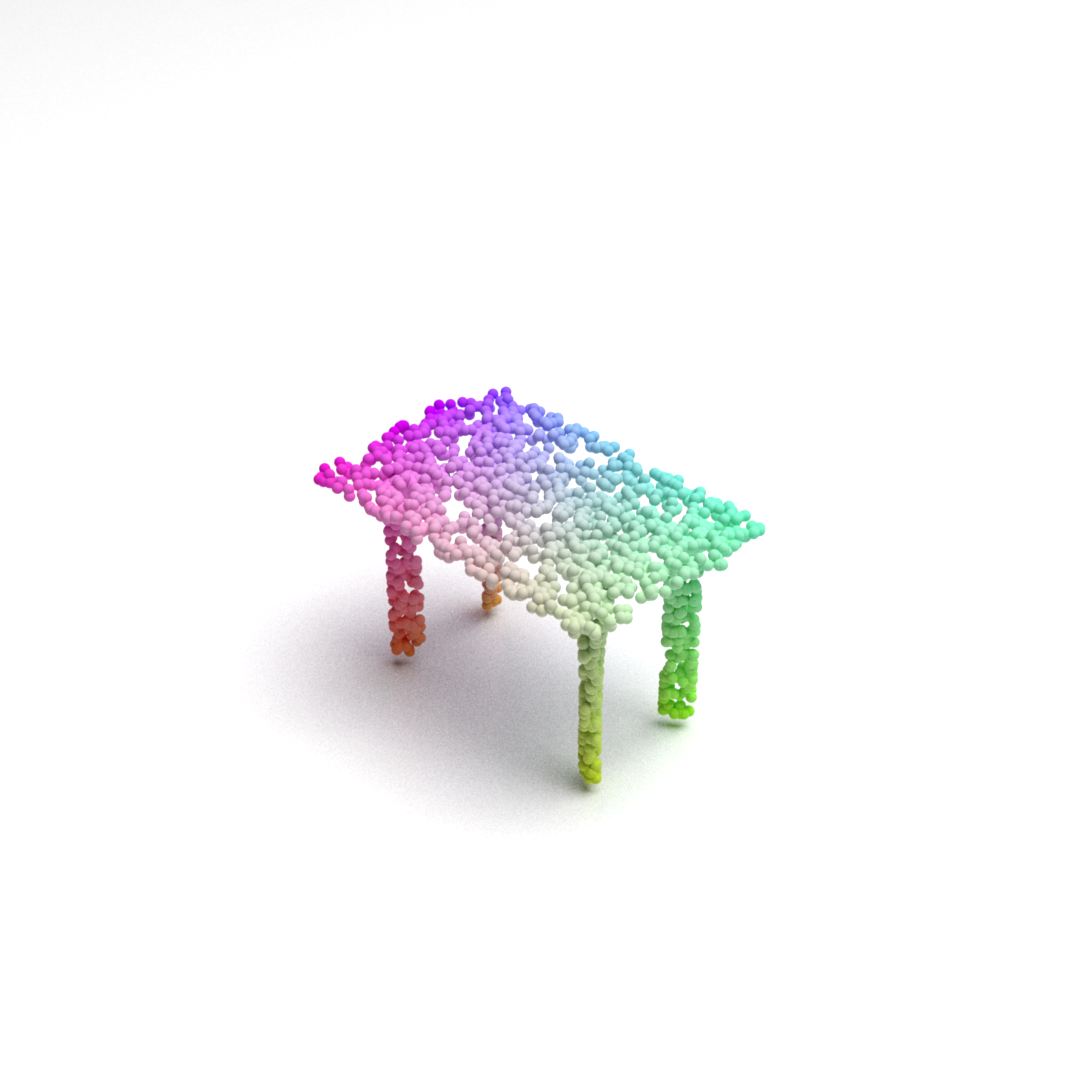}}{\whitetxt{x}\\Input}
\jsubfig{\includegraphics[height=2.2cm, trim={5cm 5cm 5cm 7cm}, clip]{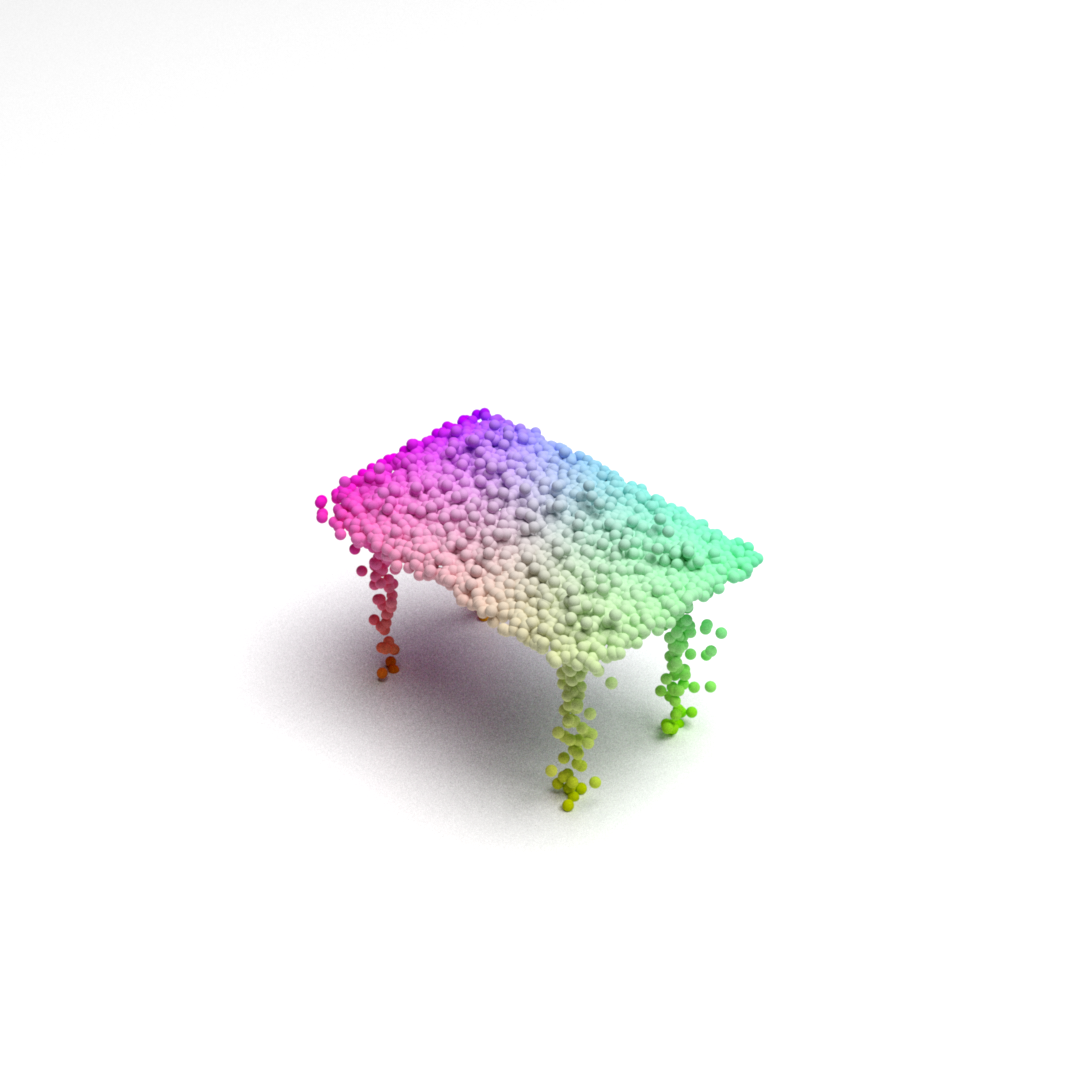}}{\hspace{-12pt}$[$\textcolor{red}{0.21} \textcolor{teal}{0.001}$]$\\\hspace{-12pt}ChangeIt3D}
\jsubfig{\includegraphics[height=2.2cm, trim={5cm 5cm 5cm 7cm}, clip]{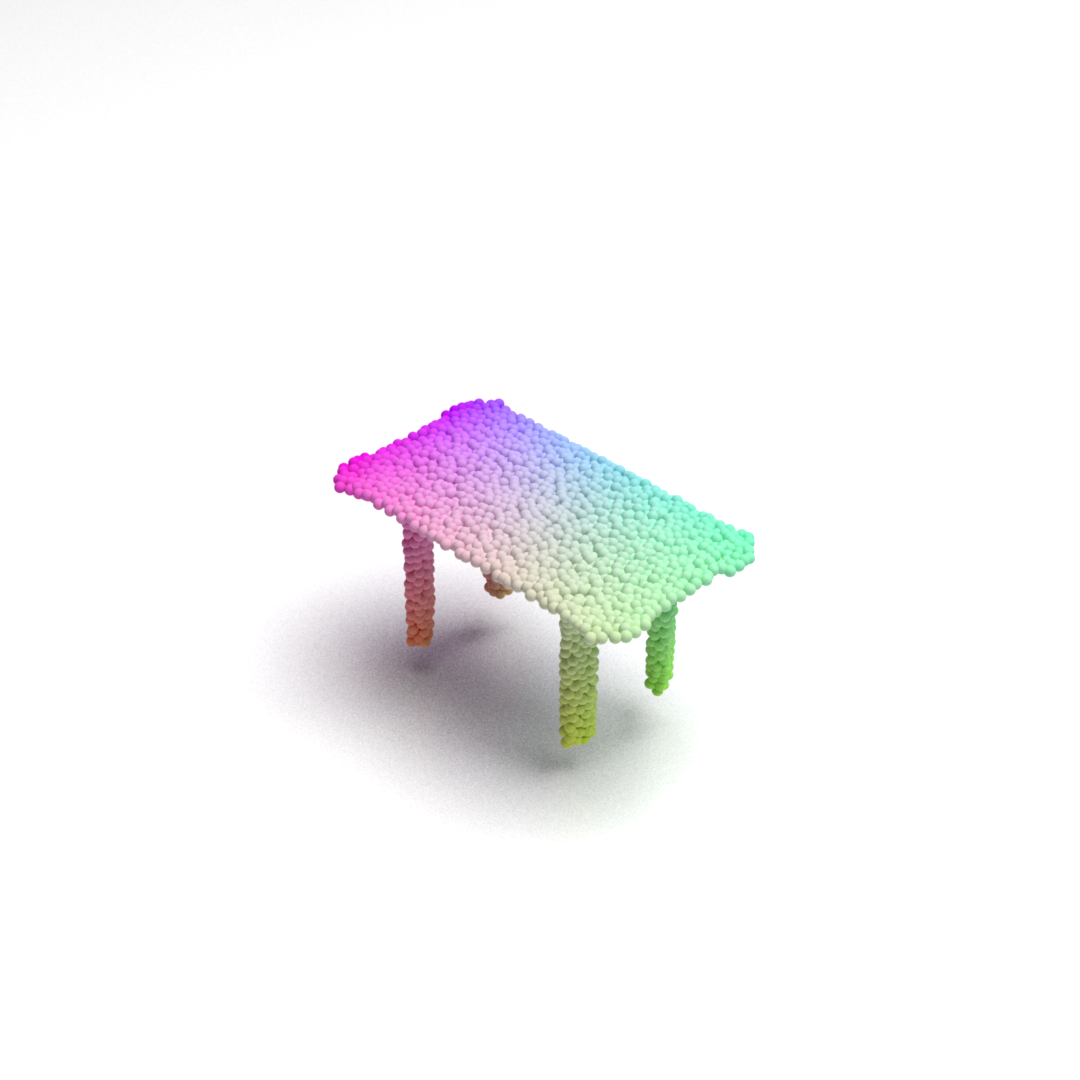}}{$[$\textcolor{red}{0.65} \textcolor{teal}{0.004}$]$\\Ours}
\vspace{-1pt}
\caption{\textbf{Semantic Shape Editing Comparison}. Above we show several examples along with their associated \textcolor{red}{LAB} and \textcolor{teal}{GD} (chamfer Distance) metrics. As illustrated above, ChangeIt3D (middle row) often fails to modify the input shape (left column), yielding smaller Chamfer distances, which do not necessarily correlate to better performance. Higher LAB scores better capture to what extent the shape agree with the target prompt. } 

\label{fig:lab_cd_issues}
\end{figure}

\begin{figure*} %
\centering
\rotatebox{90}{\hspace{12pt}\footnotesize{\emph{\bfseries The back}}}
\rotatebox{90}{\hspace{8pt}\footnotesize{\emph{ \bfseries is rectangle}}}
\hfill\jsubfig{\includegraphics[height=1.9cm]{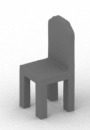}}{}
\jsubfig{\includegraphics[height=1.9cm]{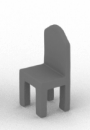}}{} 
\jsubfig{\includegraphics[height=1.9cm]{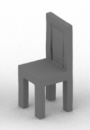}}
{}
\hfill
\hfill
\jsubfig{\includegraphics[height=1.9cm,trim={6.5cm 6.5cm 6.5cm 6.5cm},clip]{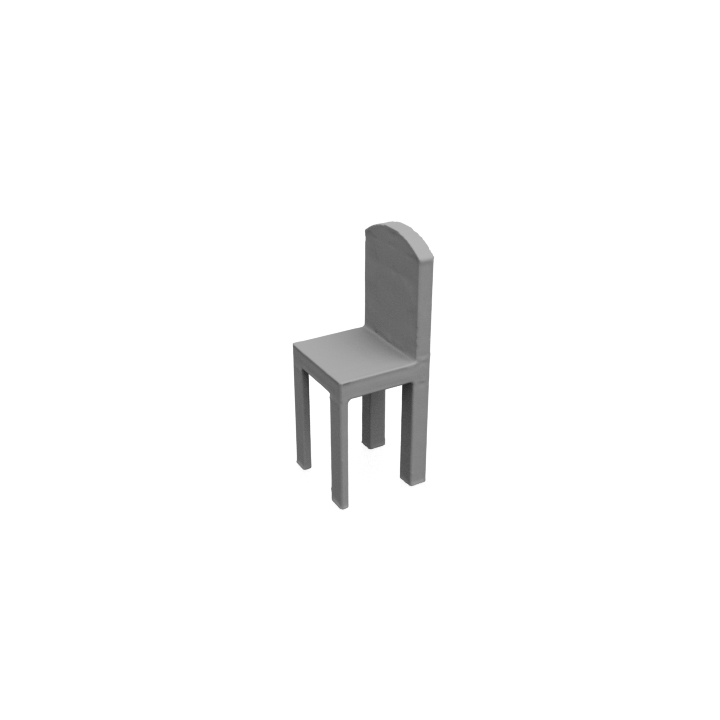}}
{}
\jsubfig{\includegraphics[height=1.9cm,trim={6.5cm 6.5cm 6.5cm 6.5cm},clip]{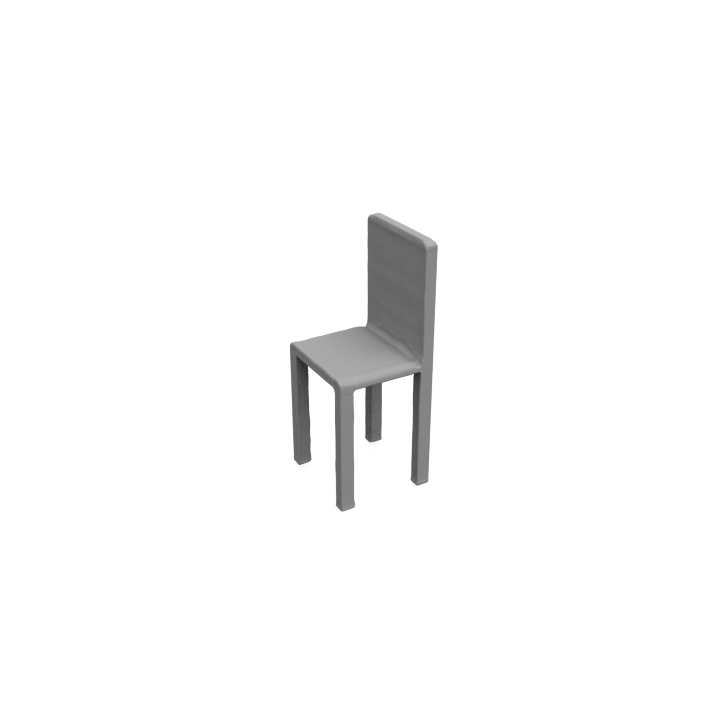}}{}
\hfill
\jsubfig{\includegraphics[height=1.9cm,trim={6.5cm 6.5cm 6.5cm 6.5cm},clip]{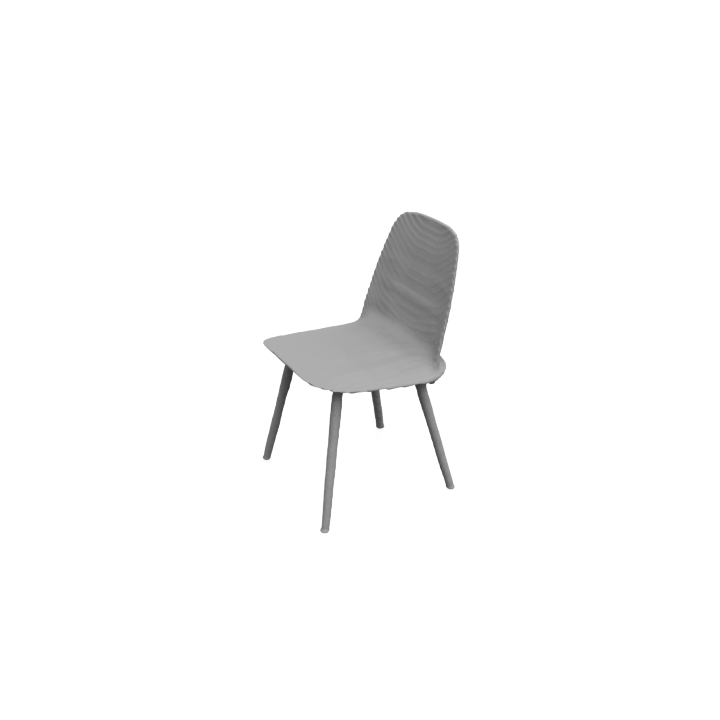}}{}
\jsubfig{\includegraphics[height=1.9cm,trim={6.5cm 6.5cm 6.5cm 6.5cm},clip]{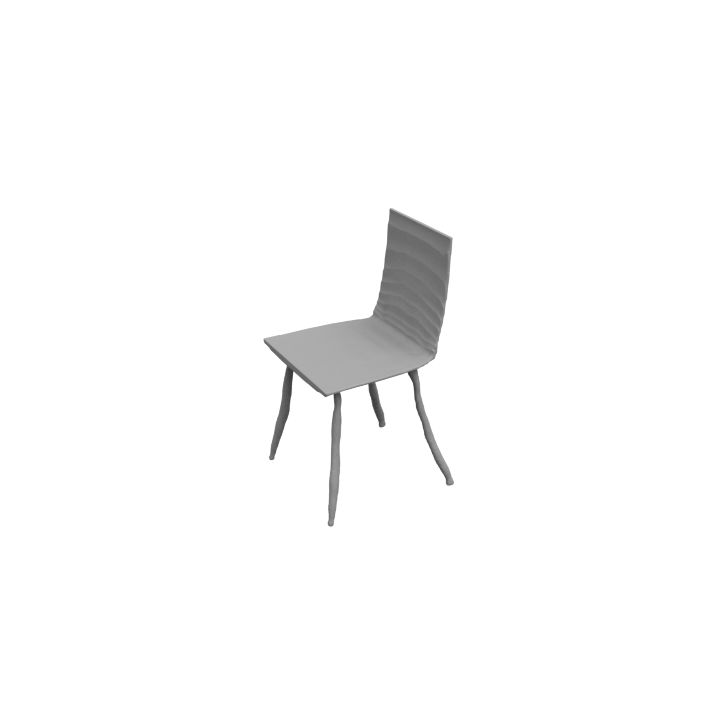}}{}
\hfill
\jsubfig{\includegraphics[height=1.9cm,trim={6.5cm 6.5cm 6.5cm 6.5cm},clip]{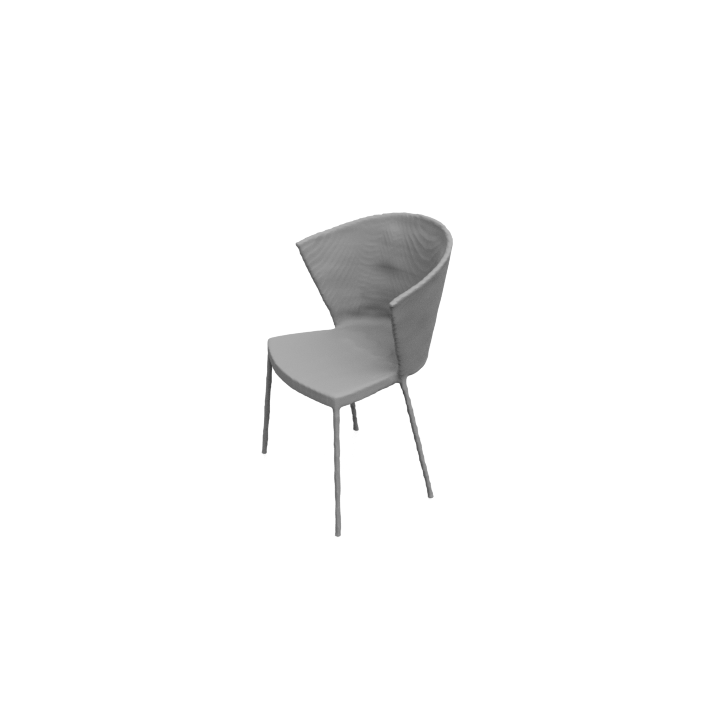}}{}
\jsubfig{\includegraphics[height=1.9cm,trim={6.5cm 6.5cm 6.5cm 6.5cm},clip]{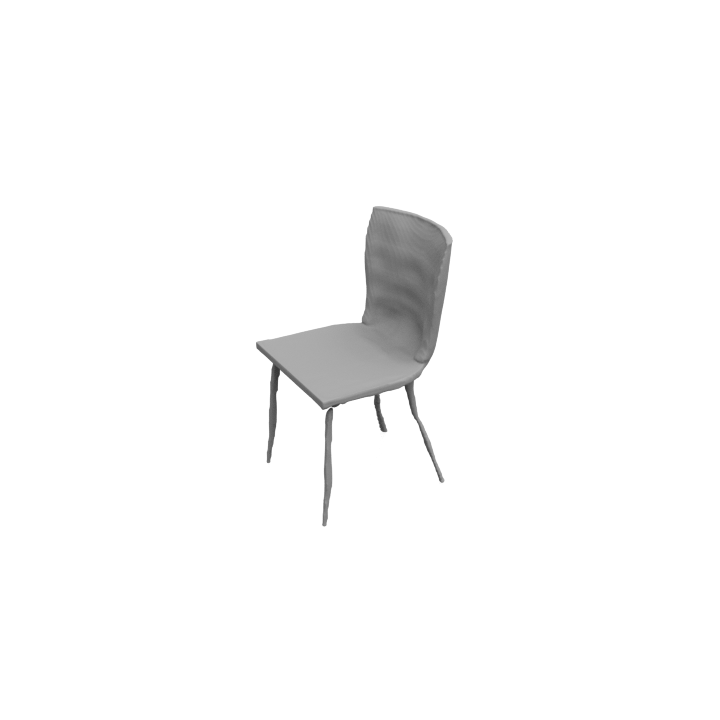}}{}
\\
\rotatebox{90}{\hspace{10pt}\footnotesize{\emph{\bfseries The target's}}}
\rotatebox{90}{\hspace{2pt}\footnotesize{\emph{\bfseries seat is less thick}}}
\hfill
\jsubfig{\includegraphics[height=1.9cm]{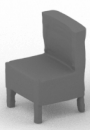}}{\footnotesize {}}
\jsubfig{\includegraphics[height=1.9cm]{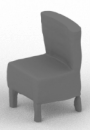}}{\footnotesize {}} 
\jsubfig{\includegraphics[height=1.9cm]{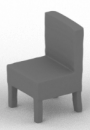}}
{\footnotesize {}}
\hfill
\hfill\jsubfig{\includegraphics[height=1.9cm,trim={6.5cm 6.5cm 6.5cm 6.5cm},clip]{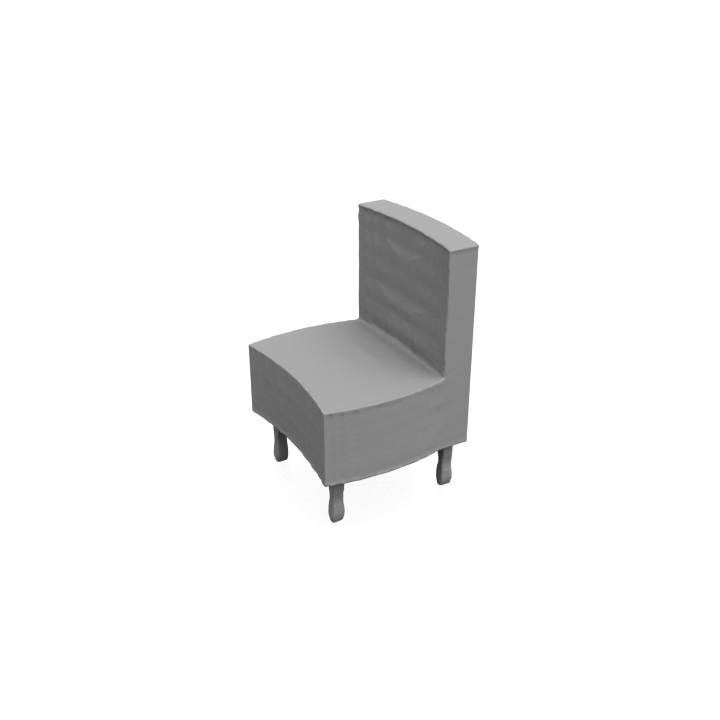}}
{\footnotesize {}}
\jsubfig{\includegraphics[height=1.9cm,trim={6.5cm 6.5cm 6.5cm 6.5cm},clip]{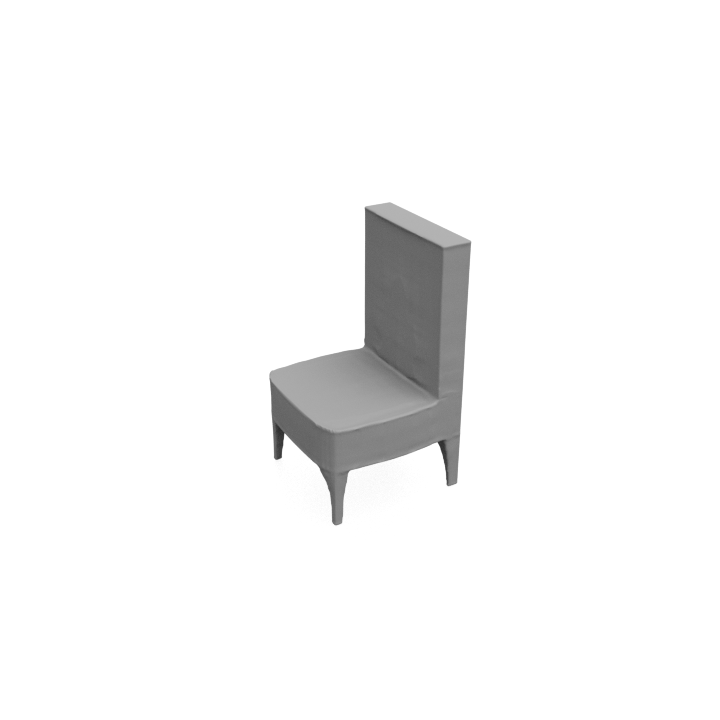}}{\footnotesize {}}
\hfill
\jsubfig{\includegraphics[height=1.9cm,trim={6.5cm 6.5cm 6.5cm 6.5cm},clip]{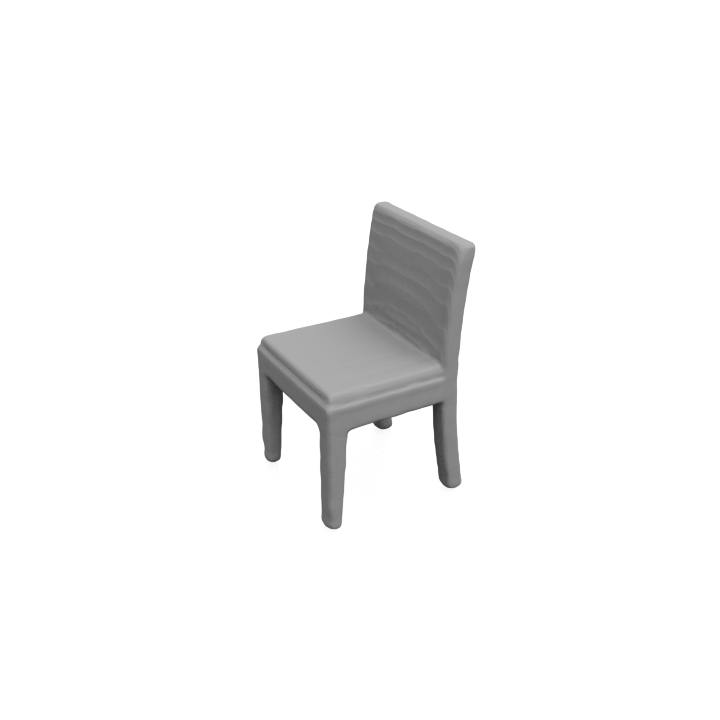}}
{\footnotesize {}}
\jsubfig{\includegraphics[height=1.9cm,trim={6.5cm 6.5cm 6.5cm 6.5cm},clip]{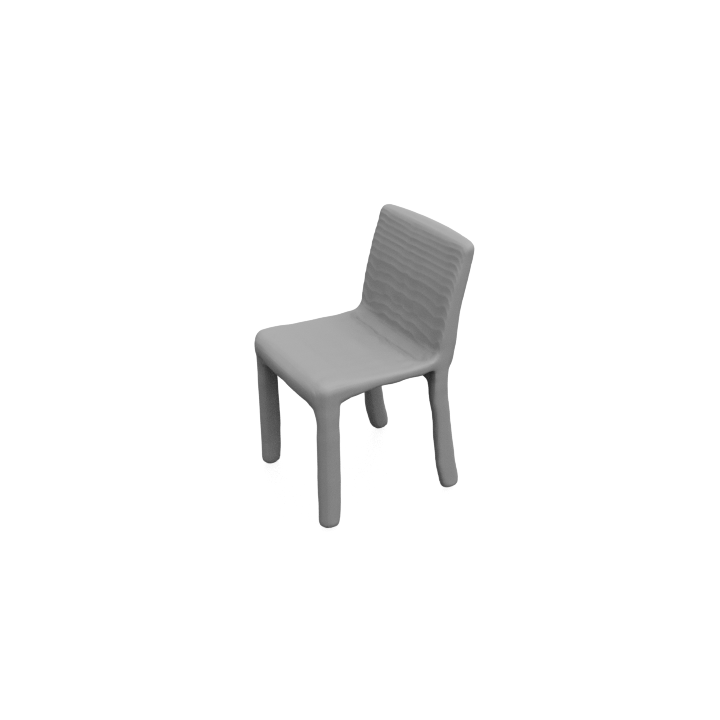}}{\footnotesize {}}
\hfill
\jsubfig{\includegraphics[height=1.9cm,trim={6.5cm 6.5cm 6.5cm 6.5cm},clip]{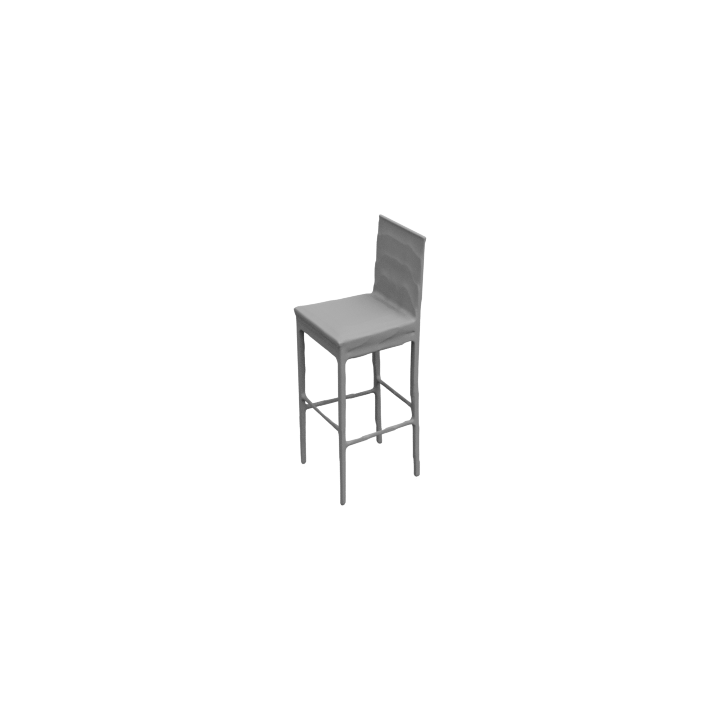}}
{\footnotesize {}}
\jsubfig{\includegraphics[height=1.9cm,trim={6.5cm 6.5cm 6.5cm 6.5cm},clip]{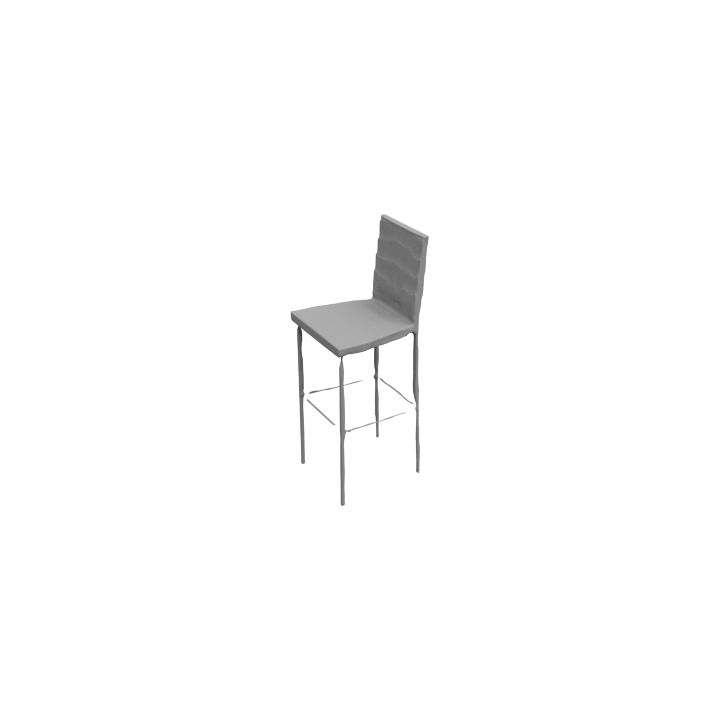}}{\footnotesize {}}
\\
\rotatebox{90}{\hspace{5pt}\footnotesize{\emph{\bfseries The target has}}}
\rotatebox{90}{\footnotesize{\emph{\bfseries smaller arm rests}}}
\hfill
\jsubfig{\includegraphics[height=1.9cm]{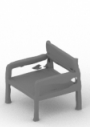}}{\footnotesize {Input}}
\jsubfig{\includegraphics[height=1.9cm]{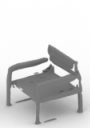}}{\footnotesize {ChangeIt3D}} 
\jsubfig{\includegraphics[height=1.9cm]{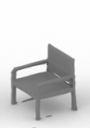}}
{\footnotesize {LADIS}}
\hfill
\hfill\jsubfig{\includegraphics[height=1.9cm,trim={6.5cm 6.5cm 6.5cm 6.5cm},clip]{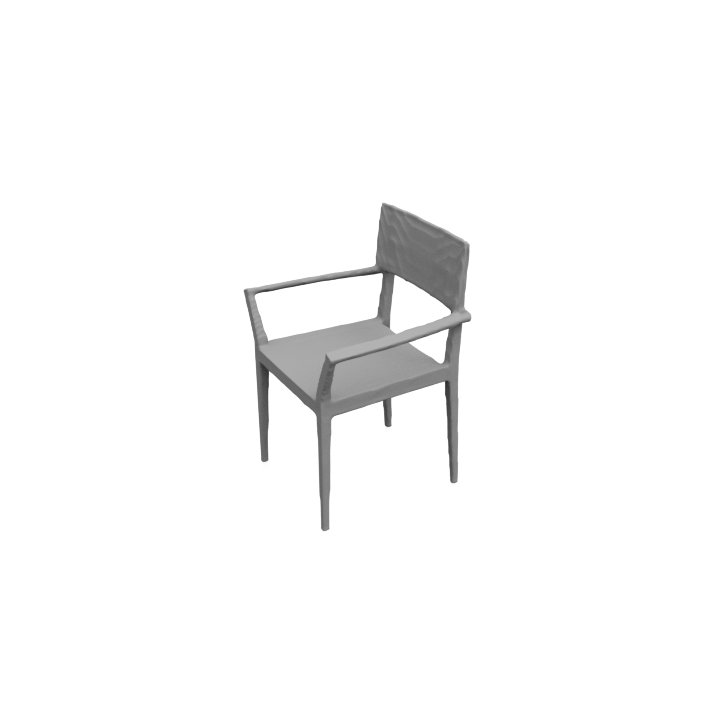}}
{\footnotesize {Guidance}}
\jsubfig{\includegraphics[height=1.9cm,trim={6.5cm 6.5cm 6.5cm 6.5cm},clip]{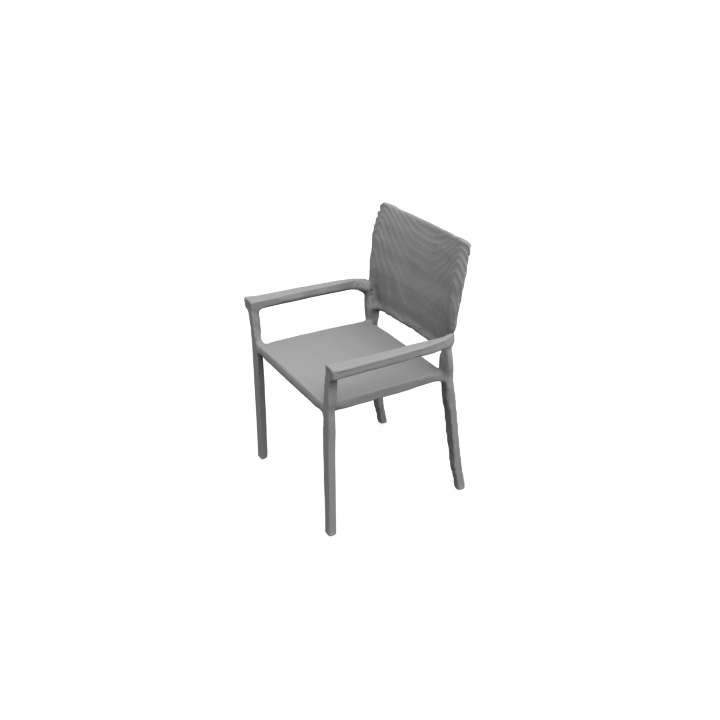}}{\footnotesize {Ours}}
\hfill
\jsubfig{\includegraphics[height=1.9cm,trim={5.5cm 5.5cm 6.5cm 6.5cm},clip]{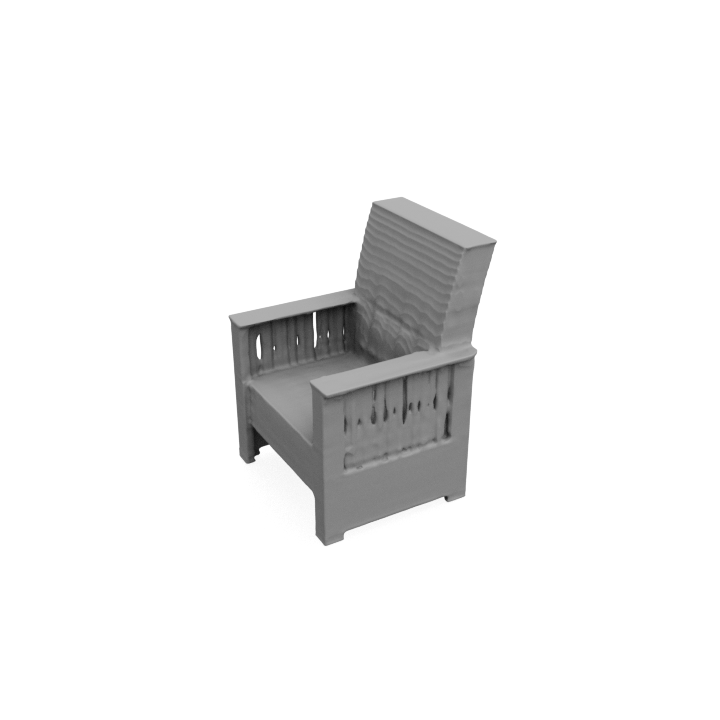}}
{\footnotesize {Guidance}}
\jsubfig{\includegraphics[height=1.9cm,trim={5.5cm 5.5cm 6.5cm 6.5cm},clip]{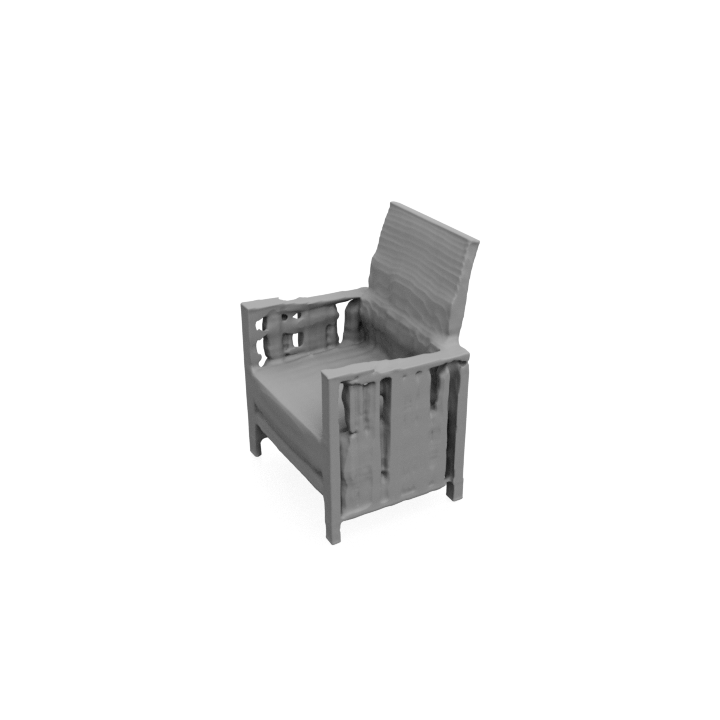}}{\footnotesize {Ours}}
\hfill
\jsubfig{\includegraphics[height=1.9cm,trim={6.5cm 6.5cm 6.5cm 6.5cm},clip]{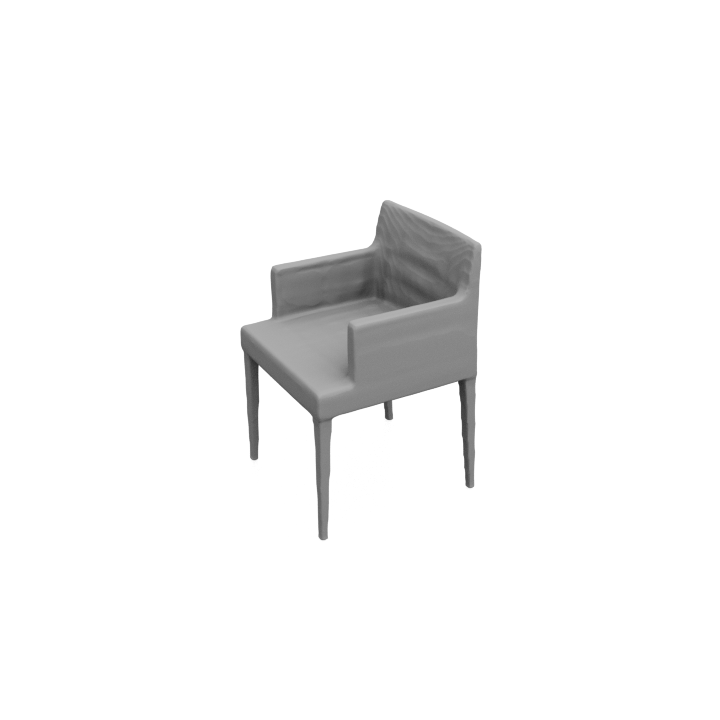}}
{\footnotesize {Guidance}}
\jsubfig{\includegraphics[height=1.9cm,trim={6.5cm 6.5cm 6.5cm 6.5cm},clip]{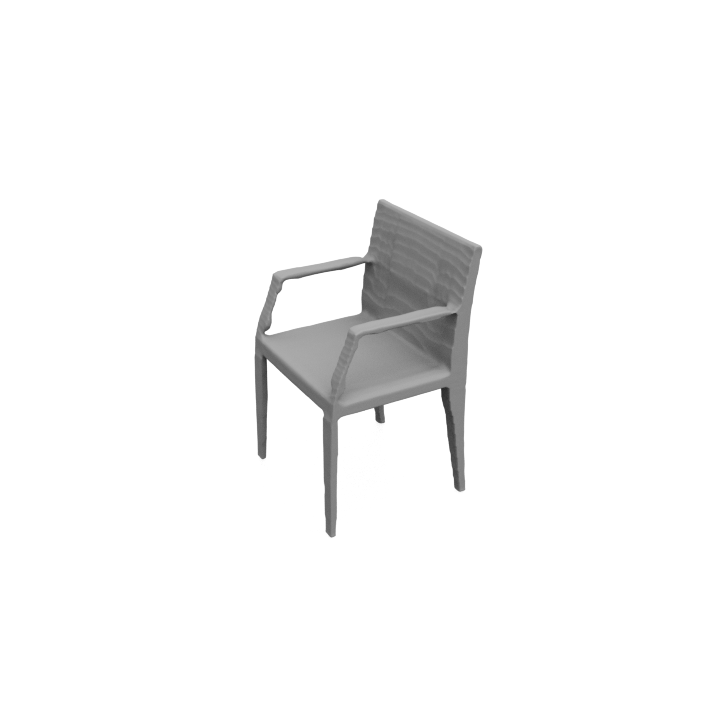}}{\footnotesize {Ours}}
\vspace{5pt} 
\caption{\textbf{Semantic Shape Editing Comparison}. We compare to the qualitative results shown in Huang \etal~\cite{huang2022ladis}, comparing LADIS with ChangeIt3D. 
While we could not locate the exact input shapes used in their paper, we compare to a similar shape from the test set, and two additional test shapes (on the right). 
As illustrated above, our model allows for significant object edits respecting the target text prompt, yielding high-quality outputs that are structurally-faithful to the guidance shape.
}
\label{fig:comparisons_shapetalk}
\end{figure*}

\subsection{Semantic Shape Editing with ChangeIt3D and LADIS}
We perform a qualitative comparison with LADIS~\cite{huang2022ladis} in Figure \ref{fig:comparisons_shapetalk} by showing our results over examples from Figure 5 in their paper. We perform this comparison over a similar guidance shape, and over two additional test shapes, as we do not have access to the exact shapes used in their figure. As illustrated in the figure, our technique can perform such edits over diverse input chairs, yielding high-quality output chairs with regions unrelated to the target text prompt largely resembling the input guidance shape. As also observed in our experiments, ChangeIt3D does not perform significant edits, yielding outputs that mostly reconstruct the inputs. LADIS indeed allows for performing local shape manipulations, but the edits may be not be as clear as the edits enabled by our method (for example, see the seat thickness in the middle row) or may cause undesirable modifications (such as the added holes in the top row).

We show additional side-by-side comparisons with ChangeIt3D, along with the associated LAB and GD (\emph{i.e.}, Chamfer distance) scores, in Figure \ref{fig:lab_cd_issues}. In these provided examples, our method achieves a worse GD score and a better LAB score, in comparison to ChangeIt3D. As illustrated in the figure (and also described in the main paper), better GD scores are often a result of unmodified shapes, whereas the LAB metric is more indicative of meaningful edits.

\begin{figure} %
\centering
\jsubfig{\includegraphics[height=2cm, trim={2cm 2cm 2cm 2cm}, clip]{images/results/stylization/cond_bottle.png}}{\footnotesize{{Guidance}}} 
\jsubfig{\includegraphics[height=2cm, trim={1cm 0cm 1cm 1cm}, clip]{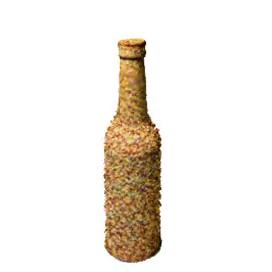}}{\footnotesize{\emph{A corked bottle}}}
\hfill
\hfill
\jsubfig{\includegraphics[height=2cm, trim={1cm 0cm 1cm 0.5cm}, clip]{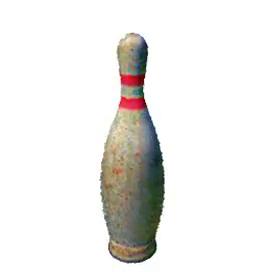}}{\footnotesize{\emph{A bowling pin}}}
\hfill
\jsubfig{\includegraphics[height=2cm, trim={1cm 0cm 1cm 1cm}, clip]{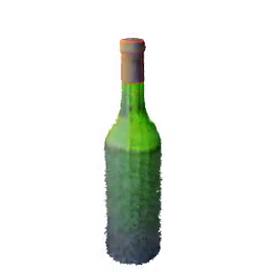}}{\footnotesize{\emph{A wine bottle}}}
\hfill
\vspace{1.5pt}
\\
\jsubfig{\includegraphics[height=1.9cm, trim={2cm 2cm 2cm 2cm}, clip]{images/results/stylization/cond_bowl.png}}{\footnotesize {Guidance}}
\hfill
\jsubfig{\includegraphics[height=1.9cm, trim={1cm 0.5cm 1cm 2cm}, clip]{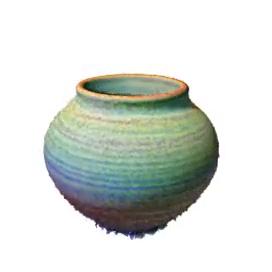}}{\footnotesize{\emph{A modern vase}}}
\hfill
\jsubfig{\includegraphics[height=1.9cm, trim={1cm 0.5cm 1cm 2cm}, clip]{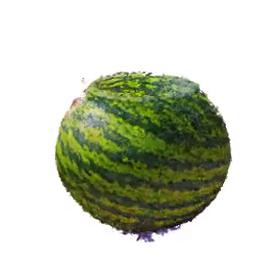}}{\footnotesize{\emph{A watermelon}}}
\hfill
\jsubfig{\includegraphics[height=1.9cm, trim={1cm 0.5cm 1cm 2cm}, clip]{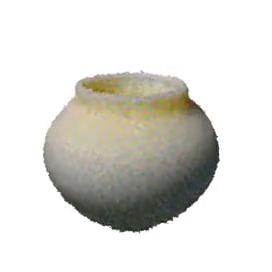}}{\footnotesize{\emph{A candle}}}
\vspace{-5pt}
\caption{\textbf{3D Stylization Comparison.} We compare our results to the ones obtained with Vox-E~\cite{sella2023vox}, a recent optimization-based technique performing text-guided editing of 3D objects. As illustrated above, Vox-E's outputs contain spurious noisy artifacts, possibly due to its volumetric regularization loss that aims at preserving the input 3D object. Our results (illustrated in Figure 8 in the main paper) better convey the target text prompts, while being orders of magnitude faster (20 seconds vs. 50 minutes per edit).}
\label{fig:voxe_comparison_}
\end{figure}

\subsection{3D Stylization with Vox-E}
We perform an additional comparison with Vox-E~\cite{sella2023vox}, a recent optimization-based method proposed for performing text-guided editing of 3D objects. Qualitative results of their method over the examples illustrated in Figure 8 in the main paper are shown in Figure \ref{fig:voxe_comparison_}. As illustrated in the figure, Vox-E's outputs contain spurious noisy artifacts, possibly due to its volumetric regularization loss which aims at preserving the input 3D object. Furthermore, we perform an experiment to quantify Vox-E's performance. As this method operates over 2D images, we randomly select ten 3D assets from the test test and manually convert these assets into a set of rendered views using Blender. 
Quantitatively, Vox-E yields $\text{CLIP}_{Sim}=0.28$ and $\text{CLIP}_{Dir}=0.01$,   comparable to our method, while being significantly slower (approximately 50 minutes per edit, in contrast to our 20 seconds inference time). 

\subsection{User Study}
\label{sec:user}

We conduct several blind perceptual studies to quantify the user's preference, comparing our results against various baselines. We perform separate studies for each task, comparing our performance to the strongest competing baseline (Table \ref{tab:semantic_editing_user_study}). Specifically, the 3D stylization and the text-conditional abstraction-to-3D models are tested against Fantasia3D, and the semantic shape editing model is tested against ChangeIt3D. 
We also compare against the ControlNet3D ablation method over the text-conditional abstraction-to-3D task (Table \ref{tab:abstraction_user_study}).

Each survey is composed of two sets of questions (Form 1 and 2), with participants first selecting the set of questions by clicking on their associated links. In all studies, users answered 10-30 questions, where each question is composed of a guidance shape, a text prompt, and two outputs (ours and an output produced by a competing baseline). Users were asked to select the output that they preferred according to a given criteria. In all surveys, the guidance shapes and prompts were chosen at random from the test sets for each task. In the text-conditional abstraction-to-3D and the semantic shape editing surveys we also made sure there was a balanced split of questions for each shape category (\emph{chair}, \emph{table}, etc.). Details regarding how we conducted each survey as well as discussions regarding their results are provided below.

\begin{table}[t]
\setlength{\tabcolsep}{3.0pt}
 \def\arraystretch{1.05}
\centering
\resizebox{0.99\linewidth}{!}{
\begin{tabular}{lcc}
\toprule
   \multirow{2}{*}{\footnotesize{Task}}  & \footnotesize{Competing}  & \footnotesize{\methodName{}} \\  
    & \footnotesize{Baseline} & \footnotesize{Preference} \\
    \midrule 
    \footnotesize{Semantic Shape Editing} & \footnotesize{ChangeIt3D} & \footnotesize{70.00\%} \\
    \footnotesize{3D Stylization} & \footnotesize{Fantasia3D} & \footnotesize{64.83\%} \\
    \footnotesize{Text-conditional Abstraction-to-3D} & \footnotesize{Fantasia3D} & \footnotesize{71.03\%} \\
\bottomrule
\end{tabular}
}
\caption{\textbf{\methodName{} user preference over competing baselines.} The results in this table indicate the percentage of users preferring our results over competing baselines across all tasks. See Section \ref{sec:user} for more details.}
\label{tab:semantic_editing_user_study}
\end{table}

\medskip
\noindent 

\textbf{Semantic Shape Editing Task (Comparison to ChangeIt3D).} 
The instructions for this survey were as follows: \emph{``Select the shape that better fits the target text while preserving the guidance shape as best as possible. If both shapes fail / succeed to match the text, choose the one that best preserves the input shape.''} In total, 47 users participated in this study (with a 19/28 splits across the forms), and each survey form was composed of 10 questions. The shapes and accompanying prompts were taken at random from the ShapeTalk \cite{achlioptas2022changeit3d} test set. Results are reported in Table \ref{tab:semantic_editing_user_study} (top row). As illustrated in the table, 70\% of the times users preferred our results.

\medskip
\noindent 
\textbf{Stylization and Text-conditional
Abstraction-to-3D  (Comparison to Fantasia3D).} We used the same instructions as in the previously detailed study. Thirty users participated in this study (with a 14/16 splits across the forms), and each survey form was composed of 20 questions (split evenly between the two tasks), depicting shapes and accompanying prompts randomly selected from the associated test sets. 
Results are reported in Table \ref{tab:semantic_editing_user_study} (bottom two rows). As illustrated in the table, users preferred our results 64.83\% of the time in the 3D stylization task and 71.03\% in the text-conditional abstraction-to-3D task.

\begin{table}[t]
\setlength{\tabcolsep}{3.0pt}
 \def\arraystretch{1.1}
\centering
\resizebox{\linewidth}{!}{
\begin{tabular}{lcccc}
\toprule
   Metric  & Chair & Table & Airplane & Overall \\ 
    \midrule 
    Conveys Text Prompt & 65.00\% & 53.16\% & 63.64\% & 60.24\% \\
    Faithful to Guidance Shape & 75.84\% & 70.47\% & 58.52\% & 67.72\%  \\
    Visually Pleasing & 67.91\% & 73.56\% & 79.58\% & 74.78\%  \\
\bottomrule
\end{tabular}
}
\caption{\textbf{\methodName{} user preference rates compared to ControlNet3D over the Abstraction-to-3D task.} The results in this table indicate the percentage of users preferring our results over different criteria (conveying the text prompt, preserving the geometry of the guidance shape or overall visual quality). See Section \ref{sec:user} for more details.}
\label{tab:abstraction_user_study}
\end{table}

\medskip
\noindent 

\textbf{Comparison to the ControlNet3D Ablation.} We compare against our ControlNet3D ablation method over random test samples from the text-conditional abstraction-to-3D task. To gain a more comprehensive understanding of how our method compares with ControlNet3D, users were asked to select their preferred shapes according to three different criterion: (i)  consistency with the target text prompt, (ii) faithfulness to the guidance shape, and (iii) visual appearance. The corresponding instructions were as follows: (i) \emph{Select the output which better conveys the text prompt}, (ii) \emph{Select the output which is more faithful to the structure of the guidance shape} and (iii) \emph{Select the output which is more visually pleasing}.
In total, 48 users participated in this study (with a 28/20 splits across the forms), and each was presented with 30 questions.

Results are reported in Table \ref{tab:abstraction_user_study}. As illustrated in the table, users more often preferred our results over the ControlNet3D baseline, when shown side-by-side examples,  across all criteria. Our results were especially preferred in terms of visual quality ($75\%$ of the time).
We believe the reason for this is that ControlNet3D is more prone to over-fitting to the dataset during training, causing its outputs to differ from the original \shapE{} output space. These results further illustrate that our attention mechanism enables more expressive 3D control, in comparison to a straightforward extension of ControlNet. Just as cross-attention layers in text-to-image models allow for connecting between each word token in the text and each pixel in an image, our system allows each layer of the guidance latent code to interact with each layer of the noisy latent propagating through the network at each block.

\begin{table}[t]
\setlength{\tabcolsep}{5.0pt}
 \def\arraystretch{1.1}
\centering
\resizebox{0.88\linewidth}{!}{
\begin{tabular}{llccccc}
\toprule
   & Method  & LAB$\uparrow$ & GD$\downarrow$ & $l$-GD $\downarrow$ & CD$\downarrow$ & FPD$\downarrow$ \\ 
    \midrule 
    \multirow{2}{*}{\rotatebox[origin=c]{90}{\emph{Chair}}} & ChangeIt3D & 0.24 & \textbf{0.003} & \textbf{0.009} & \textbf{0.01} & 22.9\\
    & Ours & \textbf{0.44} & 0.005 & 0.013 & 	0.04 &\textbf{22.5} \\
    \midrule 
    \multirow{2}{*}{\rotatebox[origin=c]{90}{\emph{Table}}} & ChangeIt3D & 0.30 & \textbf{0.003} & \textbf{0.009} & \textbf{0.03} & 32.6\\
    & Ours & \textbf{0.41} & 0.009 & 0.013 & 0.05 & \textbf{32.0} \\
    \midrule 
    \multirow{2}{*}{\rotatebox[origin=c]{90}{\emph{Lamp}}} & ChangeIt3D & 0.27 & \textbf{0.004} & 	\textbf{0.009} & 0.14 & 138.7 \\
    & Ours & \textbf{0.47} & 0.005 & 0.016 & \textbf{0.08} & \textbf{37.3} \\
    \midrule 
    \multirow{2}{*}{\rotatebox[origin=c]{90}{\emph{Avg.}}} & ChangeIt3D & 0.27 & \textbf{0.003} & \textbf{0.009} & \textbf{0.05} & 53.6 \\
    & Ours & \textbf{0.44} & 0.007 & 0.013 & \textbf{0.05} & \textbf{29.9} \\ 
\bottomrule
\end{tabular}
}
\caption{\textbf{Performance Breakdown for Semantic Shape Editing Task.  } We compare the performance of ChangeIt3D~\cite{achlioptas2022changeit3d} against ours over the three object categories in the semantic shape editing tasks.
}
\label{tab:editing_comparison_supp}
\end{table}

\begin{table}[t]
\setlength{\tabcolsep}{3.0pt}
 \def\arraystretch{1.1}
\centering
\resizebox{\linewidth}{!}{
\begin{tabular}{llcccc}
\toprule
   & Method  & $\text{CLIP}_{Sim}\uparrow$ & $\text{CLIP}_{Dir} \uparrow$ & $\text{GD} \downarrow$ & Run Time \\ 
    \midrule 
    \multirow{3}{*}{\rotatebox[origin=c]{90}{\emph{Chair}}} & SketchShape & 0.27 & 0.01 & --- & $\sim \text{15 minutes}$\\
    & Fantasia3D & \textbf{0.28} & 0.01 & 0.04 & $\sim \text{30 minutes}$  \\
    & Ours & \textbf{0.28} & \textbf{0.02} & \textbf{0.01} & $\textbf{$\sim$ 20 seconds}$ \\
    \midrule 
    \multirow{3}{*}{\rotatebox[origin=c]{90}{\emph{Airplane}}} & SketchShape & \textbf{0.27} & 0.02 & --- & $\sim \text{15 minutes}$\\
    & Fantasia3D & \textbf{0.27} & 0.02 & 0.05 & $\sim \text{30 minutes}$  \\
    & Ours & \textbf{0.27} & \textbf{0.04} & \textbf{0.02} & $\textbf{$\sim$ 20 seconds}$ \\
    \midrule 
    \multirow{3}{*}{\rotatebox[origin=c]{90}{\emph{Table}}} & SketchShape & 0.26 & 0.01 & --- & $\sim \text{15 minutes}$\\
    & Fantasia3D & 0.26 & 0.01 & 0.08 & $\sim \text{30 minutes}$  \\
    & Ours & \textbf{0.28} & \textbf{0.04} & \textbf{0.01} & $\textbf{$\sim$ 20 seconds}$ \\
    \midrule 
    \multirow{3}{*}{\rotatebox[origin=c]{90}{Average}} & SketchShape & 0.27 & 0.01 & --- & $\sim \text{15 minutes}$\\
    & Fantasia3D & 0.27 & 0.01 & 0.06 & $\sim \text{30 minutes}$  \\
    & Ours & \textbf{0.28} & \textbf{0.03} & \textbf{0.01} & $\textbf{$\sim$ 20 seconds}$ \\
\bottomrule
\end{tabular}
}
\caption{\textbf{Performance Breakdown for the Text-conditional Abstraction-to-3D Task.} Above we compare the performance of SketchShape~\cite{metzer2023latent} and Fantasia3D~\cite{chen2023fantasia3d} against ours over the three object categories in the text-conditional abstraction-to-3D task. 
Note that GD is not computed for SketchShape as it outputs a NeRF representation.
}
\label{tab:abstraction_comparison_detailed}
\end{table}

\subsection{Evaluation Breakdown}
We report performance per object category for the semantic editing task in Table \ref{tab:editing_comparison_supp} and for the text-conditional abstraction-to-3D in Table \ref{tab:abstraction_comparison_detailed}. As shown in the tables above, these detailed per-category metrics are consistent with the overall average scores reported in the main paper. In Table \ref{tab:editing_comparison_supp}, we also report the Frechet Pointcloud Distance (FPD) metric proposed by Shu \etal~\cite{fpd}, which measures the distance between the distribution of the output shapes the distribution of the input shapes. 
As illustrated in the table, our model produces outputs  that resemble the inputs better, in comparison to ChangeIt3D, as measured by lower FPD scores across all categories. 



\begin{table}[t]
\centering
\resizebox{\linewidth}{!}{
\begin{tabular}{llccc}
\toprule
 & Method & $\text{CLIP}_{Sim}\uparrow$ & $\text{CLIP}_{Dir}\uparrow$ & $\text{GD} \downarrow$ \\
 \midrule 
    \multirow{5}{*}{\rotatebox[origin=c]{90}{\emph{Chair}}} & ControlNet3D & \textbf{0.28} & \textbf{0.02} & 0.03 \\
    & SDEdit3D  & \textbf{0.28} & \textbf{0.02} & 0.06 \\ 
    & \shapE{}$_{\text{FT}}$  & \textbf{0.28} & \textbf{0.02} & 0.07 \\ 
    & CrossOnly  & 0.25 & 0.01 & 0.03 \\ 
    & Ours & \textbf{0.28} & \textbf{0.02} & \textbf{0.01} \\  
    \midrule 
    \multirow{5}{*}{\rotatebox[origin=c]{90}{\emph{Airplane}}} & ControlNet3D & 0.26 & 0.03 & 0.03 \\
    & SDEdit3D & \textbf{0.27} & \textbf{0.04} & 0.05 \\ 
    & \shapE{}$_{\text{FT}}$  & 0.26 & 0.03 & 0.05 \\ 
    & CrossOnly  & 0.25 & 0.03 & 0.03 \\
    & Ours & \textbf{0.27} & \textbf{0.04} & \textbf{0.02} \\  
\midrule 
    \multirow{5}{*}{\rotatebox[origin=c]{90}{\emph{Table}}} & ControlNet3D & \textbf{0.28} & 0.03 & 0.02 \\
    & SDEdit3D  & \textbf{0.28} & 0.03 & 0.05 \\
    & \shapE{}$_{\text{FT}}$  & \textbf{0.28} & 0.03 & 0.07 \\ 
    & CrossOnly  & \textbf{0.28} & 0.02 & 0.03 \\
    & Ours & \textbf{0.28} & \textbf{0.04} & \textbf{0.01} \\
    \midrule 
    \multirow{5}{*}{\rotatebox[origin=c]{90}{\emph{Average}}} & ControlNet3D & 0.27 & \textbf{0.03} & 0.03 \\
    & SDEdit3D  & \textbf{0.28} & \textbf{0.03} & 0.05 \\ 
    & \shapE{}$_{\text{FT}}$  & 0.27 & \textbf{0.03} & 0.06 \\
    & CrossOnly  & 0.26 & 0.02 & 0.03 \\
    & Ours & \textbf{0.28} & \textbf{0.03} & \textbf{0.01} \\  
\bottomrule
\end{tabular}
}
\caption{\textbf{Performance Breakdown for the Baseline Ablations.} We compare our cross-entity attention mechanism to three baseline methods over the three categories in the text-conditional abstraction-to-3D task. Our proposed mechanism allows for better preserving the input guidance condition, while maintaining a high fidelity to the target text prompt. For additional details on the baselines, see Section 5.2 in the main paper.}
\label{tab:ablations_abstractions}
\end{table}
\begin{table}[t]
\centering
\resizebox{0.88\linewidth}{!}{
\begin{tabular}{llccc}
\toprule
 & Method & $\text{CLIP}_{Sim}\uparrow$ & $\text{CLIP}_{Dir}\uparrow$ & $\text{GD} \downarrow$ \\
 \midrule 
    \multirow{4}{*}{\rotatebox[origin=c]{90}{\emph{Chair}}} 
    & w/o $\mathcal{Z}$ & 0.25 & 0.02 & 0.05 \\ 
    & w/ $K_{\times}$ & \textbf{0.28} & \textbf{0.02} & 0.02 \\ 
    & w/ $V_{\times}$ & \textbf{0.28} & \textbf{0.02} & 0.02 \\ 
    & Ours & \textbf{0.28} & \textbf{0.02} & \textbf{0.01} \\  
    \midrule 
    \multirow{4}{*}{\rotatebox[origin=c]{90}{\emph{Airplane}}} 
    & w/o $\mathcal{Z}$ & 0.24 & 0.03 & 0.03 \\ 
    & w/ $K_{\times}$ & 0.26 & 0.03 & \textbf{0.02} \\ 
    & w/ $V_{\times}$ & \textbf{0.27} & \textbf{0.04} & 0.03 \\ 
    & Ours & \textbf{0.27} & \textbf{0.04} & \textbf{0.02} \\  
    \midrule 
    \multirow{4}{*}{\rotatebox[origin=c]{90}{\emph{Table}}} 
    & w/o $\mathcal{Z}$ & 0.27 & 0.02 & 0.03 \\ 
    & w/ $K_{\times}$ &  \textbf{0.28} & 0.03 & 0.02 \\ 
    & w/ $V_{\times}$ &  \textbf{0.28} & 0.02 & 0.02 \\ 
    & Ours &  \textbf{0.28} & \textbf{0.04} & \textbf{0.01} \\
    \midrule 
    \multirow{4}{*}{\rotatebox[origin=c]{90}{\emph{Average}}} 
    & w/o $\mathcal{Z}$ & 0.25 & 0.02 & 0.04 \\ 
    & w/ $K_{\times}$ & 0.27 & 0.03 & 0.02 \\ 
    & w/ $V_{\times}$ & \textbf{0.28} & \textbf{0.03} & 0.02 \\ 
    & Ours & \textbf{0.28} & \textbf{0.03} & \textbf{0.01} \\  
\bottomrule
\end{tabular}
}
\caption{Additional ablations motivating our design choices for our cross-entity attention mechanism. We conduct experiments over several modifications to our proposed cross-entity attention block (more details regarding these modifications are provided in Section \ref{sec:add_ablations}). Performance is reported over the text-conditional abstraction-to-3D task, also  over each object category. As illustrated above, our proposed cross-entity attention mechanism can better preserve the input guidance shape, while maintaining higher fidelity to the target text prompt.}

\label{tab:ablations_additional_abstractions}
\end{table}
\subsection{Additional Ablations}
\label{sec:add_ablations}
We report per-category performance for the ablations shown in the main paper in
Table \ref{tab:ablations_abstractions}. As shown in the table, the baseline ablation methods cannot faithfully preserve the conditional guidance shape across all categories, yielding inferior GD scores compared to our approach. We perform these ablations on the same test sets used in our evaluations.

Next we conduct additional ablations to motivate our design choices, performing the following modifications to our cross-entity attention mechanism: (i) no zero-convolution operators, (ii) cross-entity attention over the Keys and (iii) cross-entity attention over the Values. 

Specifically, these modified cross-entity attention blocks are provided with the same input, which is a pair of latent vectors $(\mathbf{z}, \mathbf{c})$, where $\mathbf{z}$ denotes a noised latent code and $\mathbf{c}$ denotes a conditional latent that encodes structural information we would like to add to the original network. We allow all block parameters to optimize freely during model finetuning. For each experiment, the network is modified as follows:

\smallskip \noindent \textbf{No Zero-Convolution Operator} (w/o $\mathcal{Z}$). For this ablation, we remove the zero-convolution operator, which facilitates in ensuring that the network will not be effected by the conditional latent code when training (or finetuning) begins.

\smallskip \noindent \textbf{Cross-entity attention over the Keys} (w/ $K_{\times}$). Following the notations provided in the main paper, for this ablation we
perform the following additive operation:
\begin{equation}
K_{\times}=f_K(\phi(\mathbf{z}))+f_{K_c}(\mathcal{Z}(\phi_c(\mathbf{c}))),
\end{equation}
where $f_{K_c}$ and $\phi_c(\mathbf{c})$ are a learned linear layer and intermediate features, initialized randomly.  
The output of our cross-entity attention block is $\mathbf{z}_{out} = Attn(Q,K_\times,V)$. 

\smallskip \noindent \textbf{Cross-entity attention over the Values} (w/ $V_{\times}$). For this ablation we
perform the following additive operation:
\begin{equation}
V_{\times}=f_V(\phi(\mathbf{z}))+f_{V_c}(\mathcal{Z}(\phi_c(\mathbf{c}))),
\end{equation}
where $f_{V_c}$ and $\phi_c(\mathbf{c})$ are a learned linear layer and intermediate features, initialized randomly.  
The output of our cross-entity attention block is $\mathbf{z}_{out} = Attn(Q,K,V_\times)$. 
\ignorethis{
\smallskip \noindent \textbf{Cross-entity attention using only conditional Queries} (w/o $Q_{\times}$). For this ablation, we design our cross entity attention only over the conditional Queries. That is, 
the output of our cross-entity attention block is $\mathbf{z}_{out} = Attn(Q_{c},K,V)$. 
}

\medskip
Results over these additional ablations are reported in Table \ref{tab:ablations_additional_abstractions}. 
As shown in the table, our proposed cross-entity attention mechanism allows for better preserving the structure of the guidance shape, yielding significantly lower GD score. This is accomplished while also manipulating the structure of the input shape and maintaining a high fidelity to the target text prompt, achieving better $\text{CLIP}_{Dir}$ and $\text{CLIP}_{Sim}$ scores. 

\end{document}